\documentclass[final]{IEEEtran}
\usepackage[noadjust]{cite}
\usepackage{graphicx}
\usepackage{amsmath,amssymb} 
\usepackage{color}
\usepackage{epsfig}
\usepackage{comment}
\usepackage{kevinlin}
\usepackage[ruled,vlined]{algorithm2e}
\usepackage{enumitem}
\usepackage[export]{adjustbox}
\usepackage{tabularx}
\usepackage[table]{xcolor}
\usepackage{colortbl}
\usepackage{subfig}
\usepackage{multirow}
\usepackage{booktabs}
\usepackage{xr}
\usepackage{pifont}
\usepackage[hyphens]{url}
\usepackage{hyperref}

\usepackage{sidecap}

\usepackage{amssymb}
\usepackage{amsmath}

\usepackage{ragged2e}

\ifCLASSINFOpdf
\else
\fi
\begin{document}
%
\title{Cross-Domain Complementary Learning Using Pose for Multi-Person Part Segmentation}
%
%
%

\author{Kevin~Lin,~\IEEEmembership{Student Member, IEEE}, Lijuan Wang,~\IEEEmembership{Senior Member, IEEE}, Kun Luo, Yinpeng Chen,~\IEEEmembership{Member, IEEE}, Zicheng Liu,~\IEEEmembership{Fellow, IEEE}, Ming-Ting Sun,~\IEEEmembership{Life Fellow, IEEE}
\vspace{-5mm}
\thanks{Manuscript received Sep. 30, 2019; revised Feb. 1, 2020 and Apr. 7, 2020; accepted May 3, 2020.}
\thanks{K.~Lin and M.-T.~Sun are with the Department of Electrical and Computer Engineering, University of Washington, Seattle, WA, 98195. E-mail: \{kvlin,~mts\}@uw.edu}
\thanks{L. Wang, K. Luo, Y. Chen, and Z. Liu are with Microsoft Azure+AI, Redmond, WA, 98052. E-mail: \{lijuanw,~kun.luo,~yiche,~zliu\}@microsoft.com}
}

%
%

\markboth{IEEE Transactions on Circuits and Systems for Video Technology}%
{Lin \MakeLowercase{\textit{et al.}}: Cross-Domain Complementary Learning Using Pose for Multi-Person Part Segmentation}
%

\IEEEpubid{\begin{minipage}{\textwidth}\ \\[12pt] \centering
  Copyright~\copyright~2020 IEEE. Personal use is permitted. However, permission to use this material for any other purposes\\ must be obtained from the IEEE by sending an email to pubs-permissions@ieee.org.
\end{minipage}}


\maketitle

\begin{abstract}
Supervised deep learning with pixel-wise training labels has great successes on multi-person part segmentation. However, data labeling at pixel-level is very expensive. To solve the problem, people have been exploring to use synthetic data to avoid the data labeling. Although it is easy to generate labels for synthetic data, the results are much worse compared to those using real data and manual labeling. The degradation of the performance is mainly due to the domain gap, i.e., the discrepancy of the pixel value statistics between real and synthetic data. In this paper, we observe that real and synthetic humans both have a skeleton (pose) representation. We found that the skeletons can effectively bridge the synthetic and real domains during the training.  Our proposed approach takes advantage of the rich and realistic variations of the real data and the easily obtainable labels of the synthetic data to learn multi-person part segmentation on real images without any human-annotated labels. Through experiments, we show that without any human labeling, our method performs comparably to several state-of-the-art approaches which require human labeling on Pascal-Person-Parts and COCO-DensePose datasets. On the other hand, if part labels are also available in the real-images during training, our method outperforms the supervised state-of-the-art methods by a large margin. We further demonstrate the generalizability of our method on predicting novel keypoints in real images where no real data labels are available for the novel keypoints detection. Code and pre-trained models are available at \url{https://github.com/kevinlin311tw/CDCL-human-part-segmentation}.
\end{abstract}


%

\section{Introduction}\label{sec:intro}
\IEEEPARstart{H}{uman} body part segmentation~\cite{chen2014semantic,chen2018deeplab,chen2016attention,xia2016zoom} aims at partitioning persons in the image to multiple semantically consistent regions (\eg, head, arms, legs), which is important to many human-centric analysis applications~\cite{gu2018ava,Guler2018DensePose,ionescu2014human3,tong2013upper}. Supervised training with deep Convolutional Neural Networks (CNNs) significantly improves the performance of various visual recognition tasks including the human body part segmentation~\cite{Guler2018DensePose, chen2018deeplab,krizhevsky2012imagenet, long2015fully, zheng2015conditional}. However, it requires large amount of training data. Data labeling, especially at pixel level, is labor intensive and the acquisition of such annotations in large scale is prohibitively expensive.

A promising solution to address this problem is to take advantage of the graphics simulator to generate synthetic images with ground truths automatically~\cite{ marin2010learning,richter2016playing,ros2016synthia}. For example, previous study~\cite{varol2017learning} proposed to learn single-person part segmentation by directly training the neural networks using synthetic images. However, their method usually produces false alarms in the real-world background, and it does not work well for real-world images consisting of multiple person with interactions and occlusions. Also, recent studies~\cite{tobin2017domain, tremblay2018training, vazquez2014virtual} show that the discrepancy of the pixel value statistics between real and synthetic data, so called the domain gap, makes it challenging to transfer knowledge from synthetic domain to real domain. In addition to the pixel value statistics, the discrepancy of the content distributions (\eg, the background scenes and objects) between the two domains makes knowledge transfer even more difficult.

\begin{figure*}[t]
	\centering
\includegraphics[width=1.0\textwidth]{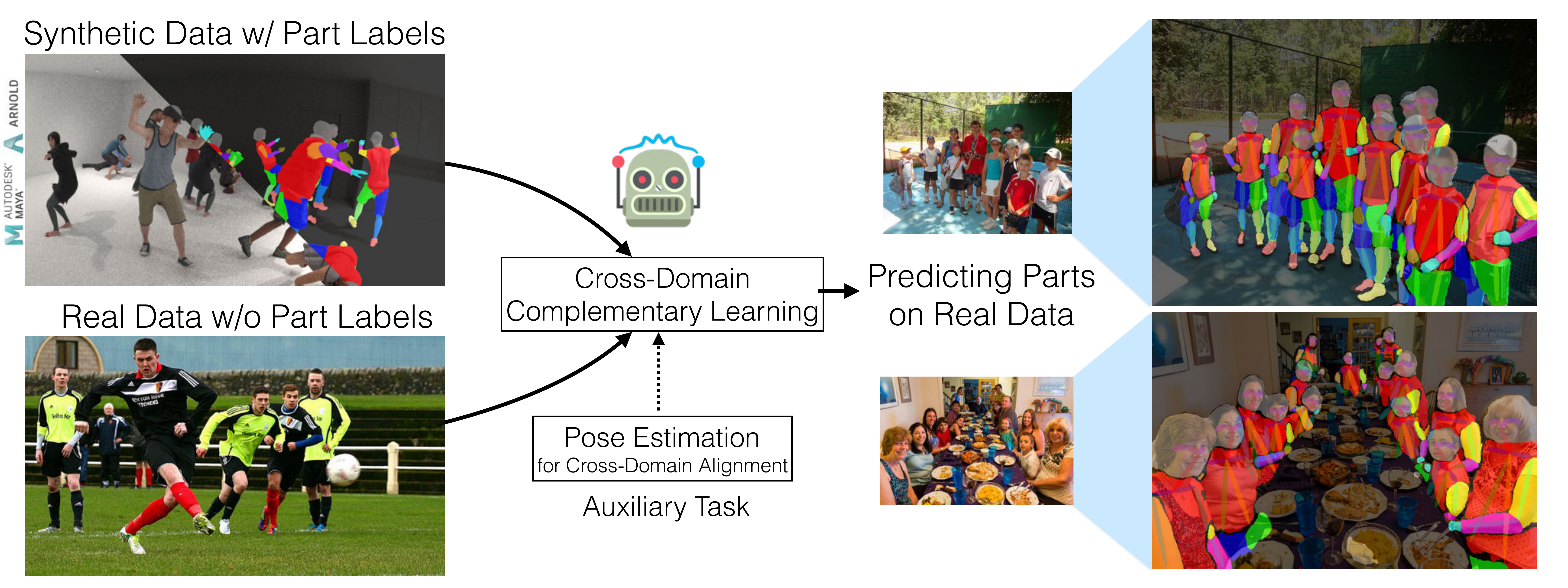}
	\caption{We address the problem of learning multi-person part segmentation without human labeling. Our proposed complementary learning technique learns a neural network model for multi-person part segmentation using a synthetic dataset and a real dataset. We observe that real and synthetic humans share a common skeleton structure. During learning, the proposed model extracts human skeletons which effectively bridges the synthetic and real domains. Without using human-annotated part segmentation labels, the resultant model works well on real world images.}
	\label{fig:first}
\end{figure*}

To address the discrepancies of the content distributions and the pixel value statistics between the two domains, recent studies~\cite{ren2017cross,Tsai_adaptseg_2018,tzeng2017adversarial} proposed to train the neural networks using adversarial training for matching the feature distributions of the real and synthetic data. They proposed to train a discriminator for distinguishing the real and synthetic images, and a generator for extracting the domain-invariant features that can fool the discriminator. However, the adversarial training may not converge due to the fact that it is difficult to maintain a balanced training between the generator and the discriminator. Previous approaches also suffer from the issue of mode collapse, where the generator may only capture a part of the real data distribution. Thus, the performances of previous approaches are much worse than the supervised training on real data with pixel-wise manual labeling.

In this paper, we observe that real and synthetic humans both have a skeleton (pose) representation and show that the skeletons can effectively bridge the synthetic and real domains during the training.  With our proposed approach, we can take advantage of the complementary nature of the real and synthetic data, i.e., rich and realistic variations of the real data and the easily obtainable labels of the synthetic data, effectively. Our technique learns multi-person part segmentation on real images without any human-annotated labels and achieves performance comparable to several state-of-the-art approaches which require human labeling. On the other hand, if part labels are also available in the real-images during training, our method outperforms the supervised state-of-the-art methods by a large margin.
\IEEEpubidadjcol
As shown in Figure~\ref{fig:first}, we have part segmentation labels from synthetic data, but do not have part segmentation labels from real data. It should be noted that our synthetic images have an extremely simple background with white walls, while the real images have complex backgrounds with a variety of non-human objects. Given such discrepancies between the two domains, we observe that real and synthetic humans both have a common skeleton representation. By learning the skeleton representation of the real and synthetic humans, our proposed model learns a shared feature space for both real and synthetic domains. Different from previous works that try to minimize the discrepancy of the pixel value statistics between the domains, we propose to perform human pose estimation to extract skeletons from the real and synthetic images, and minimize the discrepancy of the feature spaces between the two domains by learning the domain-invariant human skeleton representation. The automatically extracted skeletons capture the structural body information and can effectively bridge the real and synthetic data domains, so that both real and synthetic data can be used in the training effectively without needing the expensive manual human part labeling for the real images. 

It is worth noting that the learning of human pose estimation requires training labels. However, the pose labels are readily available on several public large-scale datasets like COCO Keypoint dataset~\cite{lin2014microsoft} and are easy to obtain than part segmentation labels. Thus, the proposed method has the advantage of saving labeling efforts in practice.

We also show that our method can be generalized to predict a new set of keypoints for real images. For example, to predict keypoints on hands and feet, we just need to generate synthetic images with hands and feet labels, and the knowledge will transfer from the synthetic domain to the real domain using our proposed approach.

In summary, the main contributions of this paper include:
\begin{itemize}
\item We discover that human pose is very effective to bridge the real and synthetic domains for human-centric analysis applications.

\item We introduce an effective framework, called cross-domain complementary learning with pose, to leverage information in both real and the synthetic images for multi-person part segmentation.

\item Through experiments, we show that without any human-annotated part segmentation label, our method performs comparably with several state-of-the-art approaches which require human labeling on Pascal-PersonParts and COCO-DensePose datasets. On the other hand, if parts labels are also available in real images during training, our method outperforms the supervised state-of-the-art methods by a large margin. 

\item We show that our method can be generalized to predict new keypoints such as those on hands and feet in real images without human labeling. 
\end{itemize}

\section{Related Work}\label{sec:intro}
\subsection{Synthetic data for computer vision tasks}
There has been a long-standing history of exploring the use of 3D synthetic data for computer vision problems~\cite{liebelt2010multi, nevatia1977description, shilane2004princeton}. Recent studies use 3D CAD models for visual recognition tasks, such as 3D model repository~\cite{chang2015shapenet,wu2016learning}, object recognition~\cite{peng2015learning, Qi_2016_CVPR, tremblay2018training}, human analysis applications~\cite{ionescu2014human3,marin2010learning,varol2017learning,vazquez2014virtual}, and semantic segmentation for urban scenes~\cite{ros2016synthia}. Among the literature, Varol~\etal~\cite{varol2017learning} proposed to render a single-person avatar on top of a static background image, and generate ground truths for training deep CNNs. However their method only works for the well-controlled environment and the single-person scenario in an image. This is because it is difficult and expensive to render photorealistic images with rich coverage of avatars, background scenes, and objects.

In this work, we address a more challenging and general scenario, where multiple people with interactions and occlusions are considered. Different from training the deep CNNs using synthetic data only~\cite{varol2017learning}, we propose to leverage the complementary natural of the real and synthetic data with human pose estimation. In the experiments, we show that our method, which learns to bridge the reality gap, performs more favorably against those proposed in previous studies~\cite{varol2017learning}. In addition, as demonstrated in the experiments, our technique reduces the requirement on the photorealism of the synthetic data generation.

\subsection{Domain adaptation}
Domain adaptation is a special case of transfer learning~\cite{pan2010survey} that aims to learn a single task from a source domain, so that it performs well on a target domain. Many approaches have been proposed to address the \textit{visual} dataset bias~\cite{torralba2011unbiased} for domain adaptation, including active learning with human-in-the-loop~\cite{ vazquez2014virtual}, training deep CNNs with reverse gradient~\cite{ganin2015unsupervised}, learning with auxiliary tasks to reduce domain variations~\cite{bousmalis2016domain, yu2016learning}, and matching feature distributions of two domains by adversarial training~\cite{zhao2019multi,Tsai_2019_ICCV,chen2019learning,chen2019crdoco, zhang2018fully,luo2019taking,ren2017cross,Tsai_adaptseg_2018,tzeng2017adversarial}.
In particular, Chen~\etal~\cite{chen2019learning} proposed an image-level adaptation approach which tries to make the appearance of synthetic images similar to real images. One key assumption of~\cite{chen2019learning} is that the content distribution of the synthetic data is similar to the content distribution of the real data. It is not suitable to our research problem because all of our synthetic images have an extremely simple background (empty room with white walls) while real images have complex backgrounds with a variety of objects. The discriminators can easily distinguish our synthetic images from real images thus making the adversarial learning scheme ineffective. Instead of image-level adaptation, Ren and Lee~\cite{ren2017cross} proposed a feature-level adaptation approach to learn image classifiers and object detectors using synthetic images with adversarial training.
A recent study~\cite{Hoffmann:GCPR:2019} proposed to learn human pose estimation with synthetic data using adversarial teacher-student network.
Tsai~\etal~\cite{Tsai_2019_ICCV} further proposed to enhance the adversarial learning with patch-level alignment. However, existing domain adaptation approaches do not work as well as the fully supervised training approaches.
Instead of adversarial training, our approach uses an auxiliary task of human pose estimation to bridge synthetic and real domains, which is shown to be more effective from our experiments.

\subsection{Multi-task learning}
Prior works~\cite{du2017rpan,gkioxari2014r,he2017mask,li2014heterogeneous, pan2010survey, popa2017deep, zhou2017towards} have shown that multi-task learning is effective for many vision problems. Given multiple different tasks, where a subset of these tasks are related, multi-task learning aims to improve the learning of the original task by using knowledge from all or some of the other tasks~\cite{ruder2017overview,zhang2017survey}.  However, many previous studies assume that, for all the tasks, the labeled data have to be available for training~\cite{pan2010survey}. Different from the previous works, our method learns without human-annotated segmentation labels in a cross-domain scenario, and learns to bridge the domain gap between real and synthetic data.

\begin{figure}[t]
	\centering
\includegraphics[width=.8\columnwidth]{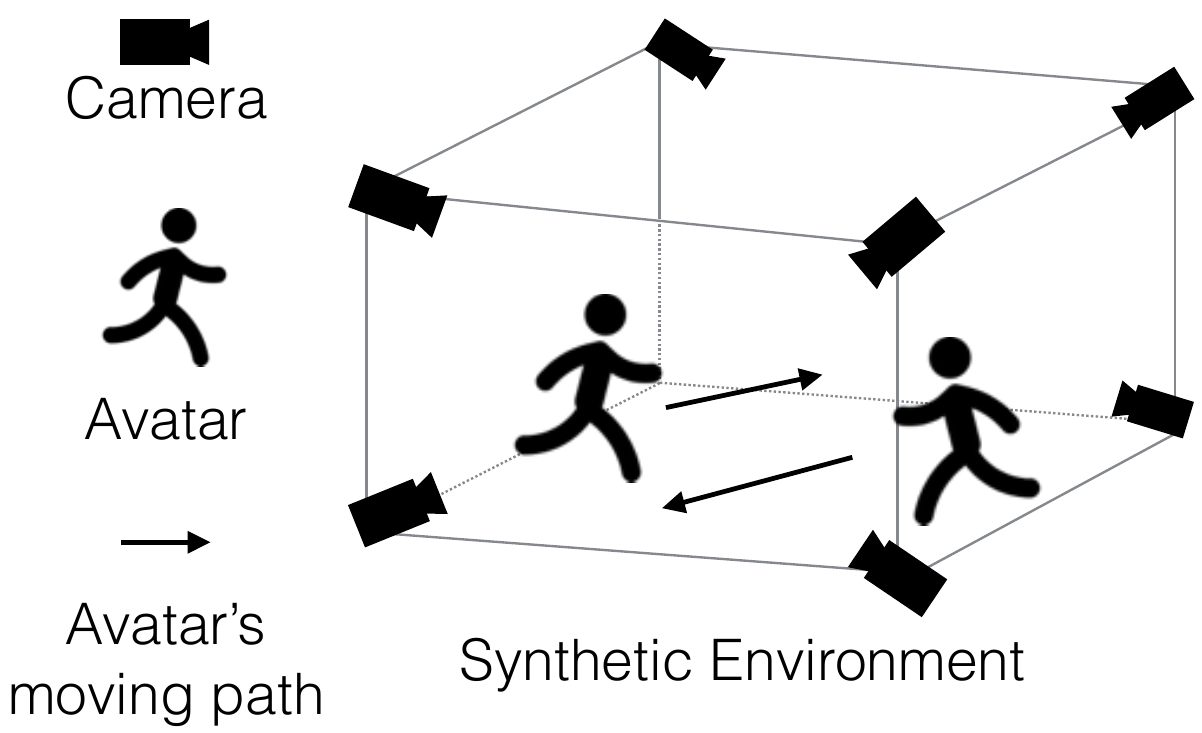}
	\caption{The layout of our synthetic environment. We render multiple avatars performing different actions in a 3D room, and capture the animations from multiple different viewpoints. }
	\label{fig:syn_design}
\end{figure}

\subsection{Supervised and semi-supervised part segmentation}
Recent studies~\cite{dong2014towards, gong2017look, ladicky2013human,xia2017joint,xia2016pose,papandreou2018personlab} proposed to jointly train human part segmentation and human pose estimation for improving the performance of part segmentation. However, the successes of the previous studies are mainly attributed to the supervised training with the pixel-wised manual labeling. Different from the fully supervised approaches, we propose to remove the manual labeling requirement by learning with synthetic data. On the other hand, Fang~\etal~\cite{fang2018weakly} proposed a semi-supervised approach that aims to augment training samples by transferring the human-labeled part segmentation from an existing dataset to another unlabeled dataset. Our method differs from theirs in that our method does not require any human-labeled part segmentation dataset at all.

Bearman~\etal~\cite{bearman2016s} proposed a point-level supervision which is related to our work. The key insight of their method is to use the foreground masks (called objectness prior in~\cite{bearman2016s}) to help find the foregrounds. Their method is effective for extracting the foreground regions, but it remains challenging to find the boundaries between different foreground objects such as the object parts in our problem. For example, when a person's arm is in front of the torso, the arm region is overlapped with the torso region making it difficult to find the boundary of the arm. Instead of relying on objectness prior, we leverage synthetic data to learn the boundaries between different parts.

\begin{figure*}[tb]
 \centering
 \setlength{\tabcolsep}{1pt}
 \begin{tabular}{ccccccc}
 	\includegraphics[width=.14\textwidth]{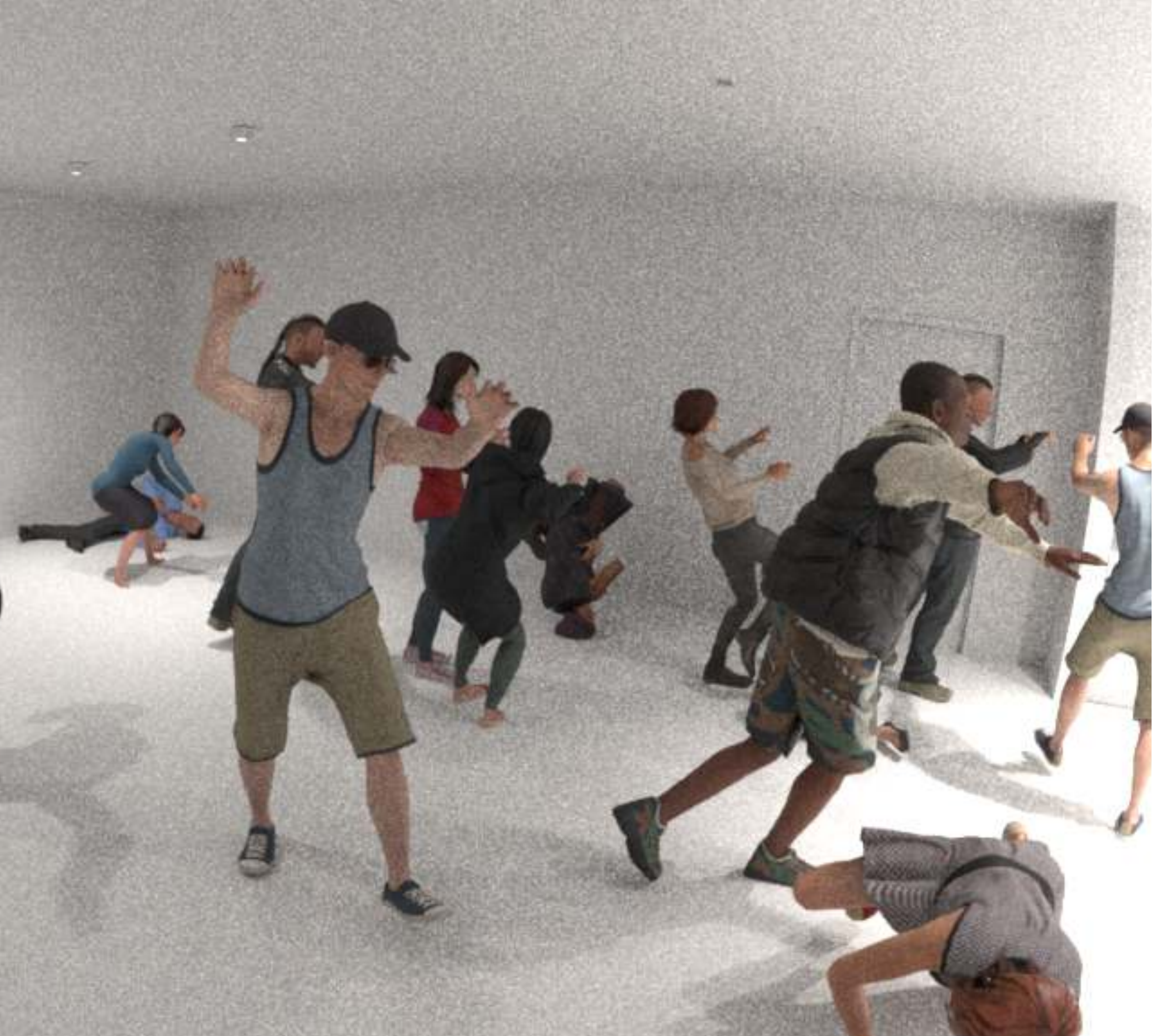}
 	&\includegraphics[width=.14\textwidth]{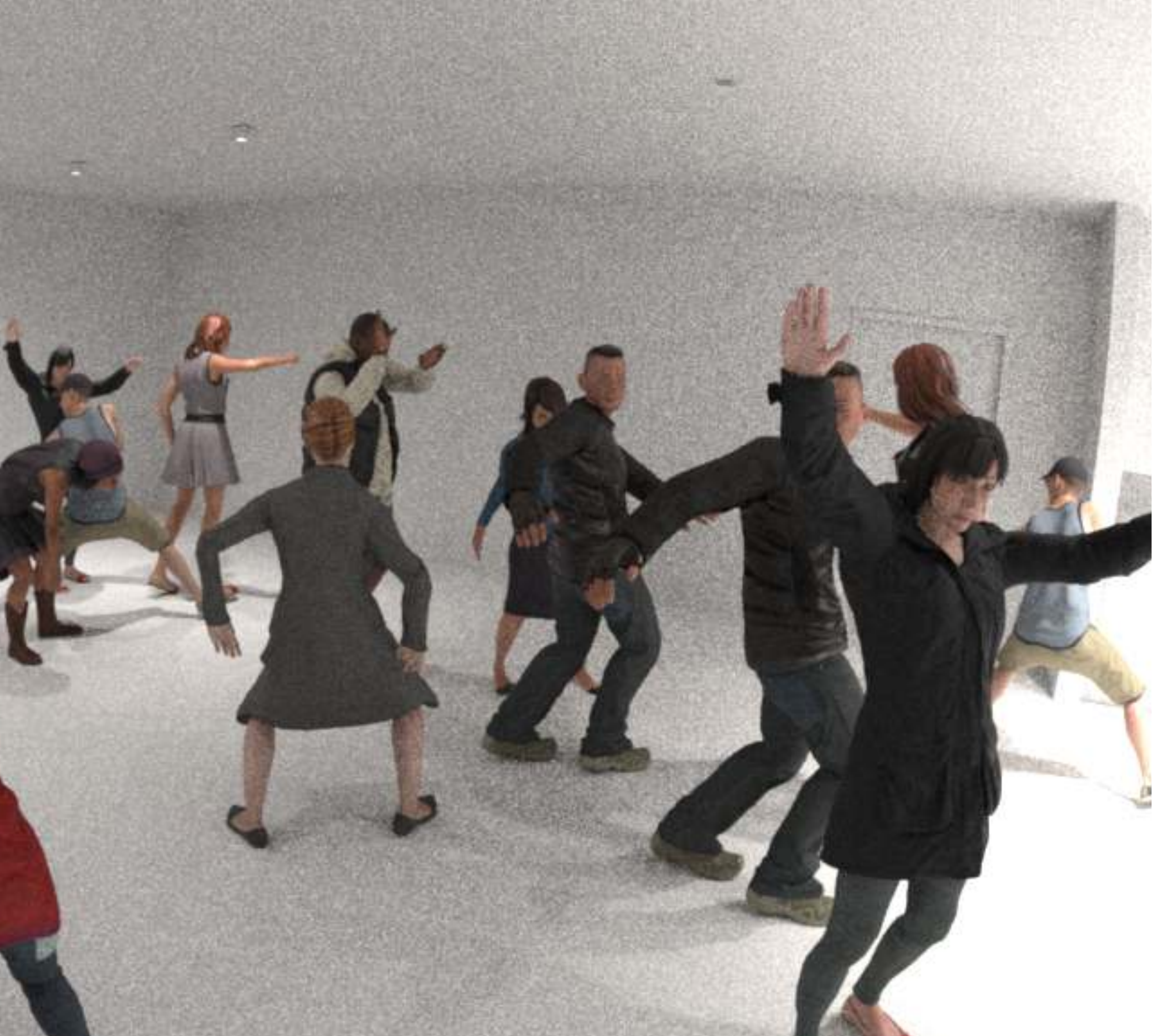}
 	&\includegraphics[width=.14\textwidth]{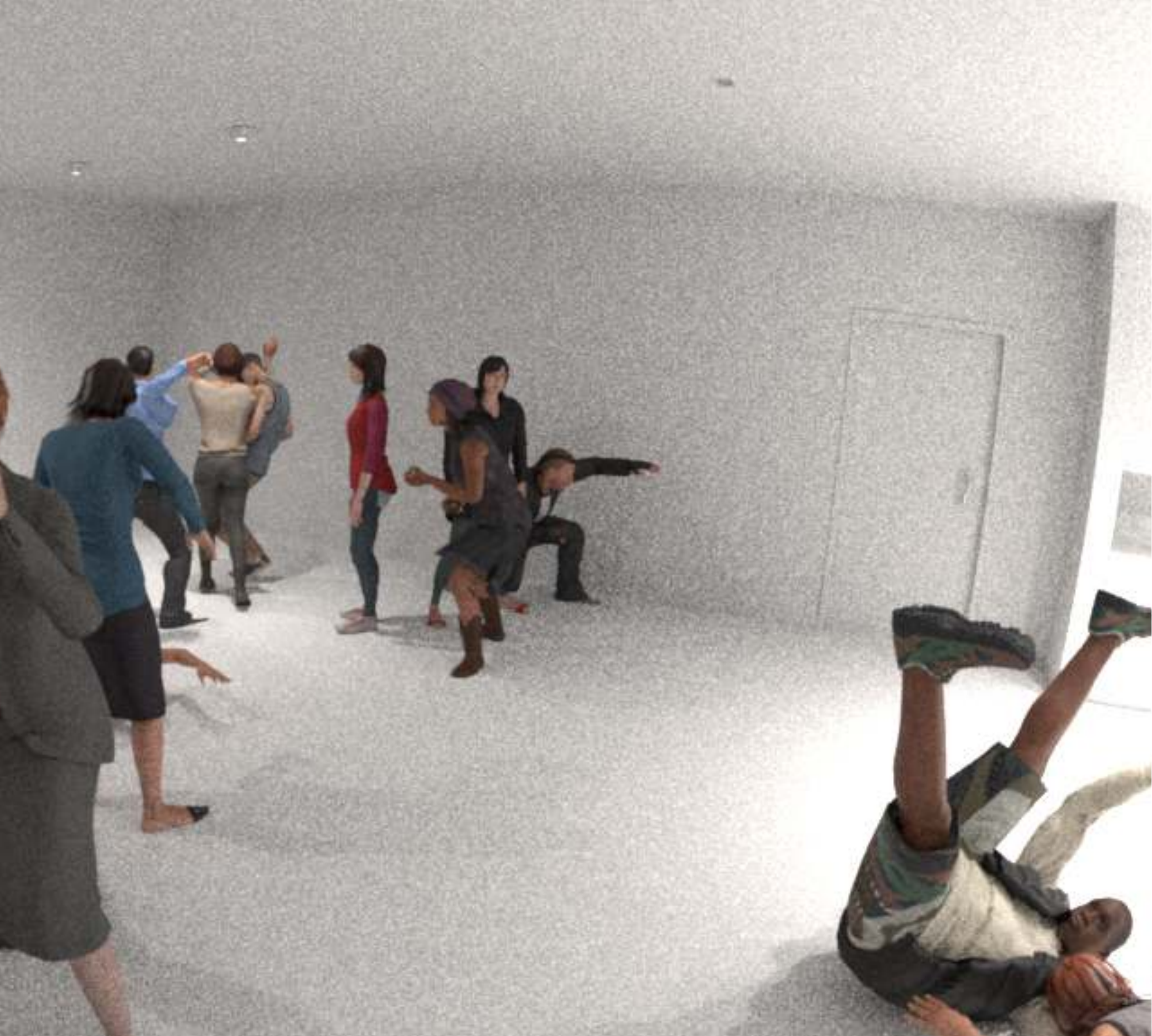}
 	&\includegraphics[width=.14\textwidth]{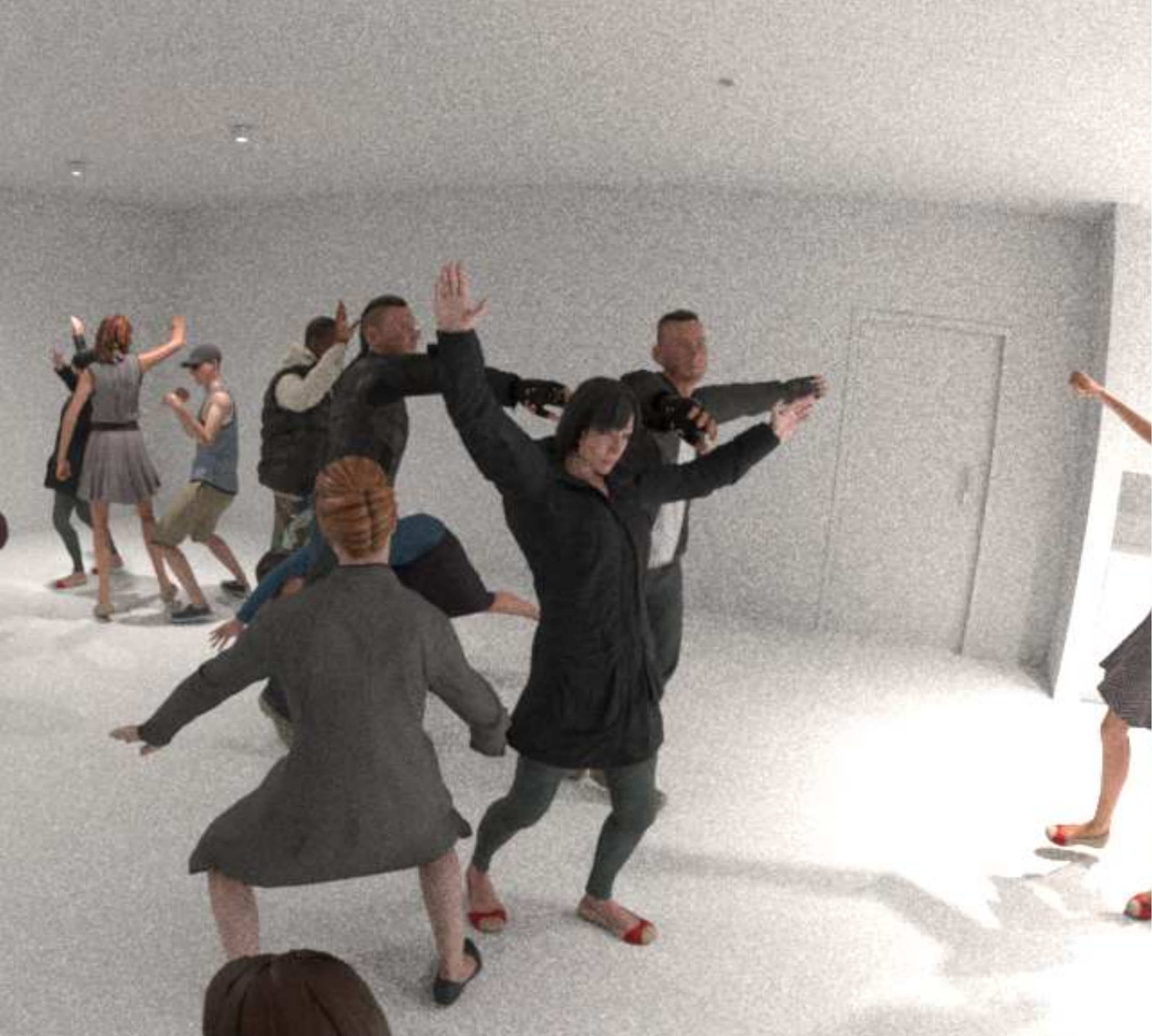}
 	&\includegraphics[width=.14\textwidth]{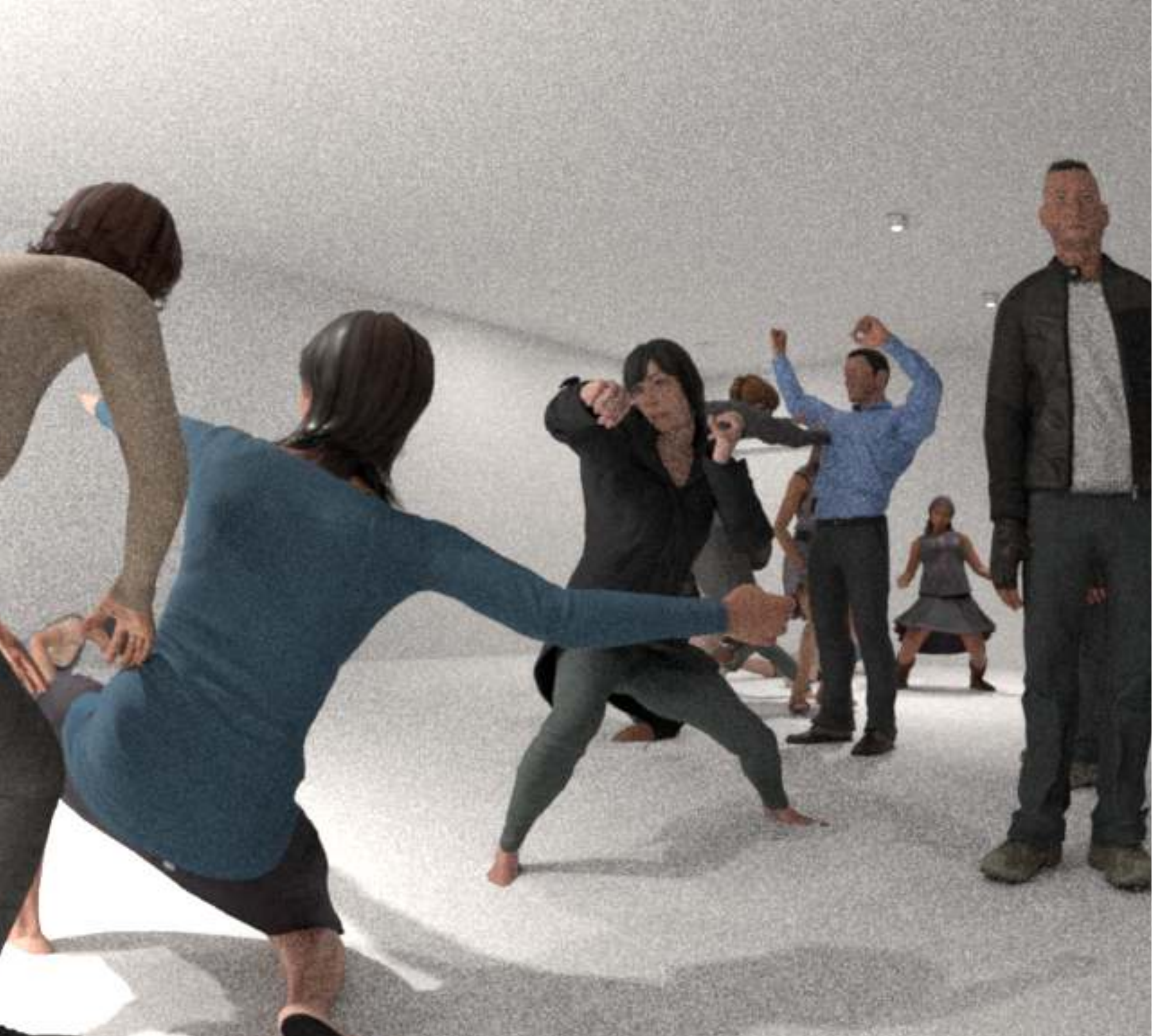}
 	&\includegraphics[width=.14\textwidth]{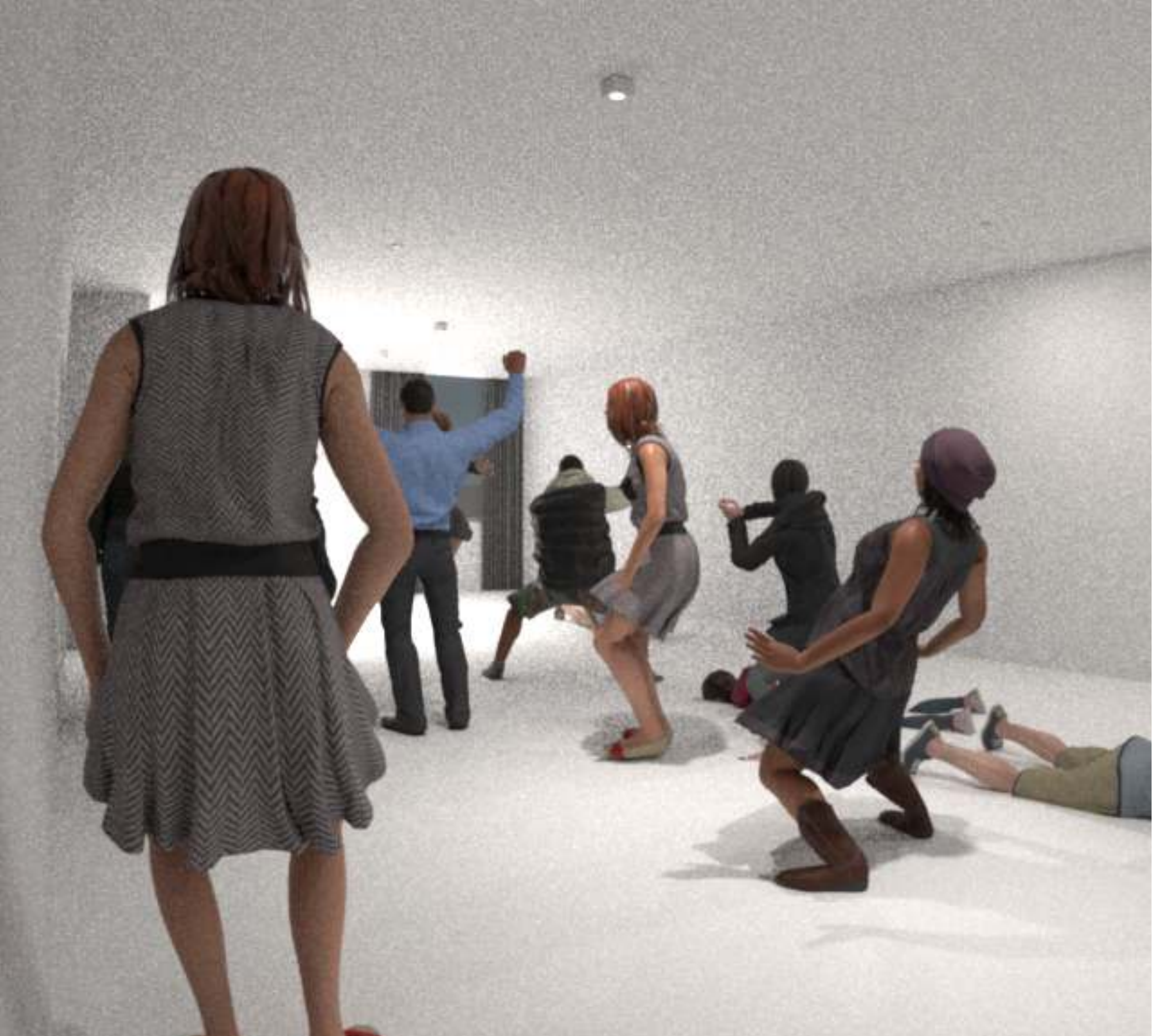}
 	&\includegraphics[width=.14\textwidth]{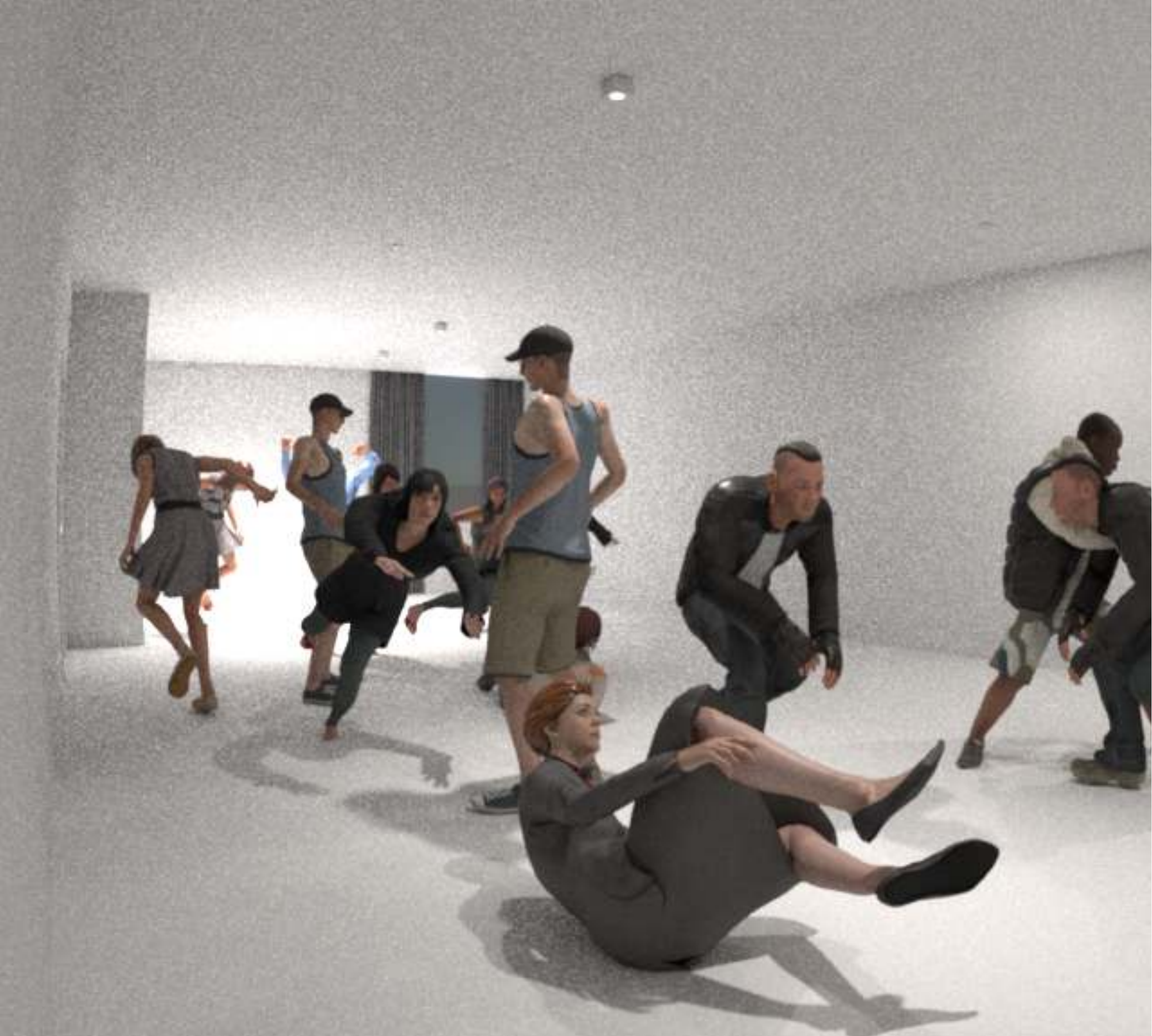}\\
 	\includegraphics[width=.14\textwidth]{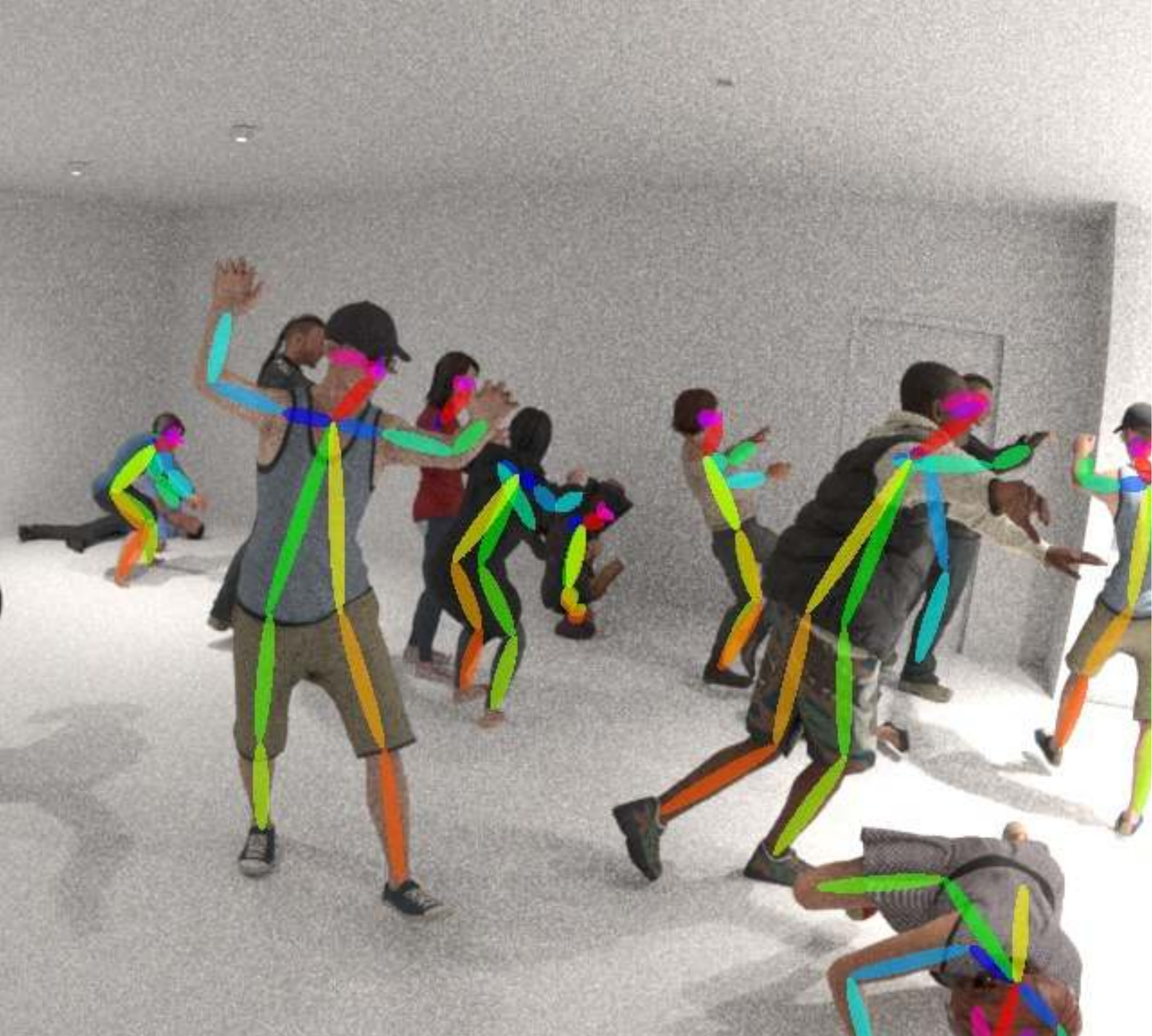}
 	&\includegraphics[width=.14\textwidth]{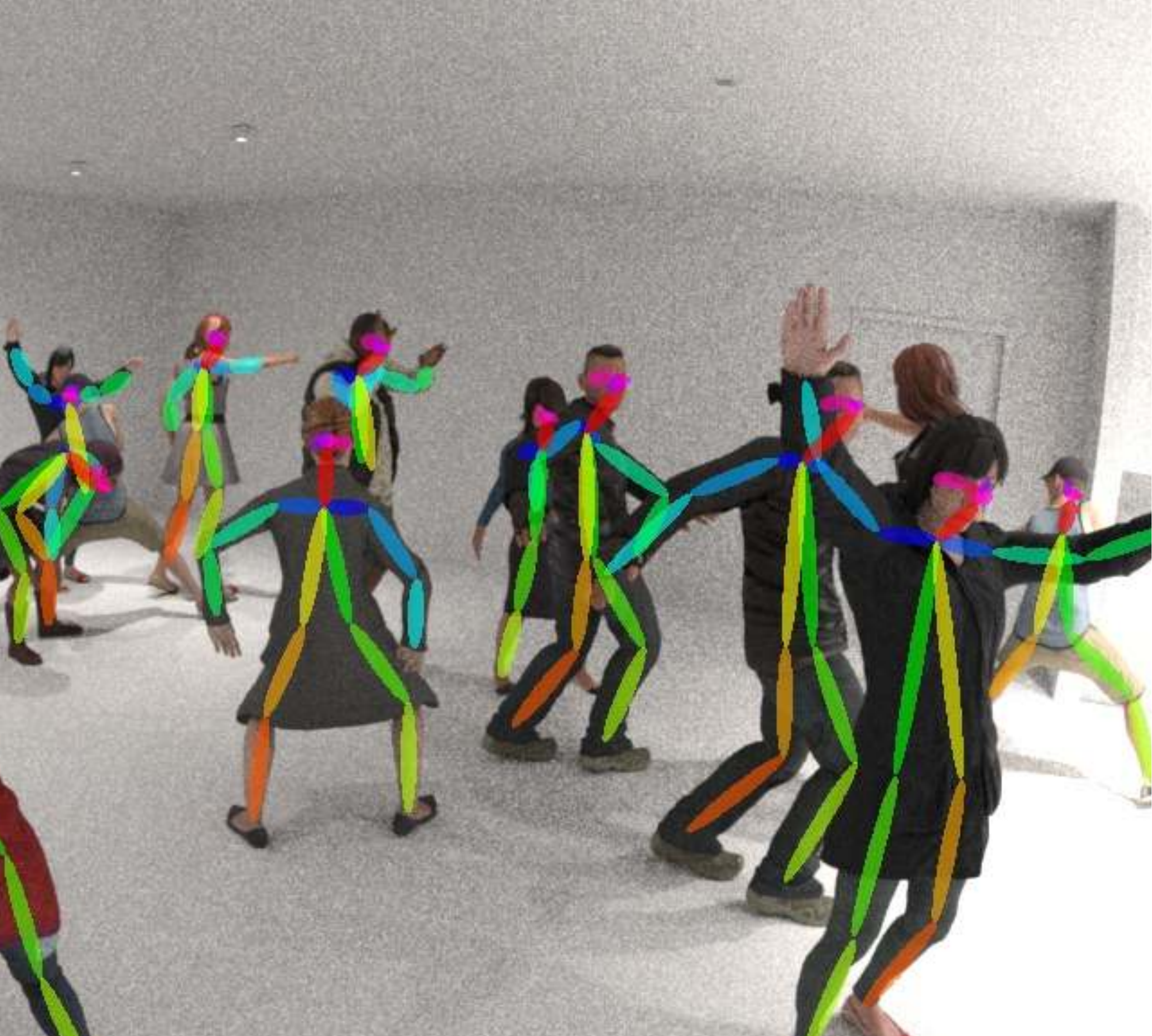}
 	&\includegraphics[width=.14\textwidth]{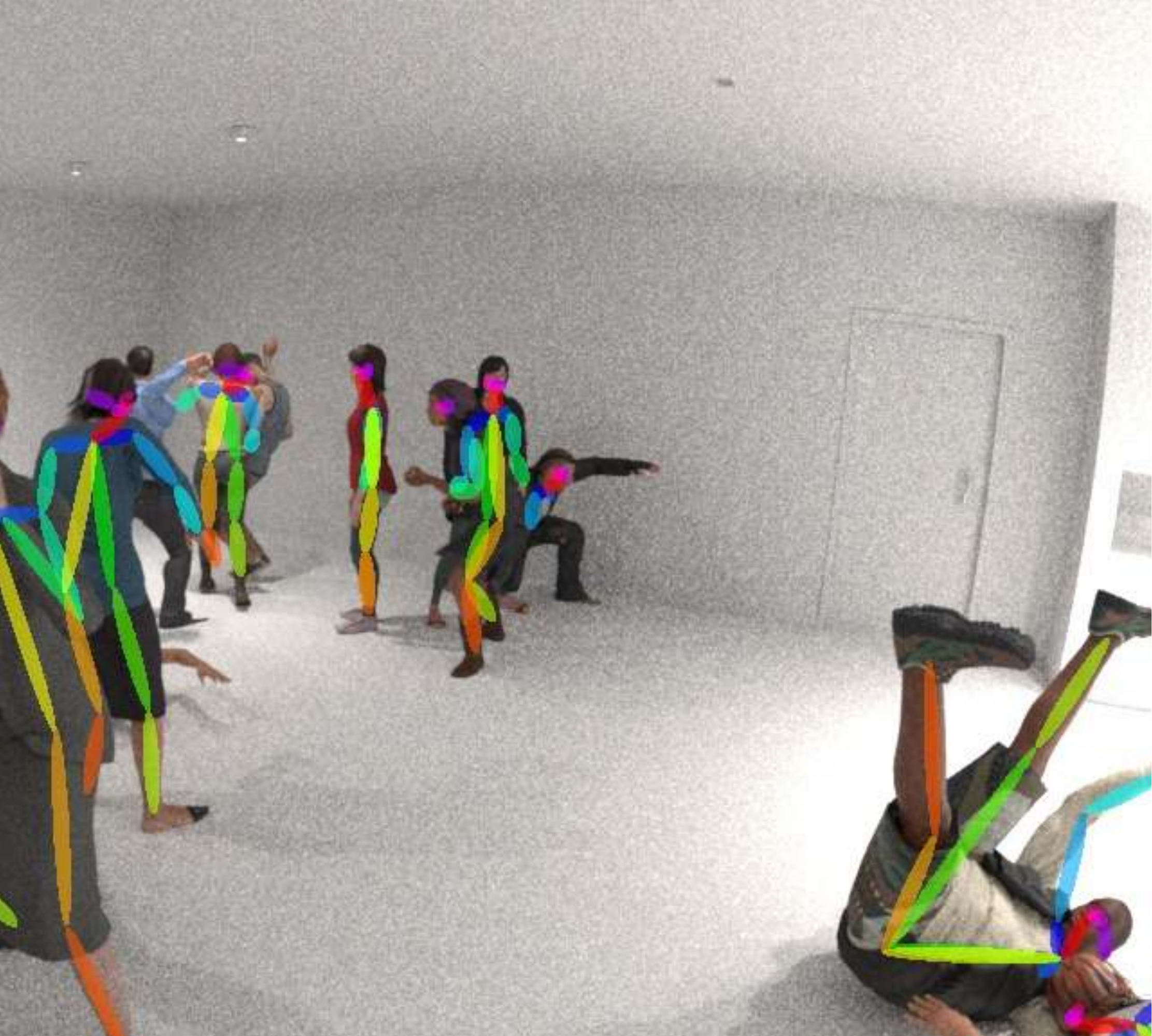}
 	&\includegraphics[width=.14\textwidth]{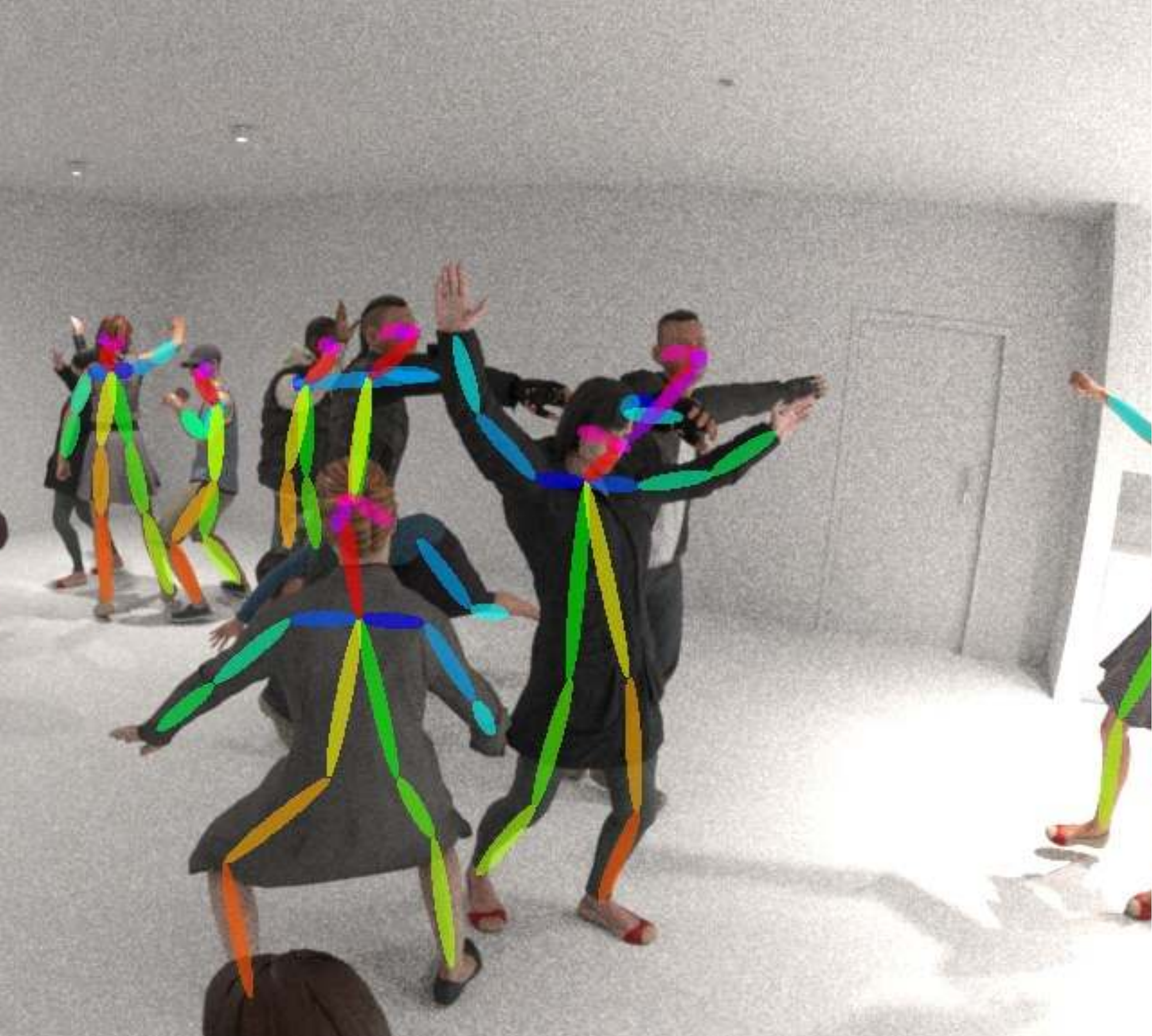}
 	&\includegraphics[width=.14\textwidth]{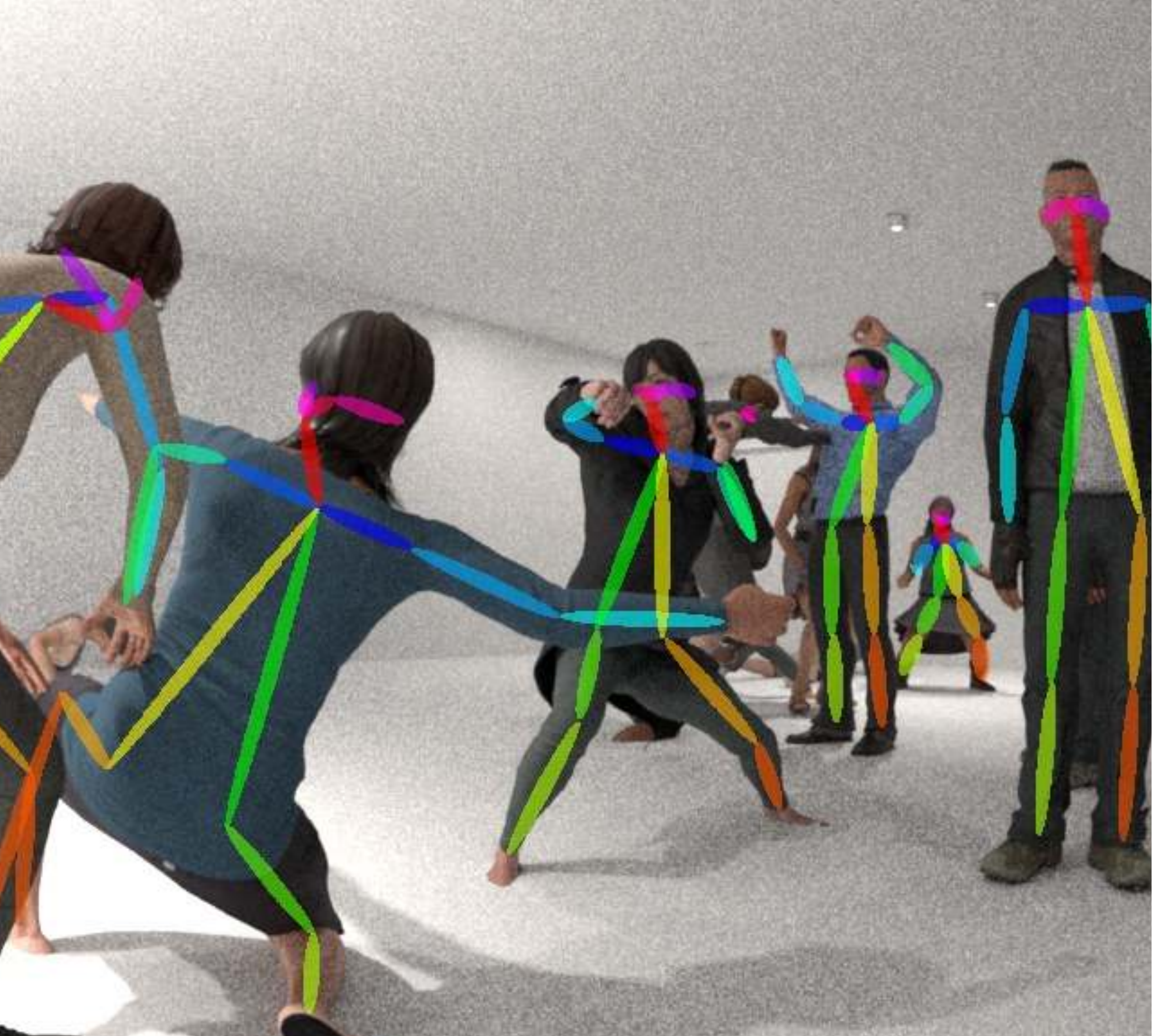}
 	&\includegraphics[width=.14\textwidth]{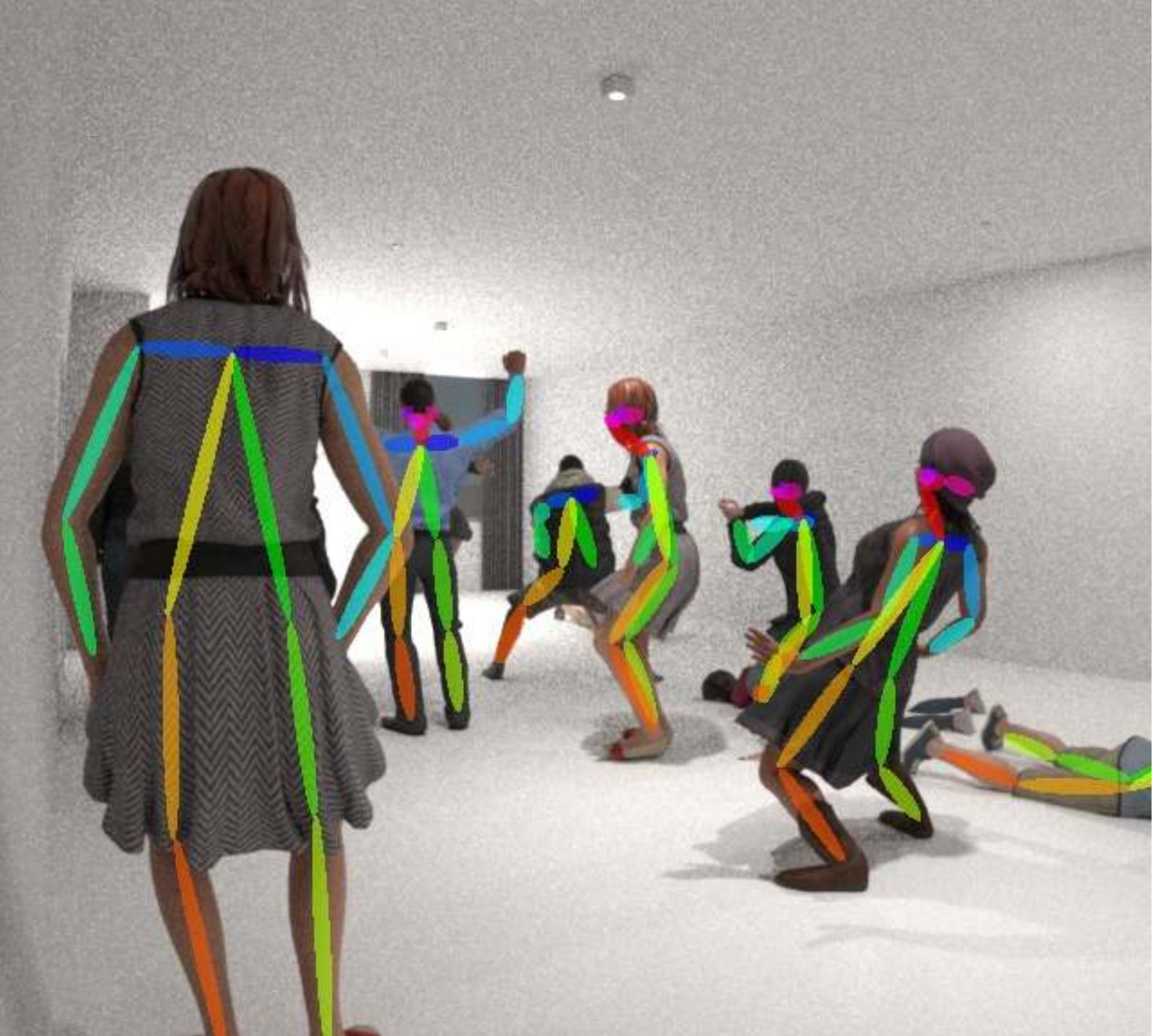}
 	&\includegraphics[width=.14\textwidth]{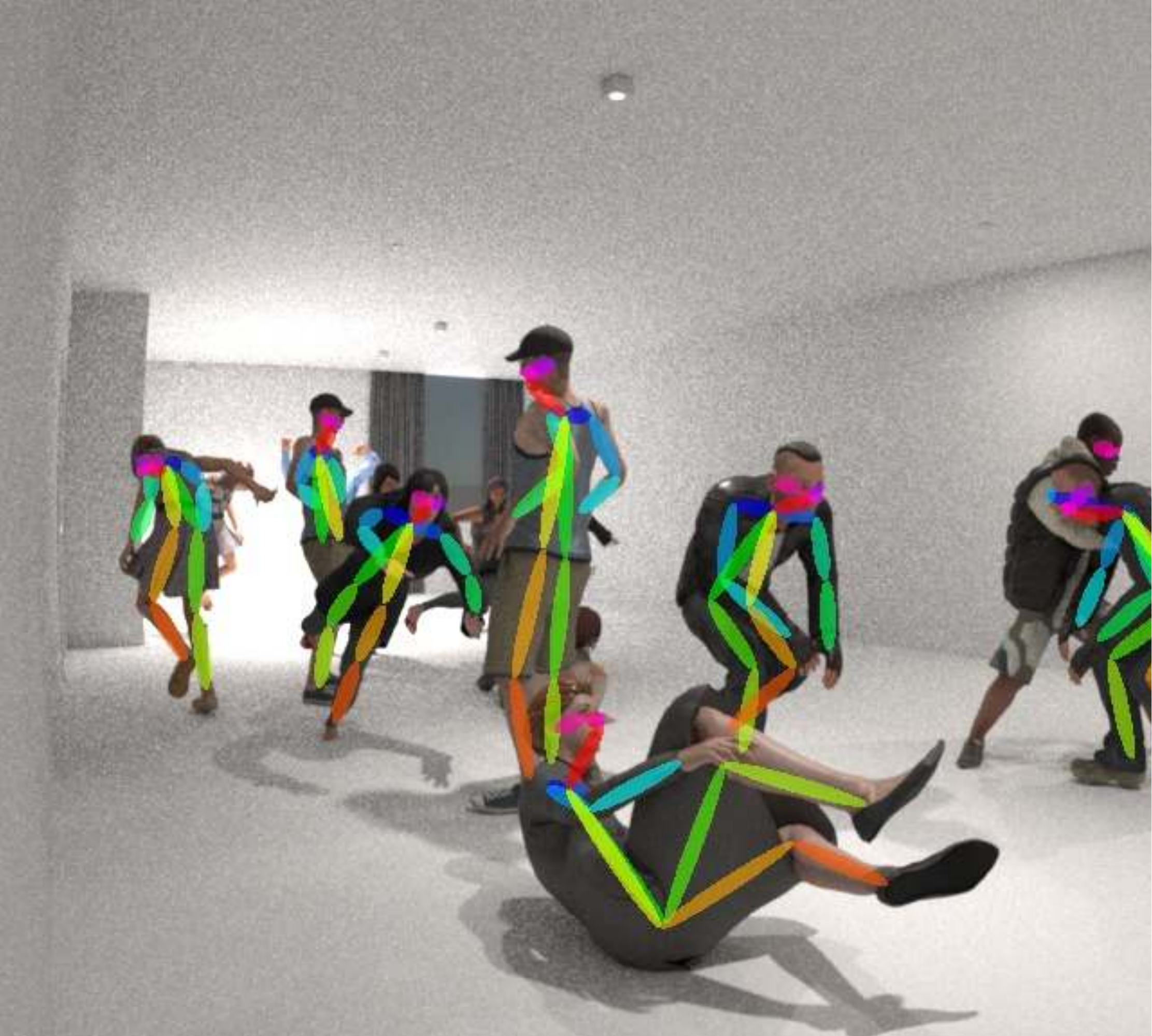}\\
 	\includegraphics[width=.14\textwidth]{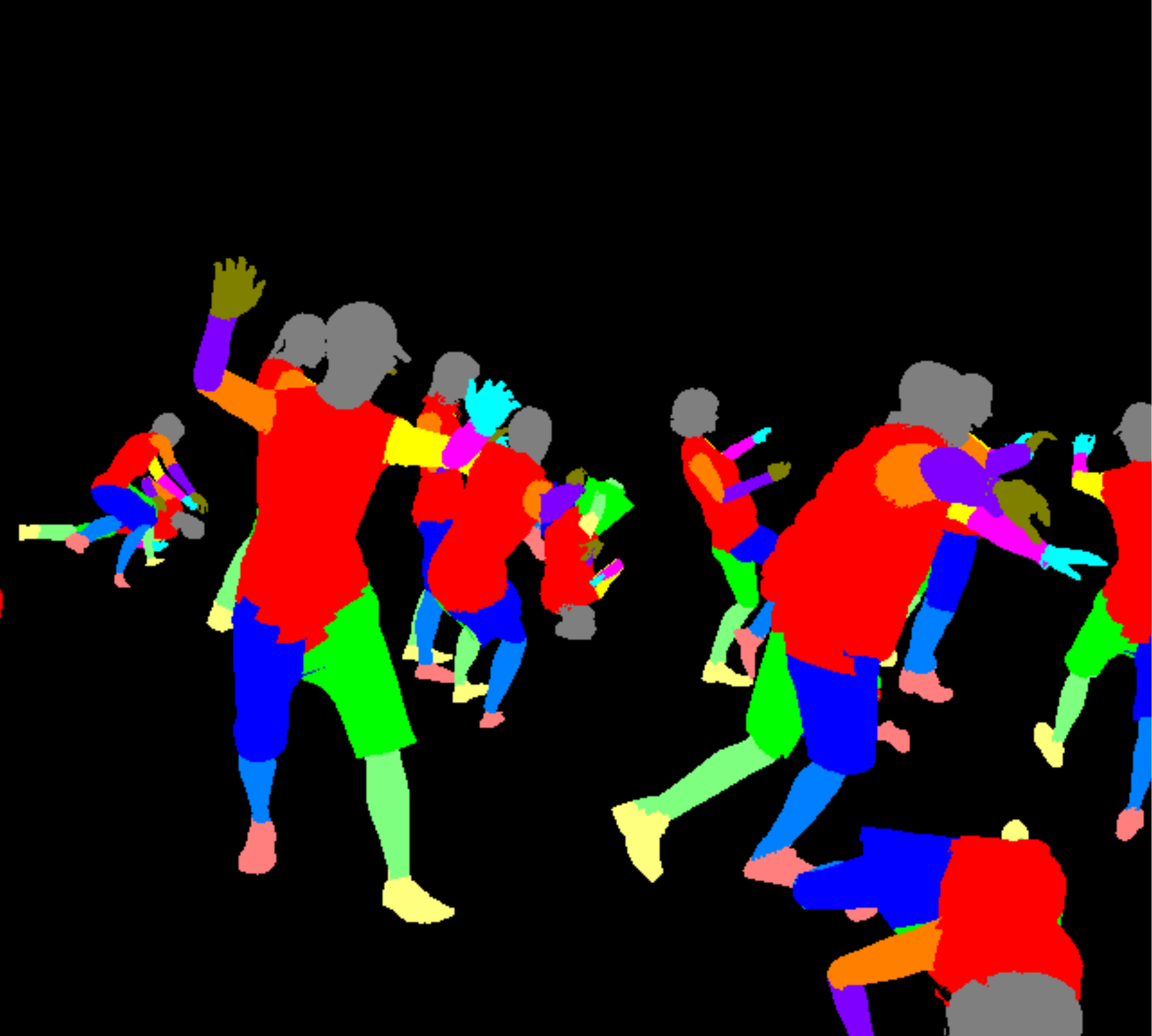}
 	&\includegraphics[width=.14\textwidth]{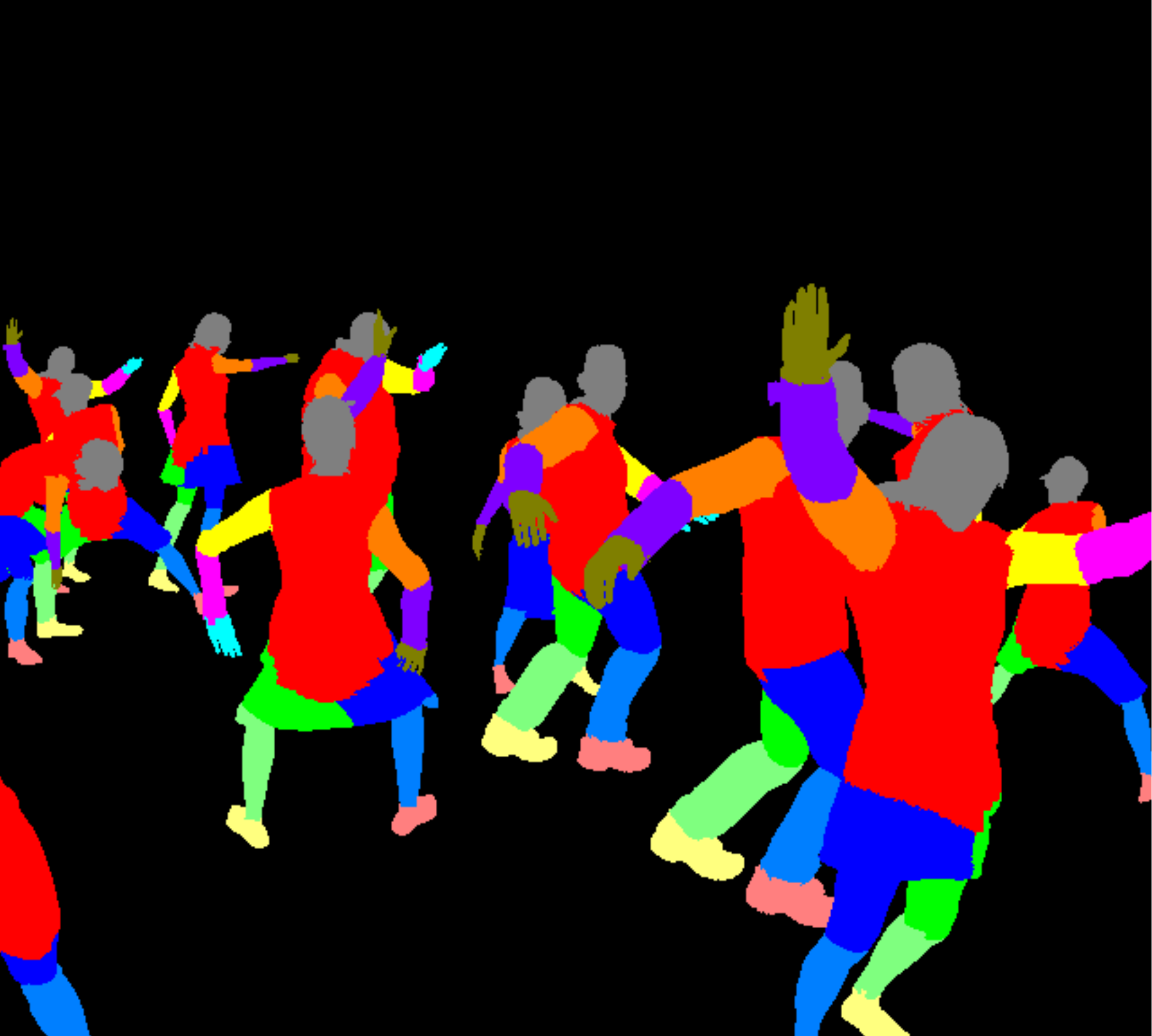}
 	&\includegraphics[width=.14\textwidth]{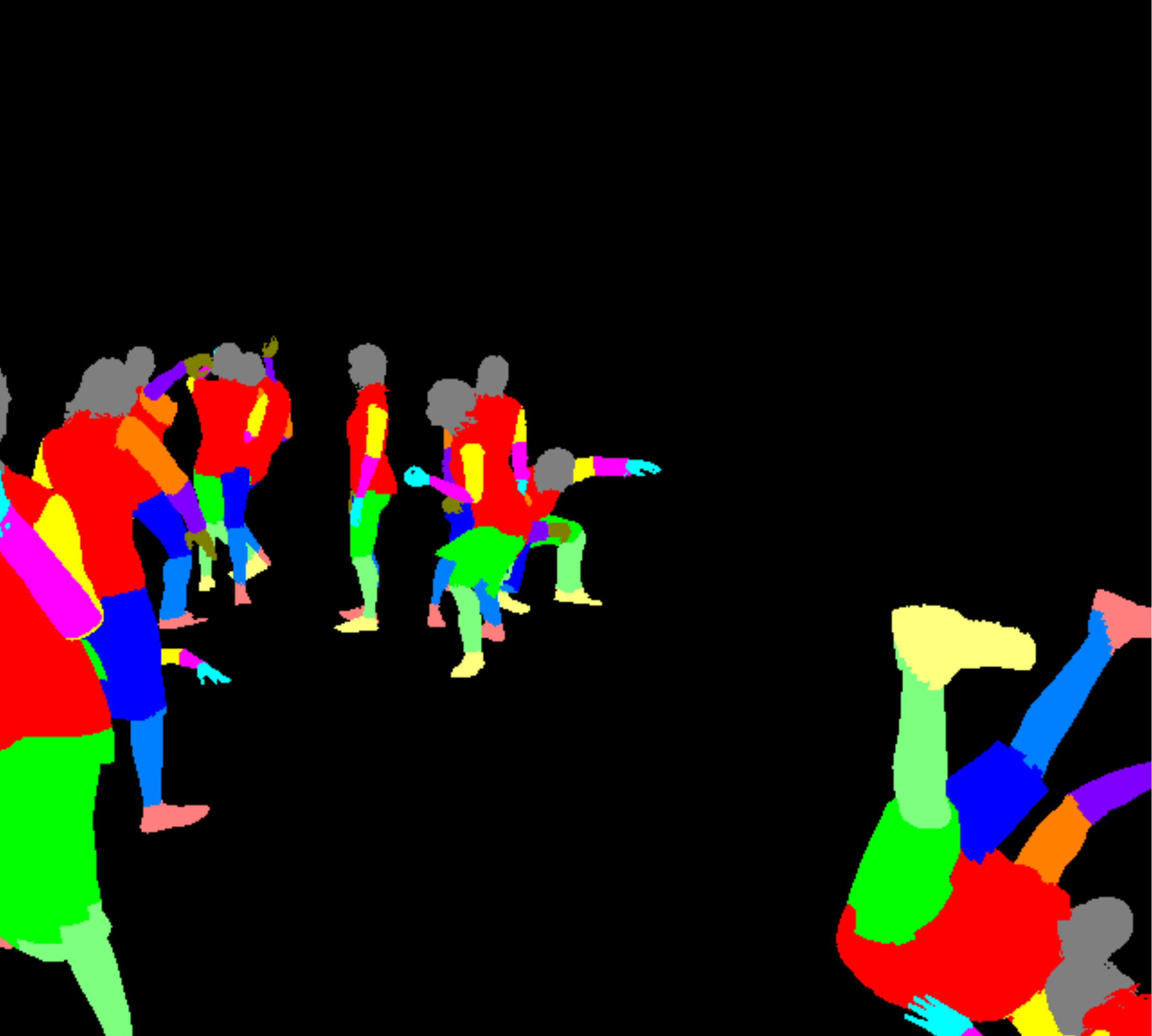}
 	&\includegraphics[width=.14\textwidth]{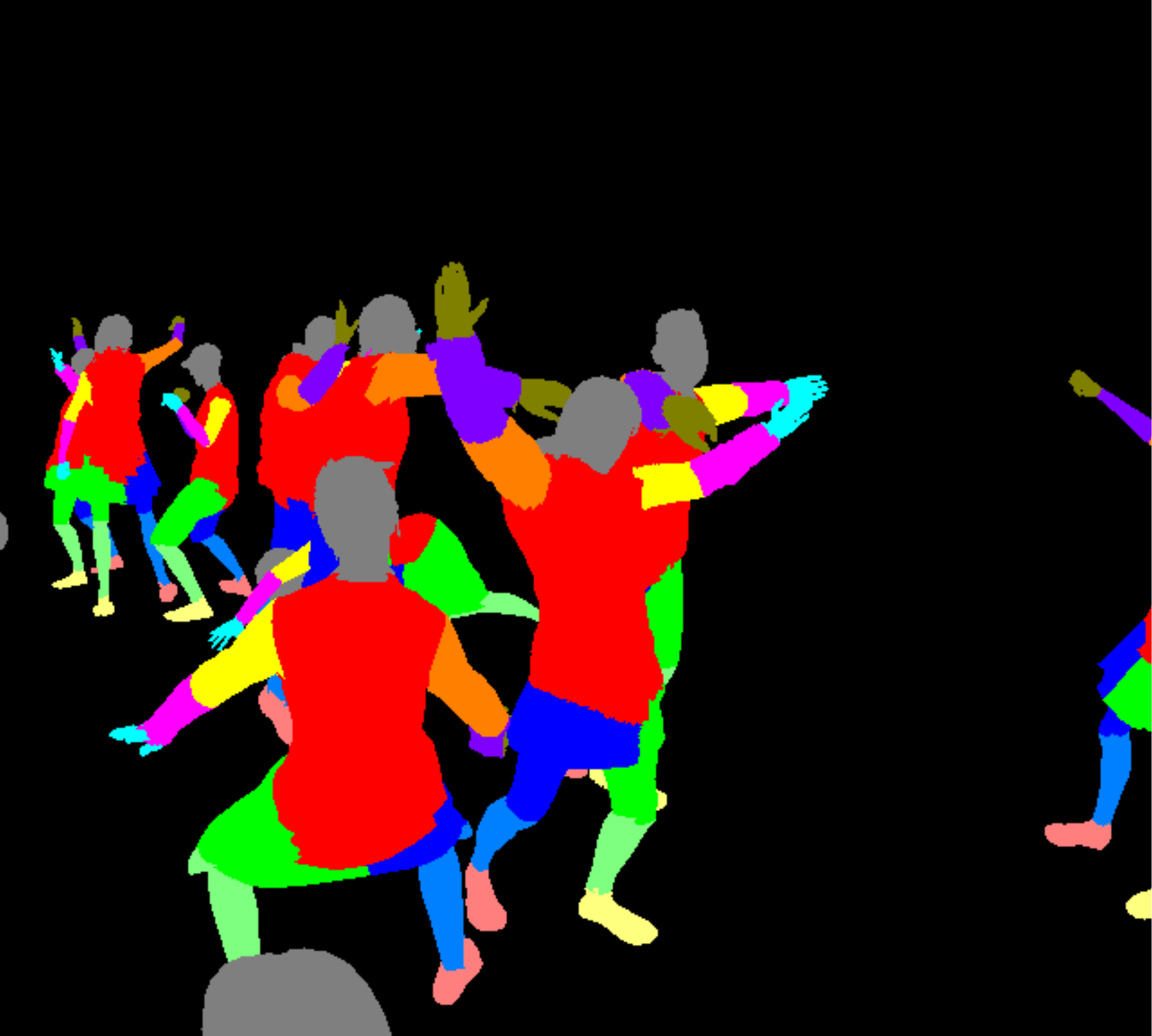}
 	&\includegraphics[width=.14\textwidth]{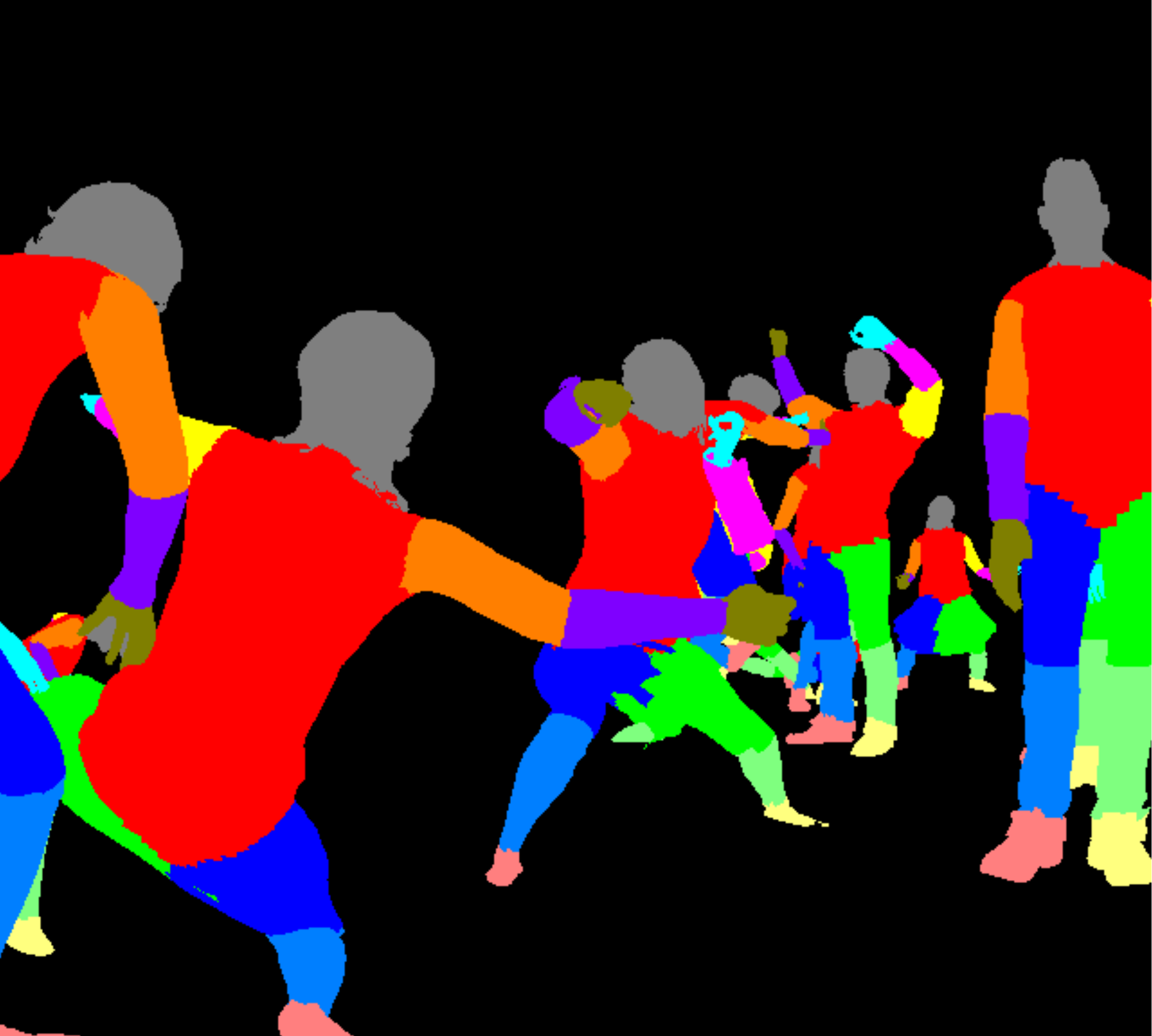}
 	&\includegraphics[width=.14\textwidth]{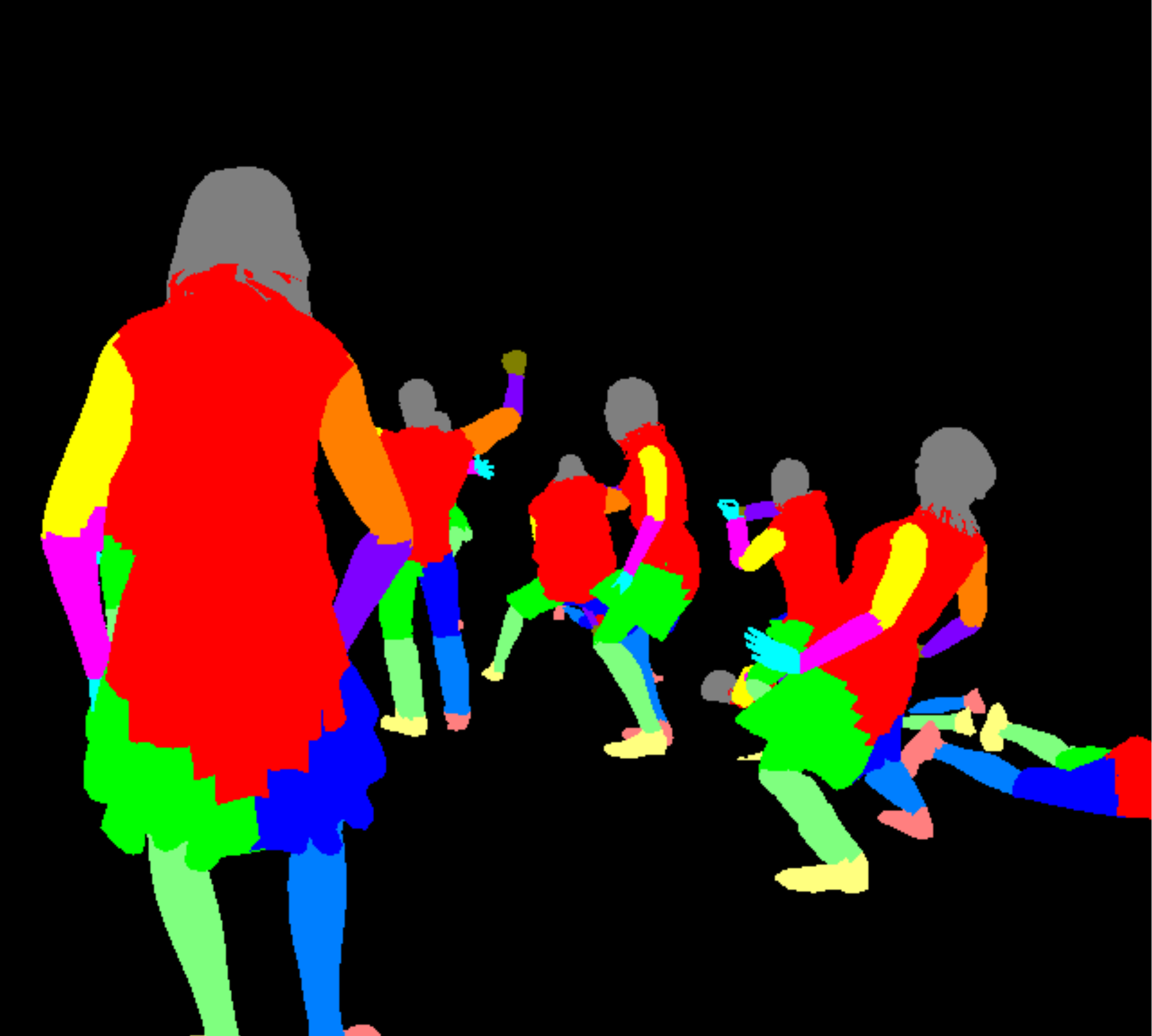}
 	&\includegraphics[width=.14\textwidth]{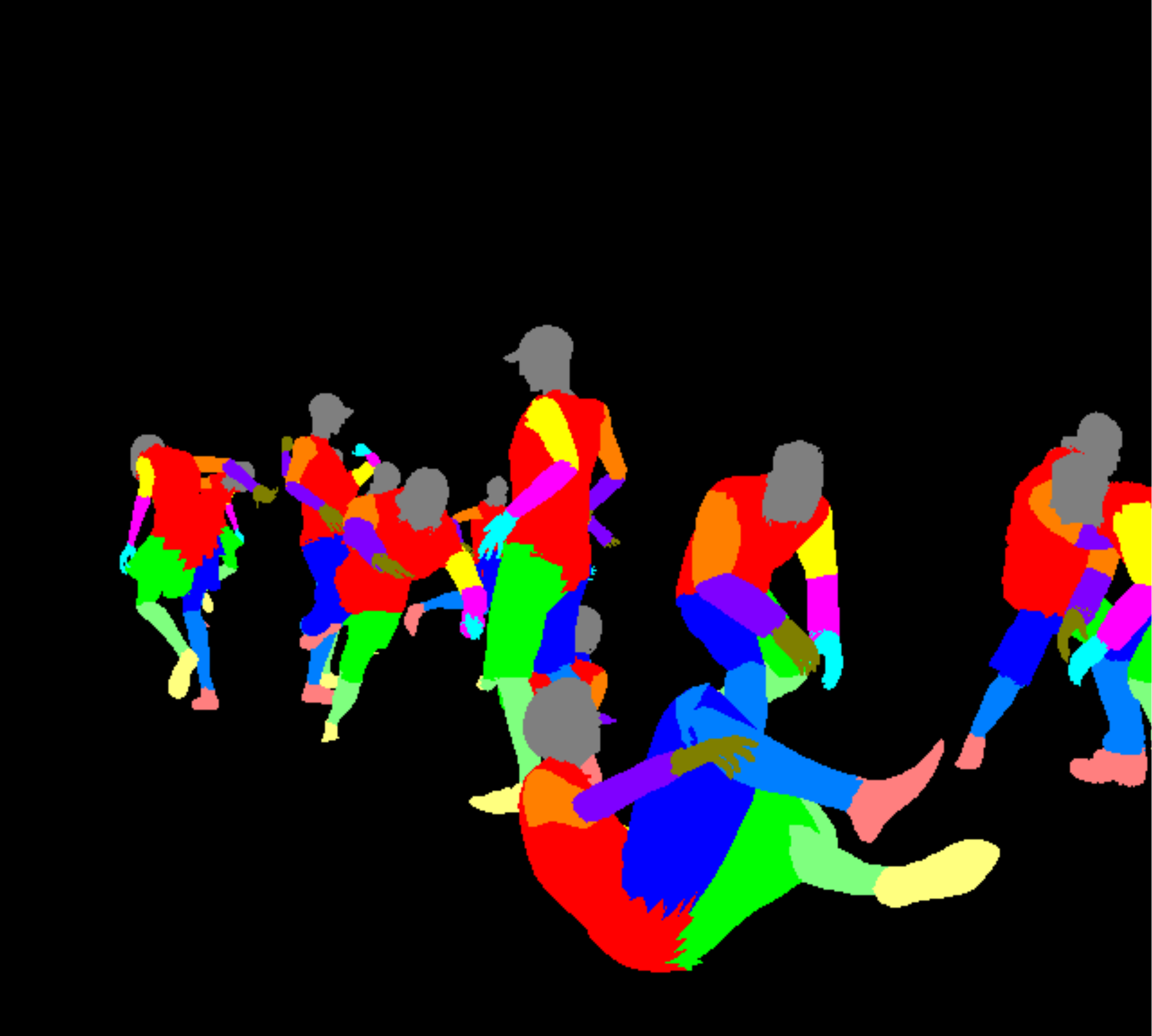}\\
    \end{tabular}
 \caption{Samples of our synthetic data. Our synthetic data contain multiple persons performing various actions in a 3D room. Top row: the synthetic RGB images. Middle row: the synthetic pose labels. Bottom row: the synthetic part labels.}
\label{fig:syn-samples} 
\end{figure*}
\section{Synthetic Data}\label{sec:data}

It is a common belief that high-quality synthetic data should be created as similar as possible to the real-world scenarios. For example, in generating single-person synthetic data~\cite{varol2017learning}, the authors composed their synthetically generated human images with a variety of real world background images. An advantage of our technique is that we reduce the requirement on the photorealism of the synthetic data generation. In particular, we use a simple empty room as the background for all of our synthetic data. The reason why our technique works well even with such a simple synthetic background is that our technique learns about the background from the real data.

We have $20$ 3D human models with different body shapes and clothing. These avatars are randomly placed at different positions in the virtual room, and they are animated to perform a variety of actions such as walking, jumping, crawling, etc. To create realistic human motions, we retarget the motion capture data from CMU MoCap database~\cite{cmu-mocap} to the avatars. We use a ray-tracing based rendering engine~\cite{arnoldrenderer,maya} to render the scene.

Multiple virtual cameras are set up at different positions in the environment to capture the scene from a variety of viewpoints. Figure~\ref{fig:syn_design} shows the layout of our simulation environment. The virtual camera model we used is a pinhole camera with a $90$ degree FoV. The exposure of the camera is 1/30-th of a second. The focal length is $35$ mm.

Figure~\ref{fig:syn-samples} shows the examples of our synthetic data and the ground truths. Our graphics simulator generates different types of per-pixel ground truth labels for the animations.
Following the common definitions of body parts and human pose~\cite{Guler2018DensePose,lin2014microsoft}, we generate $14$ categories of body part ground truth labels, and $17$ types of keypoint ground truth labels. It is worth noting that the labels for the synthetic data can be freely extended depending on user preferences, and are more flexible than those in the conventional real datasets. 
For example, as shown in Section~\ref{sec:novelkey}, we generate a new set of keypoints including hands and feet from synthetic data thus allowing our model to predict new keypoints.

Another advantage of the graphics simulation is that we can easily generate large amount of data. In this work, we generate a total of $17,211$ frames and their corresponding ground truths for model training. 
\section{Method}\label{sec:method}

Given a set of synthetic data with human part segmentation labels, we would like to learn a function that performs human part segmentation on real world data. If we directly train a neural network with synthetic data labels, it does not generalize well to real data due to the reality gap. Unlike existing methods~\cite{ren2017cross} that try to transform the synthetic data to real data domain to make them look similar to each other, we use a complementary learning strategy that effectively leverages the rich variation of the real data and the part segmentation labels of the synthetic data. To make sure the synthetic data and real data are aligned in a common latent space, we use an auxiliary task, pose estimation, to bridge the two domains. In summary, our training data consist of part segmentation labels and pose labels from synthetic data, and pose labels from real data. We learn a part segmentation function without any part segmentation labels from real data.

\begin{figure}[t]
	\centering
\includegraphics[width=.99\columnwidth]{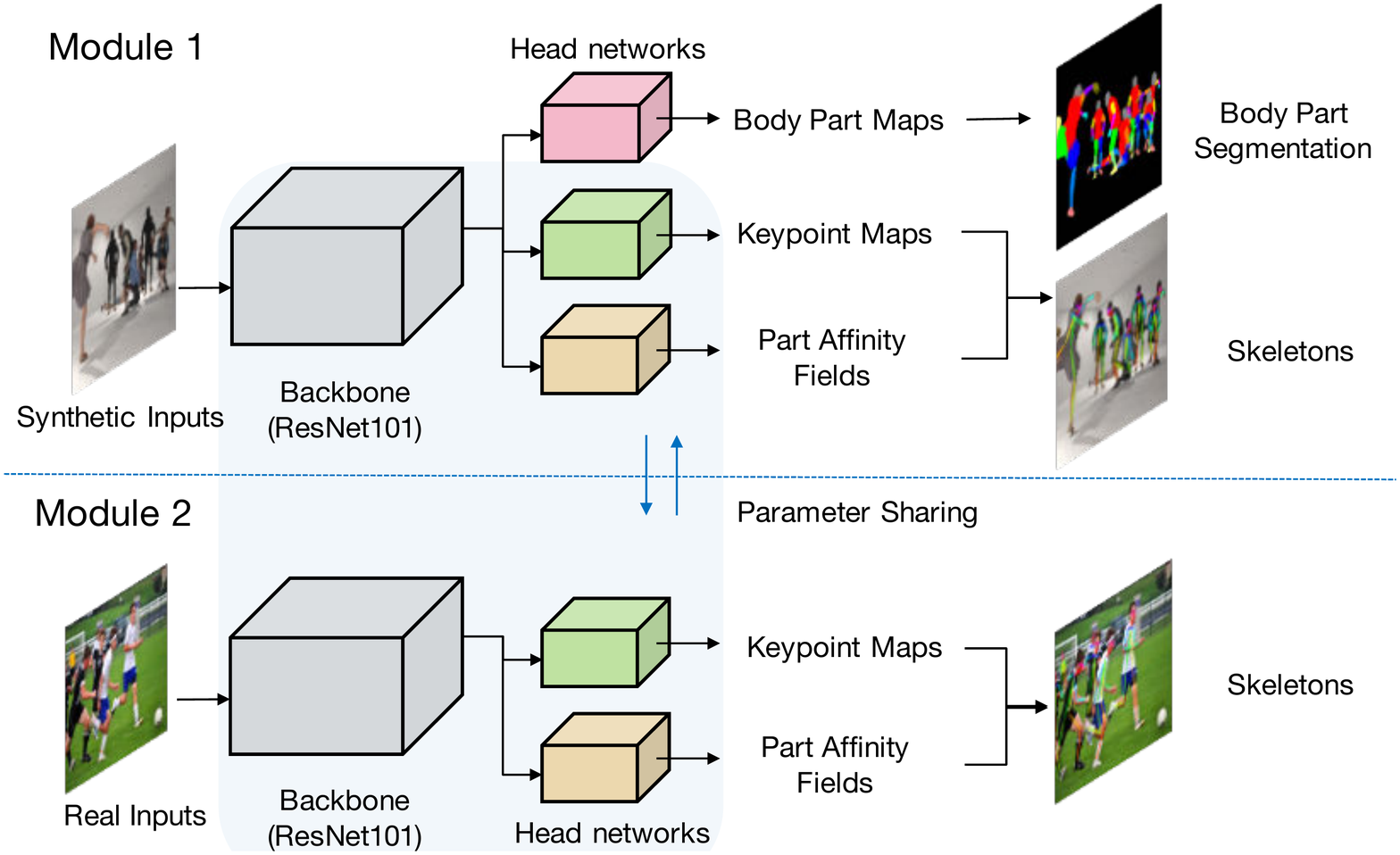}
	\caption{An overview of the proposed framework. Our framework consists of two main components. The first is the synthetic input training to learn body parts and human poses \textcolor{black}{in} the synthetic domain. In the second component for real input training, we share the network parameters of the backbone, keypoint map head, and part affinity field head with the first component. During learning, we train our network using two modules within a mini-batch, and optimize the network using back-propagation. }
	\label{fig:network}
\end{figure}

\subsection{Learning objective}
Given a real dataset with pose labels $D_r^{pose}$, a synthetic dataset with pose labels $D_s^{pose}$, and a synthetic dataset with part segmentation labels $D_s^{part}$, we formulate the cross-domain complementary learning (CDCL) as the following optimization problem:
\begin{equation}
\label{eqn:abstract}
L = \alpha L_{pose}(D_r^{pose}) +  \beta L_{pose}(D_s^{pose})+ \gamma L_{part}(D_s^{part}),
\end{equation}
where $L_{pose}$ is the loss function for pose estimation, and $L_{part}$ is the loss function for part segmentation. The first two terms together form the objective function for learning the auxiliary task of pose estimation from both real and synthetic data. The third term learns part segmentation from synthetic data. Note that $\alpha$, $\beta$, $\gamma$ are the hyperparameters for balancing the losses among the three terms.

Human pose estimation aims at detecting human skeletons in a given image. Previous study~\cite{cao2017realtime} proposed to detect the joint locations (i.e. keypoints) and the associations between the joints (i.e. Part Affinity Fields). After that, human skeletons are reconstructed with a greedy algorithm.
Following the common definition of pose labels~\cite{cao2017realtime,lin2014microsoft}, we use the annotations of keypoints and Part Affinity Fields (PAFs)~\cite{cao2017realtime} for learning pose estimation. In particular, let $D_r^{pose}=\{I_{r}^i, {K}_{r}^i, {P}_{r}^i\}_{i=1}^{M}$, where $M$ is the total number of real images, $I_{r} \in R^{w\times h \times 3}$ denotes a real RGB image, ${K}_{r} \in R^{w\times h \times J}$ denotes a real keypoint ground truth, which has $J$ different maps, one per keypoint, ${P}_{r} \in R^{w\times h \times C}$ denotes a real part affinity ground truth, which has $C$ affinity vector fields. Also, we have a synthetic dataset with pose labels $D_s^{pose}=\{I_{s}^i, {K}_{s}^i, {P}_{s}^i\}_{i=1}^{N}$, where $N$ is the total number of images in the synthetic data. Furthermore, we have a synthetic dataset with part segmentation labels $D_s^{part}=\{I_{s}^i, {B}_{s}^i\}_{i=1}^{N}$, where ${B}_{s} \in R^{w\times h \times Z}$ is the synthetic body part segmentation ground truth and $Z$ is the total number of body part categories.
Note that it is convenient to assume $D_s^{pose}$ and $D_s^{part}$ share the same set of images. In this work, we use COCO Keypoint dataset~\cite{lin2014microsoft} as $D_r^{pose}$.

In the following, we omit the subscript \textit{r} and \textit{s} and use $D^{pose}$ to represent either real or synthetic data. The loss function we use for learning pose estimation is $L_{pose}(D^{pose}) = L_{kpts}(I, K, \hat{K}) + L_{paf}(I, P, \hat{P})$ where $L_{kpts}(\cdot)$ and $L_{paf}(\cdot)$ are the Euclidean loss functions minimizing the differences between the predictions and the ground truths, and they are defined below:
\begin{equation}
\label{eqn:l2}
L_{kpts}(I, K, \hat{K}) = \sum_{j=1}^{J} \sum_{\theta} \mathcal{M}(\theta)||K(\theta) - \hat{K}(\theta)||_{2}^{2},
\end{equation}
\begin{equation}
\label{eqn:l3}
L_{paf}(I, P, \hat{P}) = \sum_{c=1}^{C} \sum_{\theta} \mathcal{M}(\theta)||P(\theta) - \hat{P}(\theta)||_{2}^{2},
\end{equation}
where $\hat{K}$ and $\hat{P}$ denote the predicted keypoint confidence map and the predicted part affinity field, respectively, and $K$ and $P$ denote the ground truths. $\mathcal{M}$ is a binary mask, where $\mathcal{M}(\theta) = 0$ if the ground truth is missing at the location $\theta$ of the image. The mask is used to avoid penalizing the correct predictions as discussed in~\cite{cao2017realtime}.

The loss function of learning part segmentation is denoted as $L_{part}(D^{part}) = L_{part}(I, B, \hat{B})$, which is defined as the categorical cross entropy loss for classifying pixels to different human parts, that is:
\begin{equation}
\label{eqn:l1}
L_{part}(I, B, \hat{B}) = -\sum_{z=1}^{Z} \sum_{\theta} \mathcal{M}(\theta) B(\theta)\log(\hat{B}(\theta)),
\end{equation}
where $\hat{B}$ denotes the predicted body part maps, $B$ denotes the synthetic part segmentation ground truths.

In summary, the overall objective function is
\begin{equation}
\begin{aligned}\label{eqn:overall-loss}
L
& = \alpha \Big( L_{kpts}(I_r, K_r, \hat{K}) + L_{paf}(I_r, P_r, \hat{P})\Big)\\
& + \beta \Big( L_{kpts}(I_s, K_s, \hat{K}) + L_{paf}(I_s, P_s, \hat{P}) \Big)\\
& + \gamma L_{part}(I_s, B_s, \hat{B}).
\end{aligned}
\end{equation}
\subsection{Network architecture}

Figure~\ref{fig:network} illustrates the proposed network. Our network takes an image of arbitrary size as input, and predicts three different outputs including (1) a set of body part segmentation maps $\hat{B}$, (2) a set of confidence keypoint maps $\hat{K}$, and (3) a set of Part Affinity Fields (PAFs) $\hat{P}$~\cite{cao2017realtime}. For clarity, we describe our network in two components: \textit{backbone} and \textit{head} networks. 
\subsubsection{\textbf{Backbone network}}
In this paper, all the results are obtained by using ResNet101~\cite{he2016deep} with pyramid connections~\cite{lin2017feature,chen2017cascaded} as our backbone network. We denote
the output feature maps of the residual blocks in ResNet$101$ as $\{ C_1, C_2, C_3, C_4, C_5\}$ for $conv1$, $conv2$, $conv3$, $conv4$, and $conv5$, respectively. Following~\cite{chen2017cascaded}, we normalize the size of the feature maps $\{ C_1 - C_5\}$ to a fixed size $\{\tilde{C}_1-\tilde{C}_5\}$ as the input of the subsequent convolution layers.
We denote $f$ as our backbone network, and the output of our backbone is $F = f(I)$, where $I$ is an input image.
\subsubsection{\textbf{Head network}}
We detect multi-person body parts and human poses in a bottom-up strategy, which is in spirit similar to OpenPose~\cite{cao2017realtime}. Our network predicts three target outputs in parallel, which are $\hat{B}$, $\hat{K}$, and $\hat{P}$. Each head network is a fully convolutional network consisting of $8$ convolution layers with $3\times3$ filters. Note that this is different from prior studies~\cite{cao2017realtime,wei2016convolutional} that have a cascaded multi-stage head architecture. Our head networks do not have such a cascaded design, and can be seen as a single-stage network compared to prior works. Finally, we denote the three head networks as $H_B$, $H_K$, and $H_P$, respectively. The body part segmentation maps $\hat{B}$ are computed by $\hat{B} = H_B(F)$, where $F$ is the output of our backbone. The confidence keypoint maps $\hat{K}$ are computed by $\hat{K} = H_K(F)$, and the Part Affinity Fields~\cite{cao2017realtime} $\hat{P}$ are computed by $\hat{P} = H_P(F)$.

\begin{figure*}[h]
 \centering
 	\includegraphics[height=.161\textwidth]{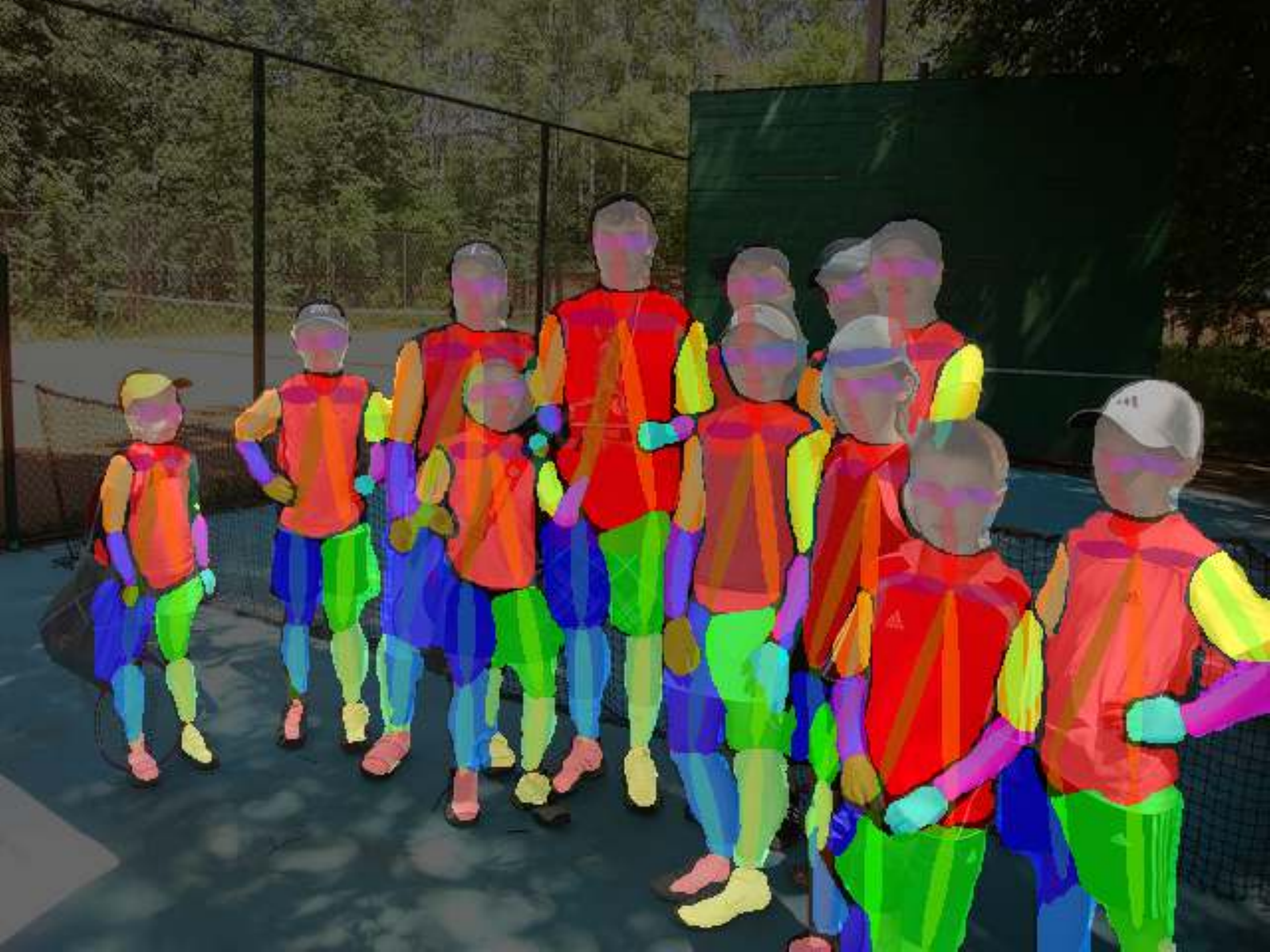}
 	\includegraphics[height=.161\textwidth]{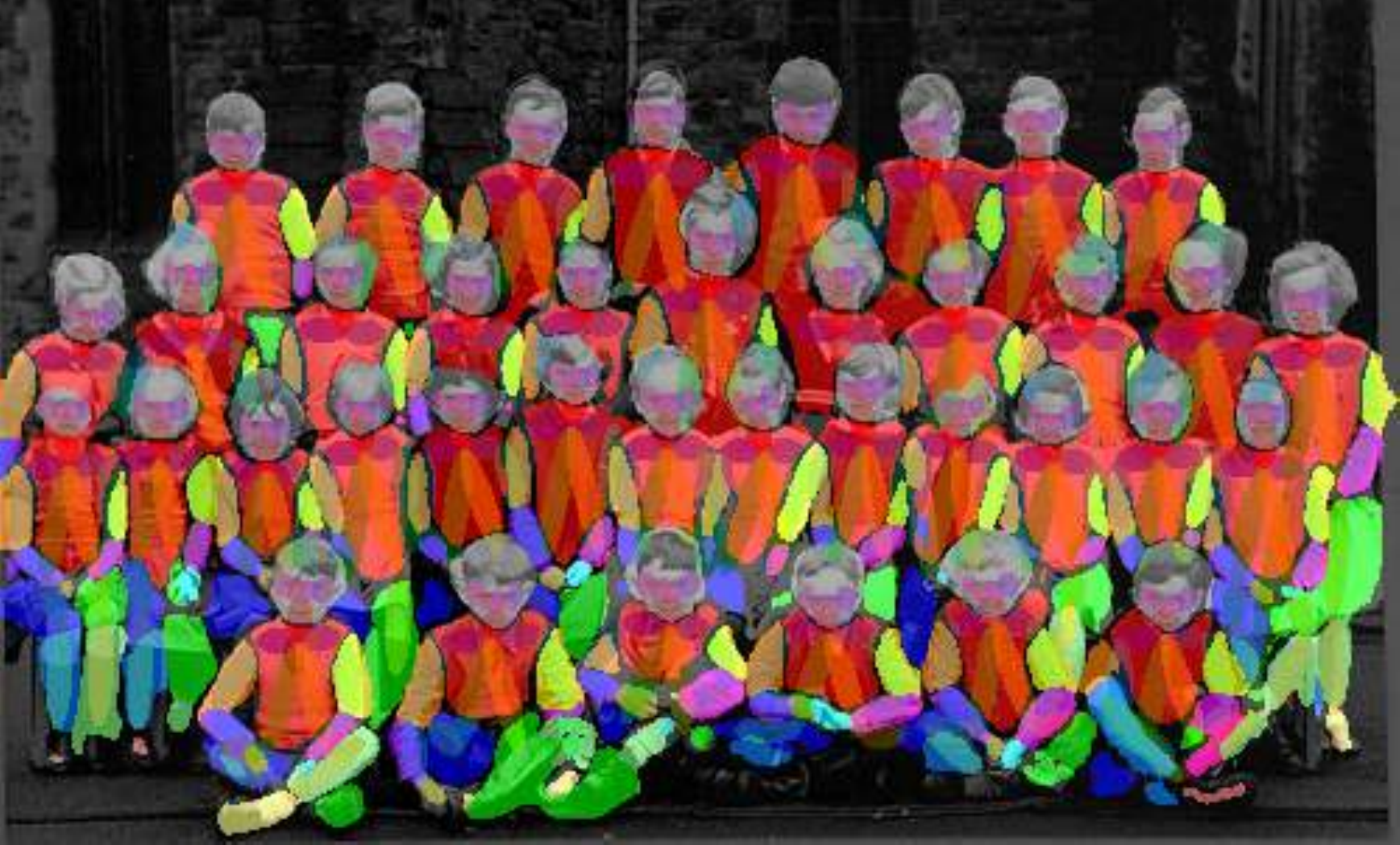}
 	\includegraphics[height=.161\textwidth]{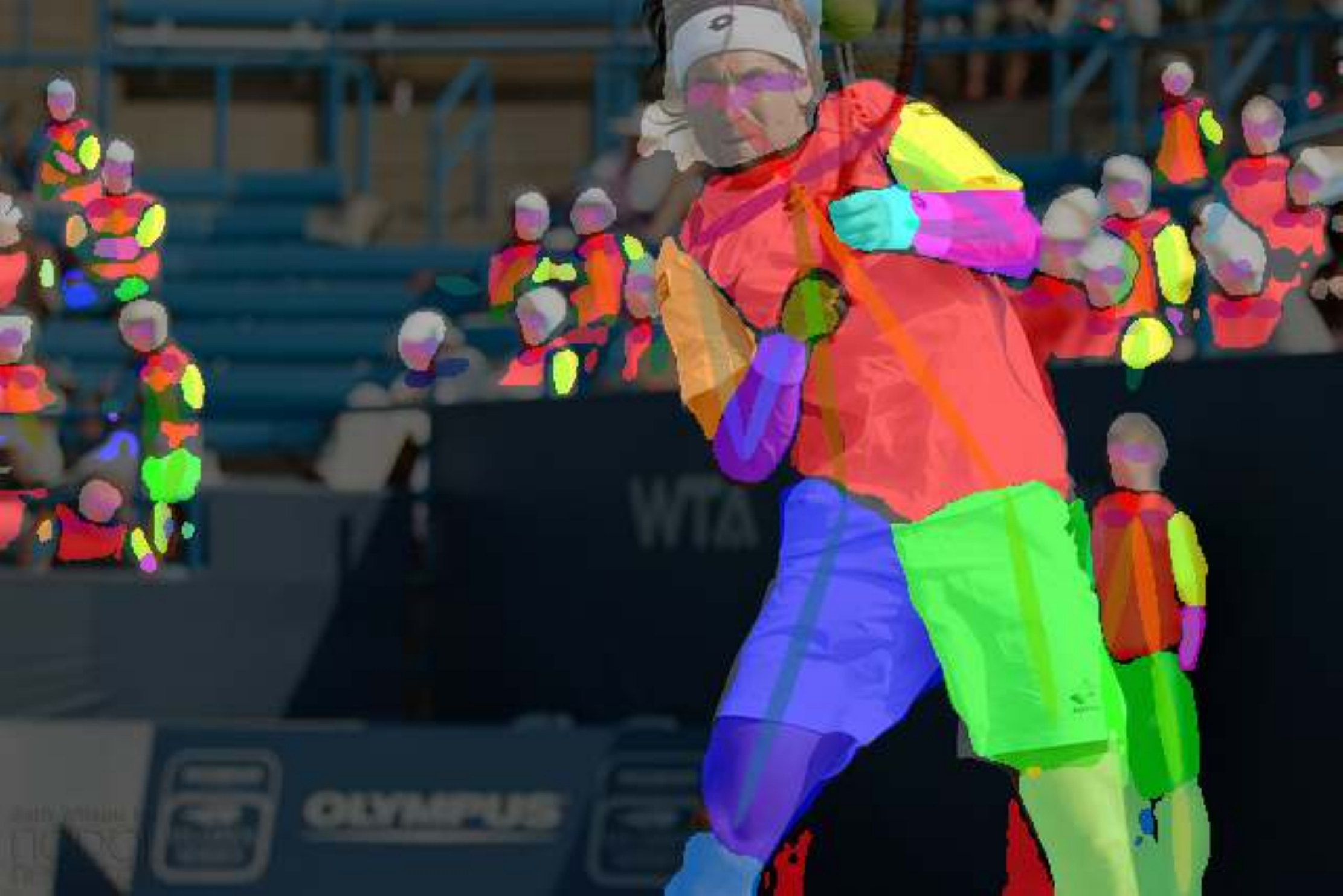}
 	\includegraphics[height=.161\textwidth]{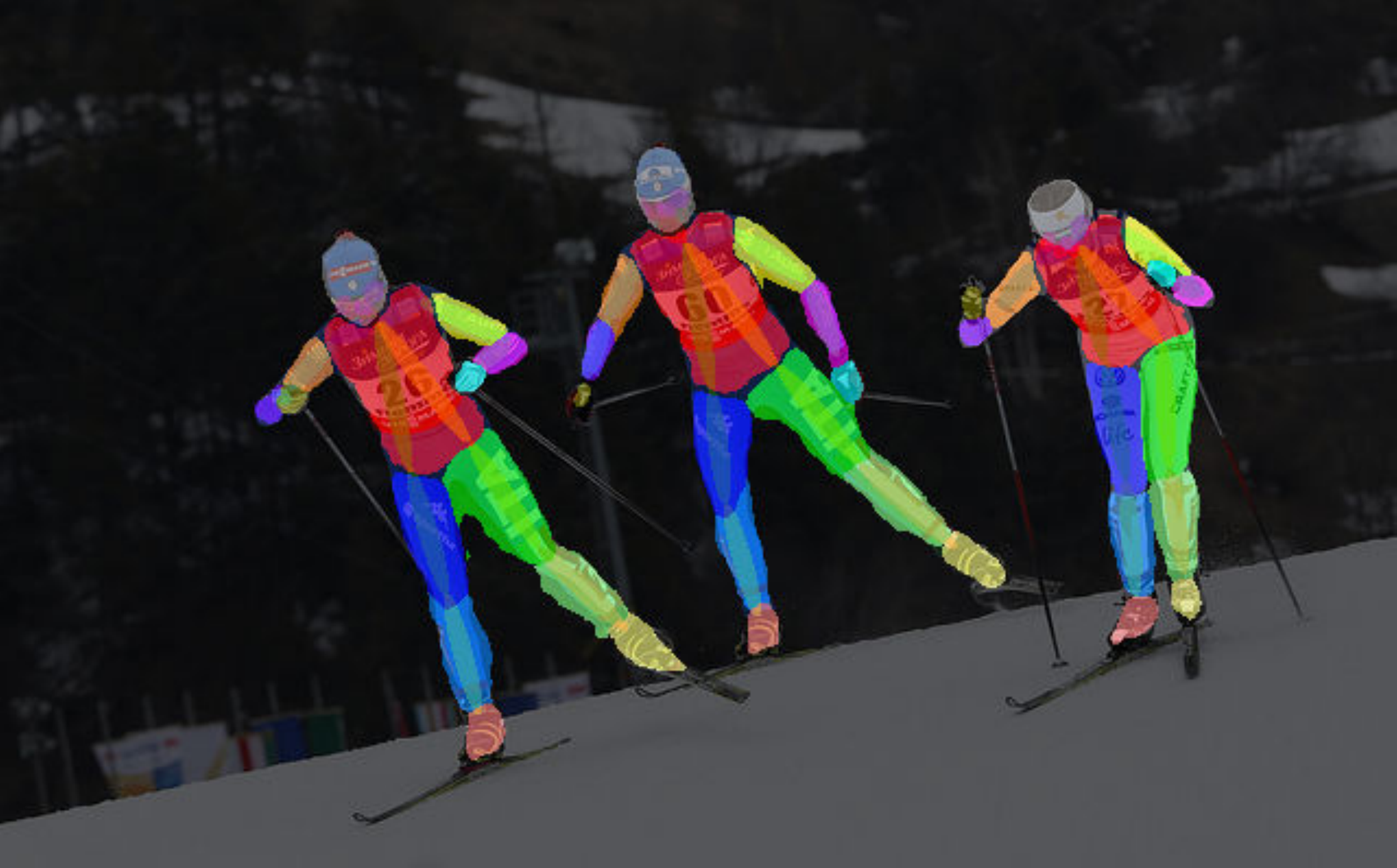}\\
 	\includegraphics[height=.158\textwidth]{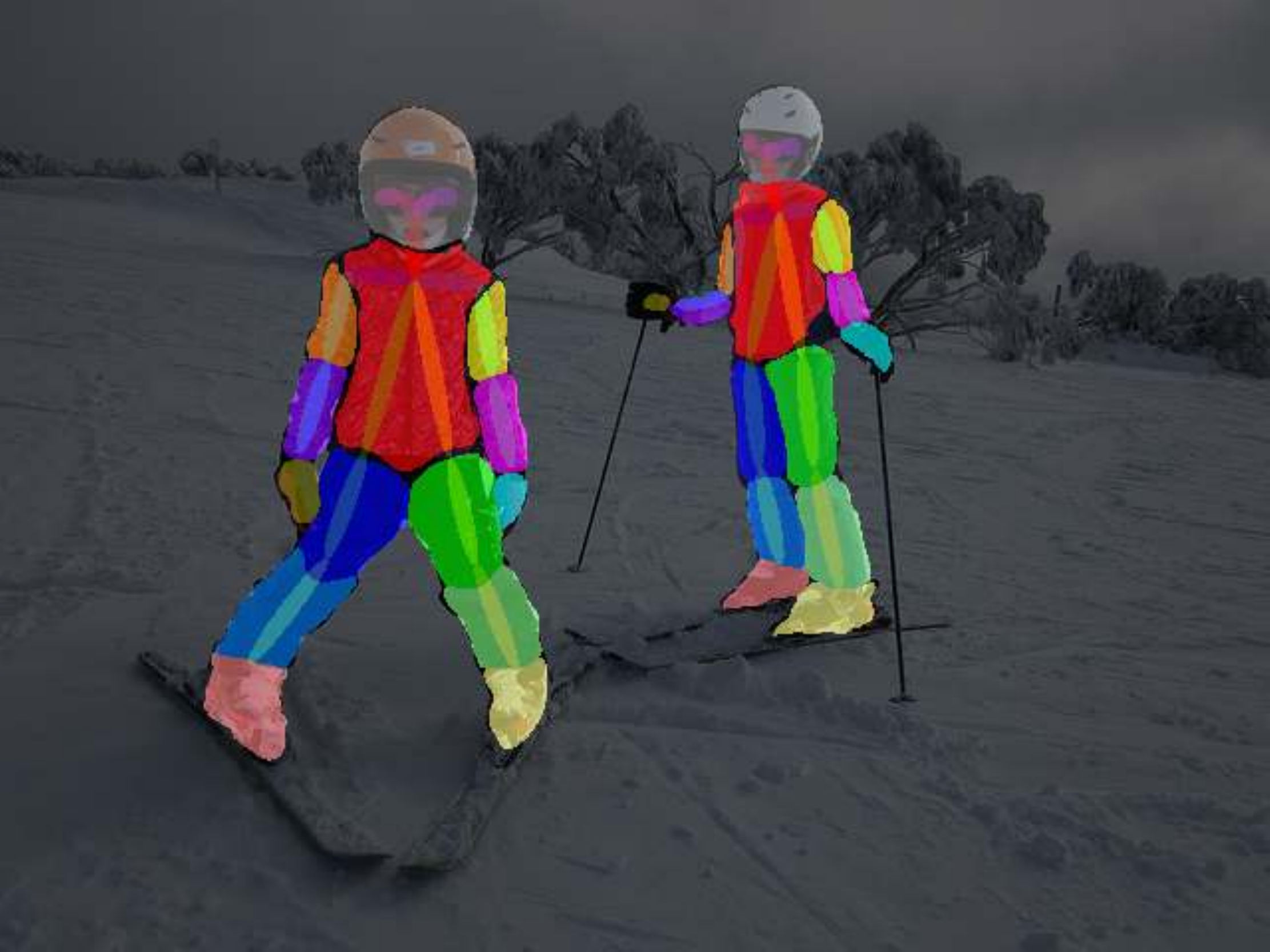}
 	\includegraphics[height=.158\textwidth]{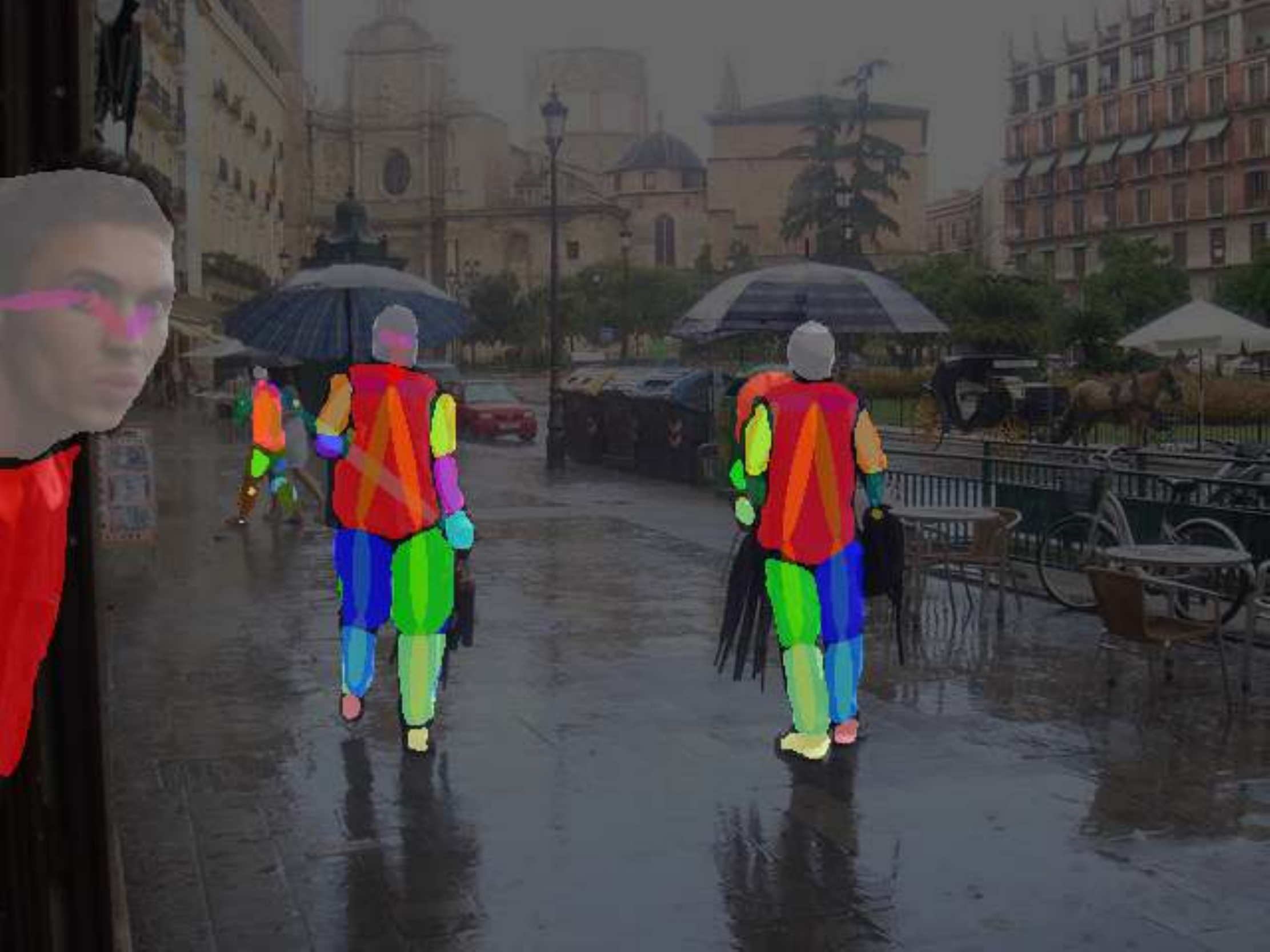}
 	\includegraphics[height=.158\textwidth]{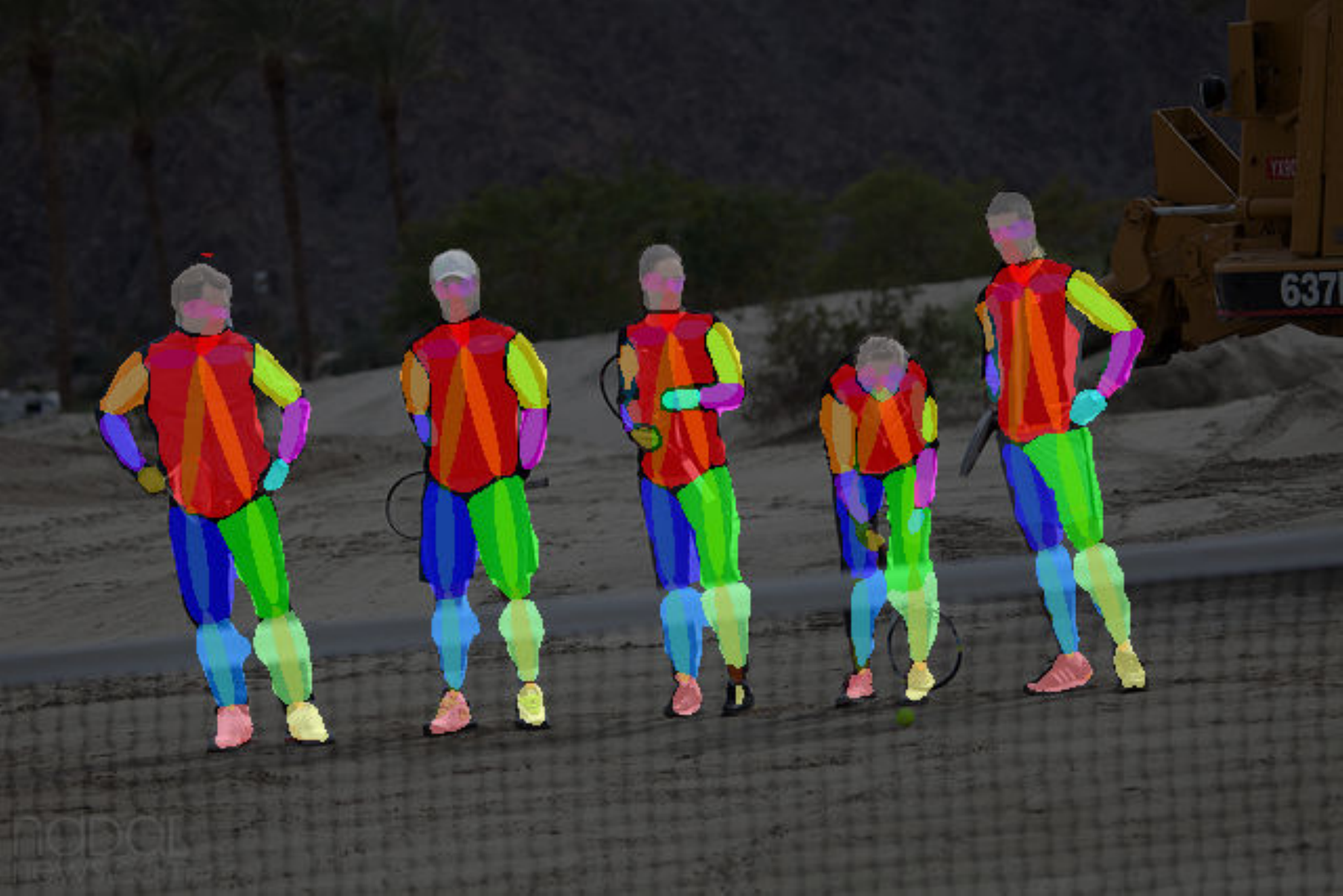}
 	\includegraphics[height=.158\textwidth]{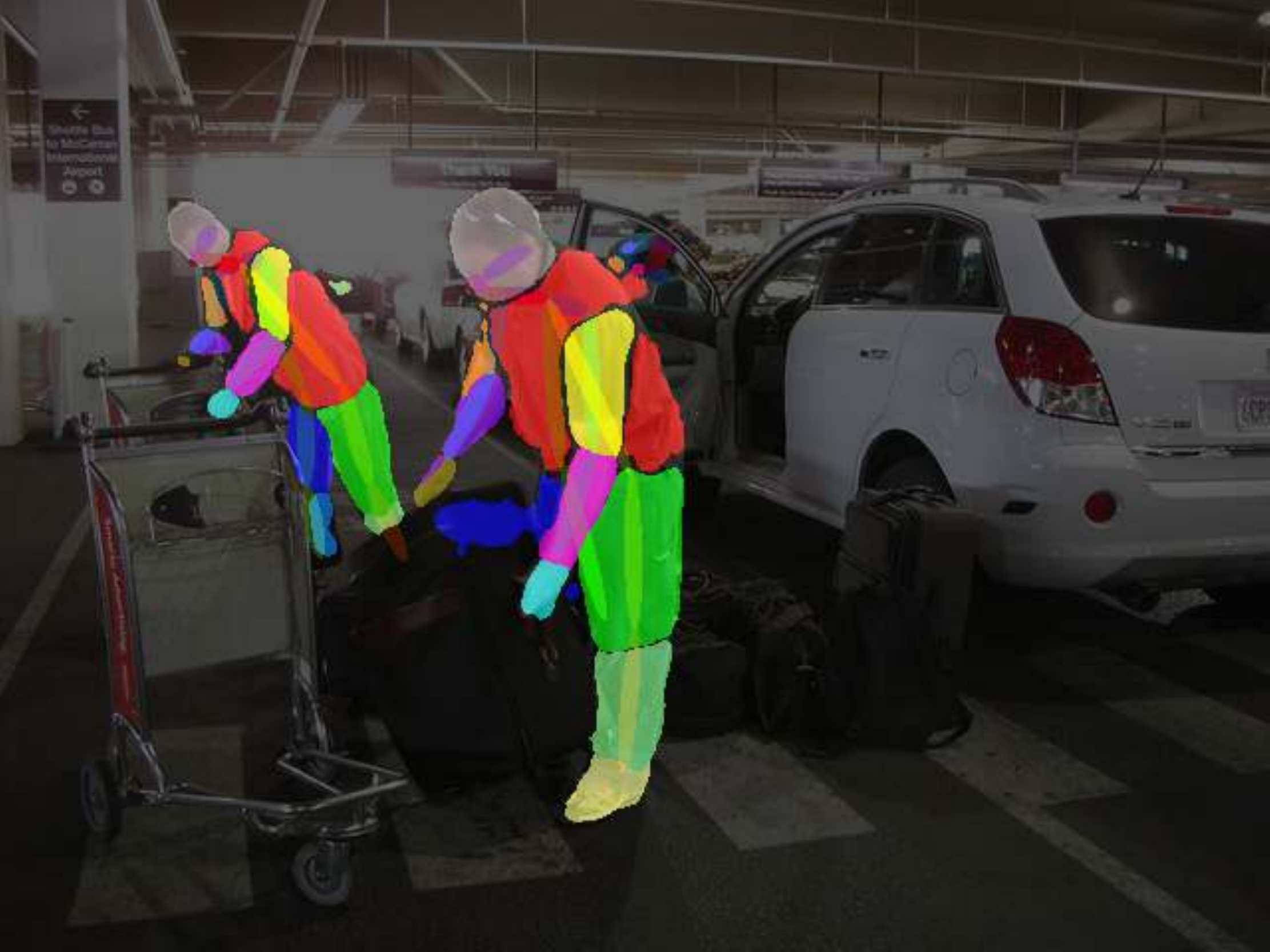}
 	\includegraphics[height=.158\textwidth]{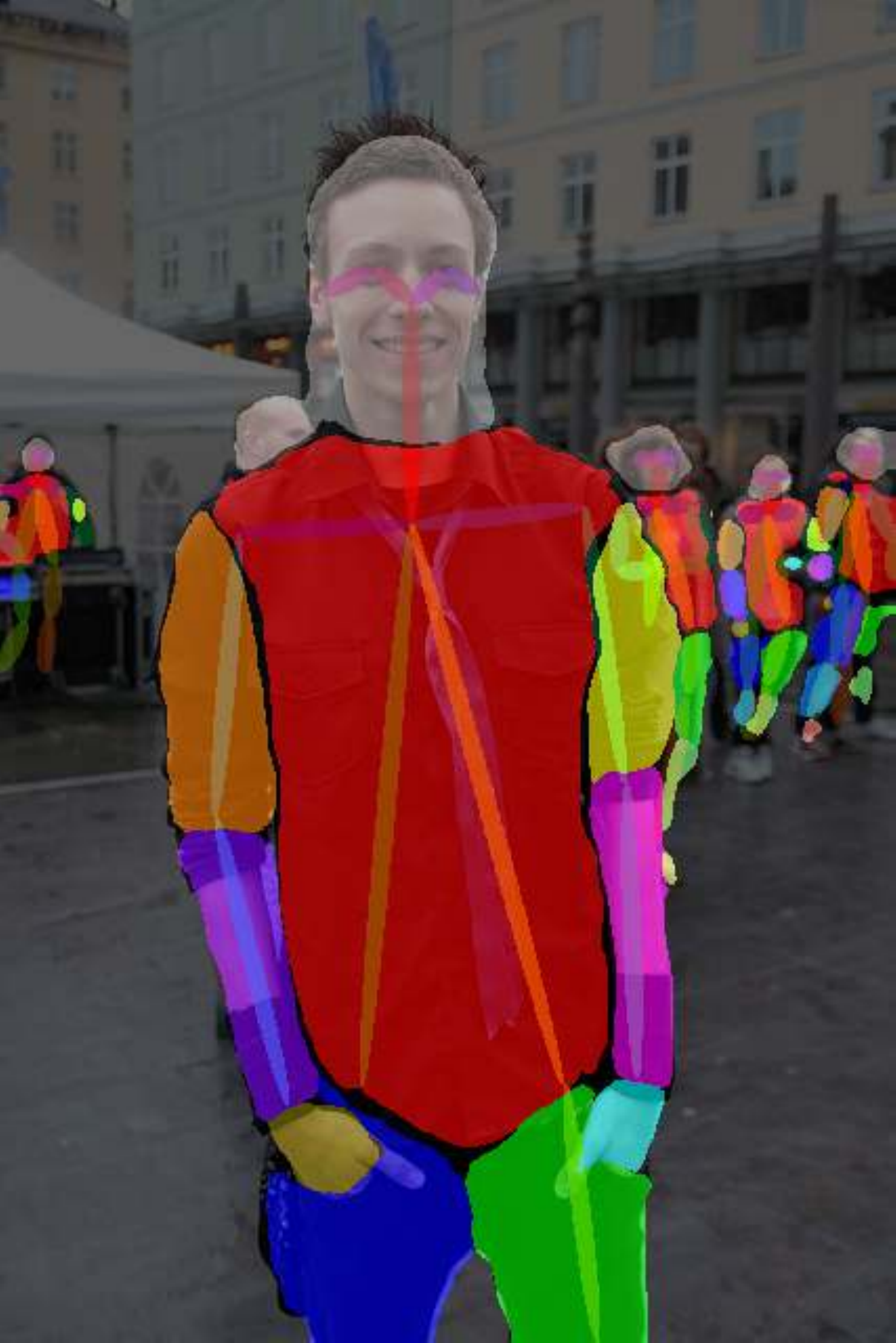}\\
 	\includegraphics[height=.184\textwidth]{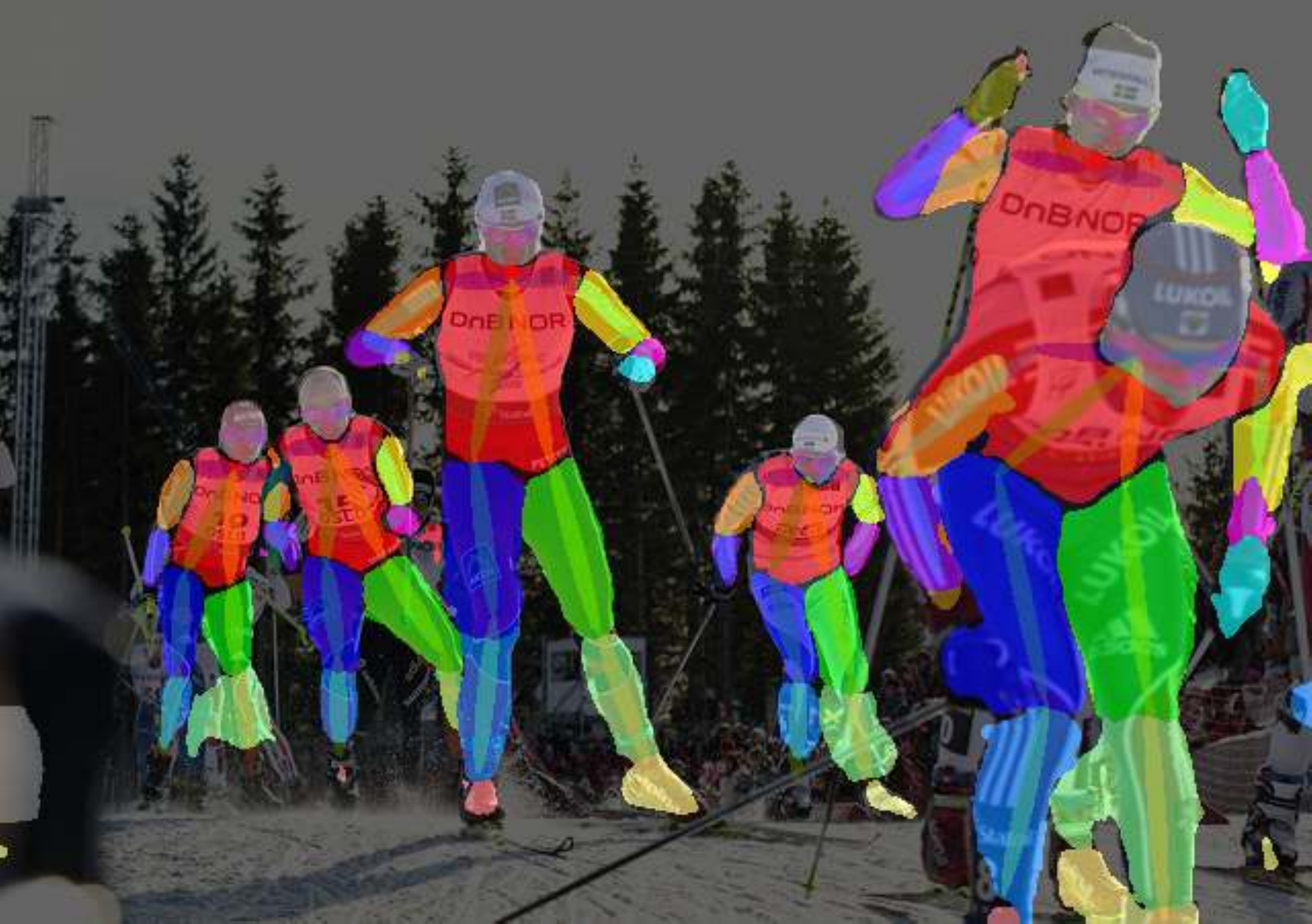}
 	\includegraphics[height=.184\textwidth]{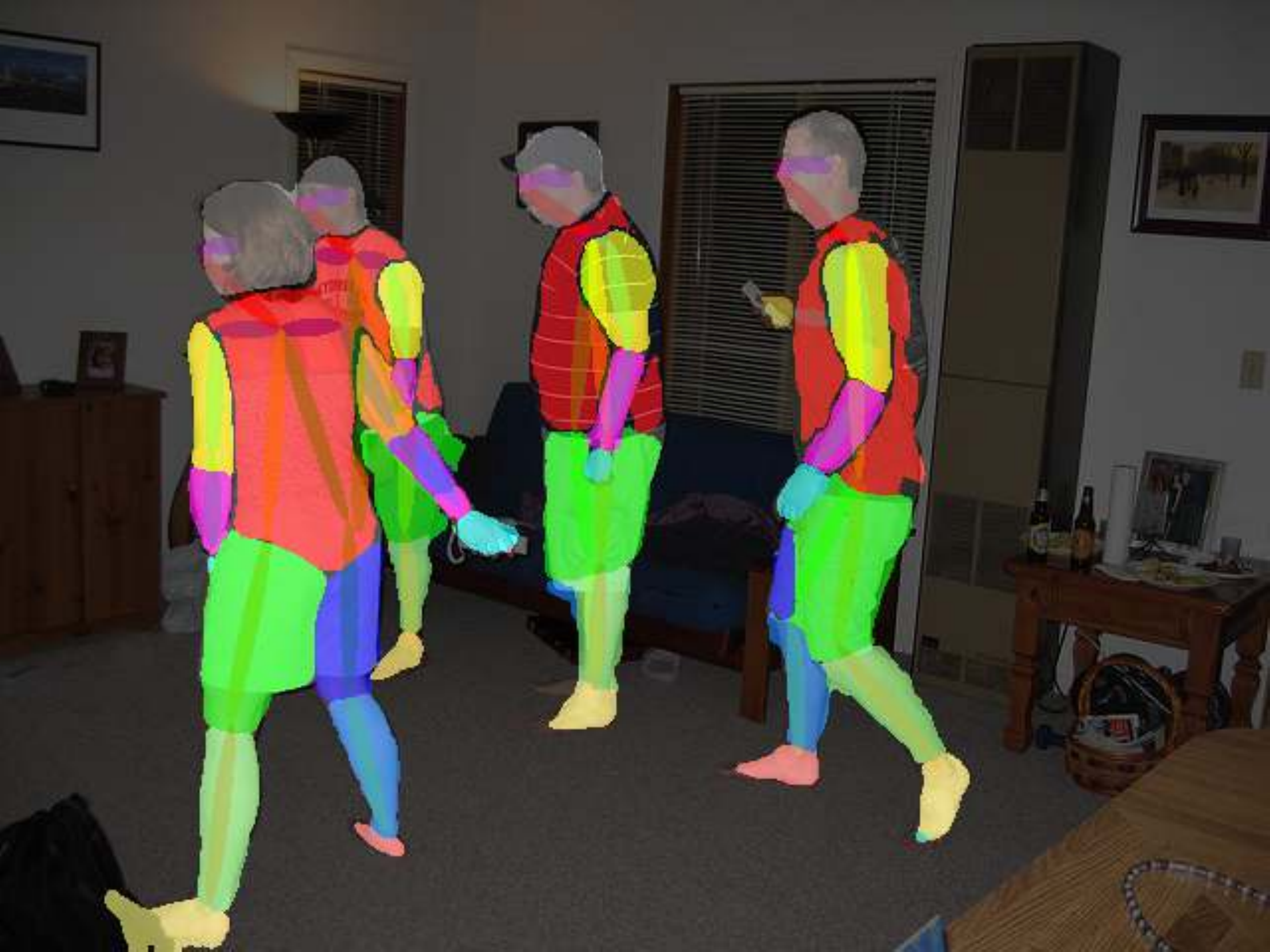}
 	\includegraphics[height=.184\textwidth]{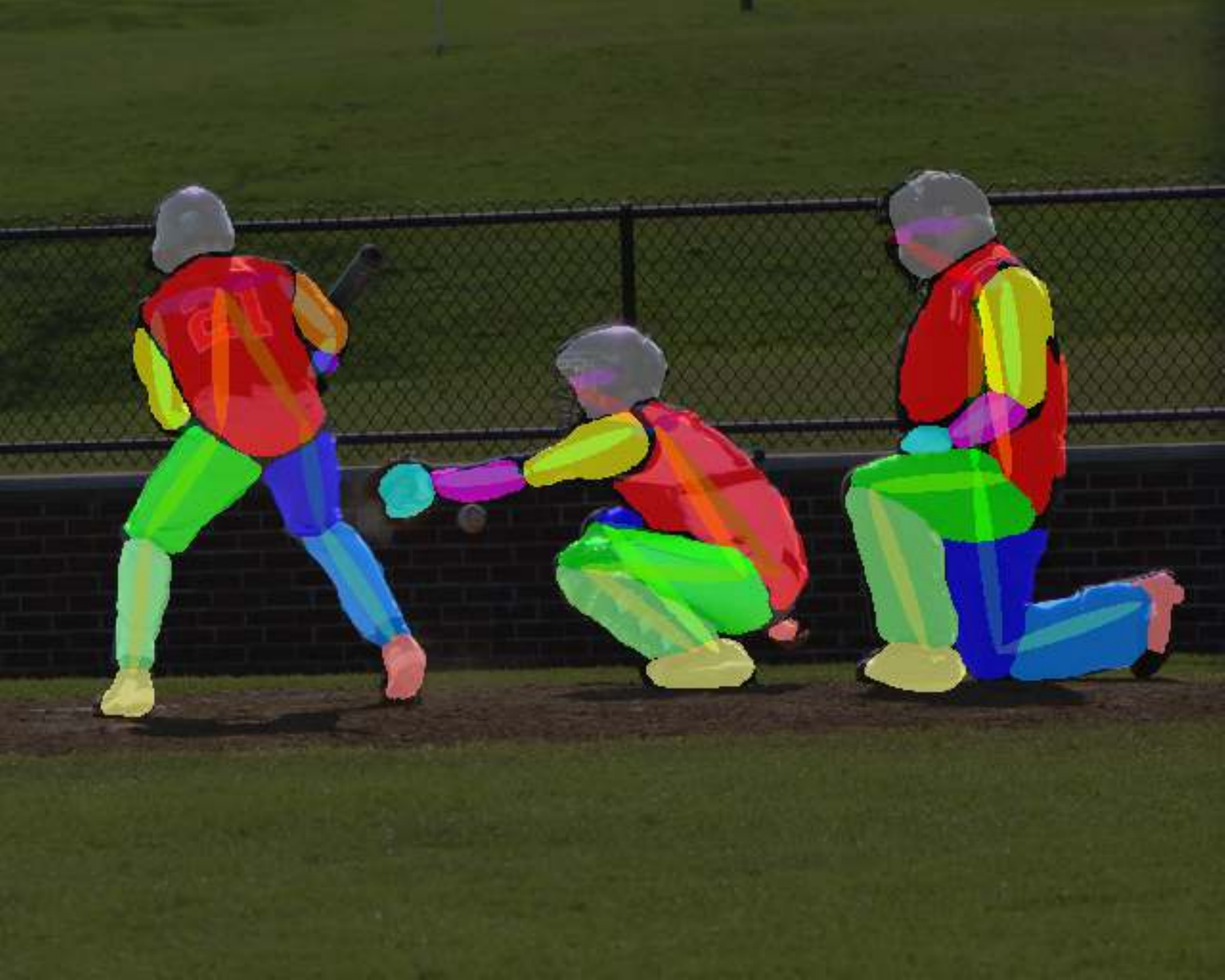}
 	\includegraphics[height=.184\textwidth]{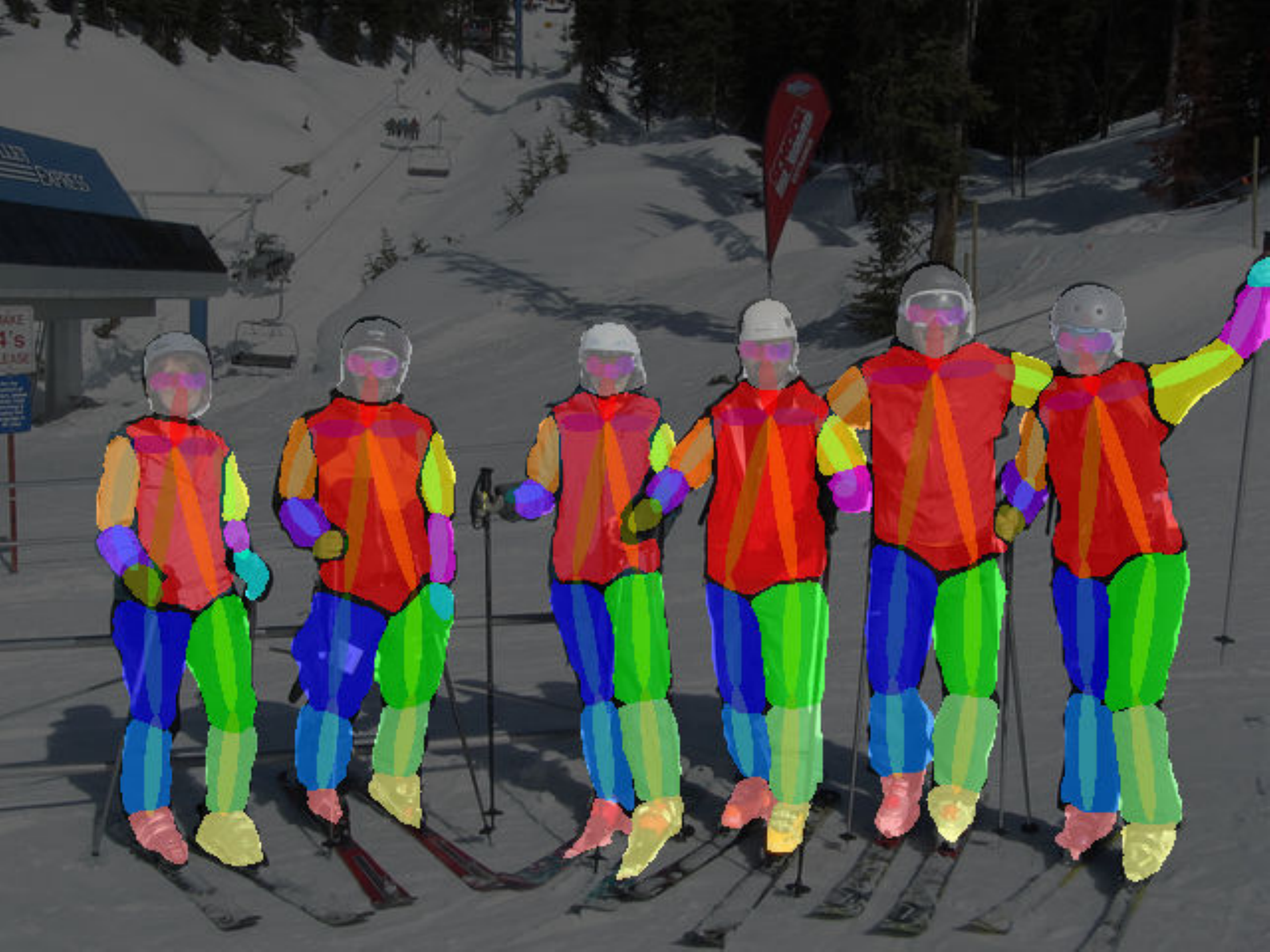}\\
 	\includegraphics[height=.1726\textwidth]{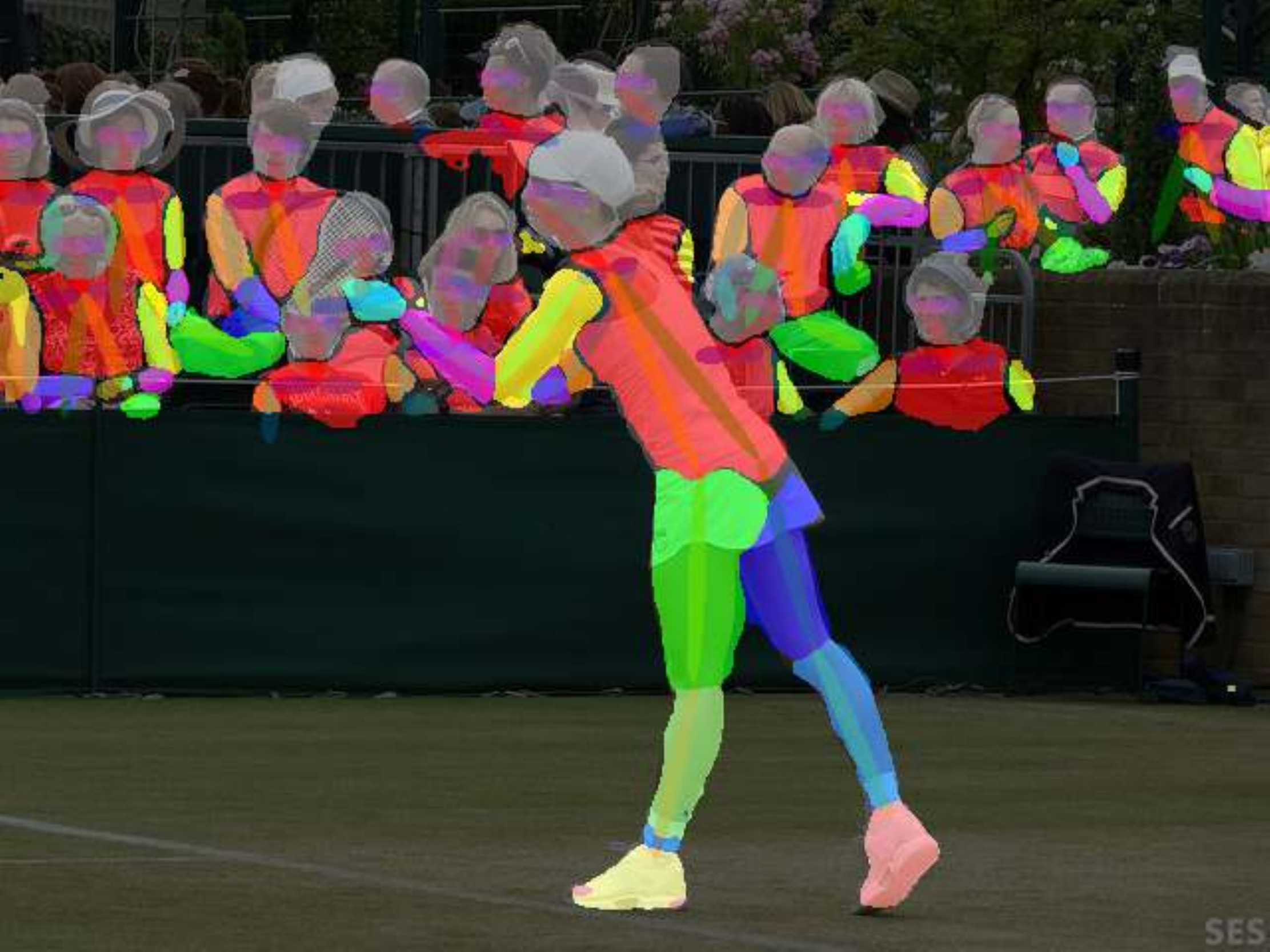}
 	\includegraphics[height=.1726\textwidth]{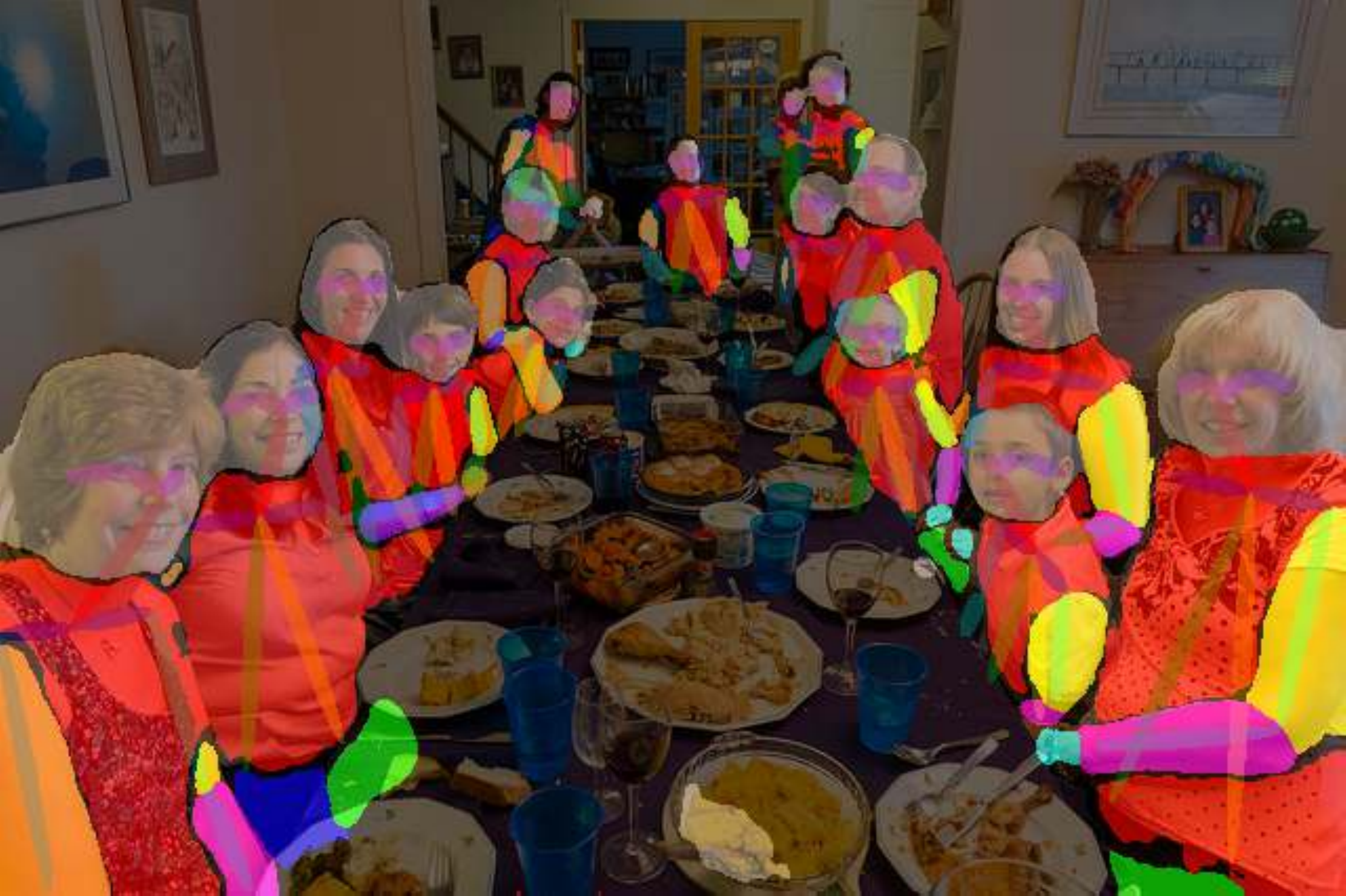}
 	\includegraphics[height=.1726\textwidth]{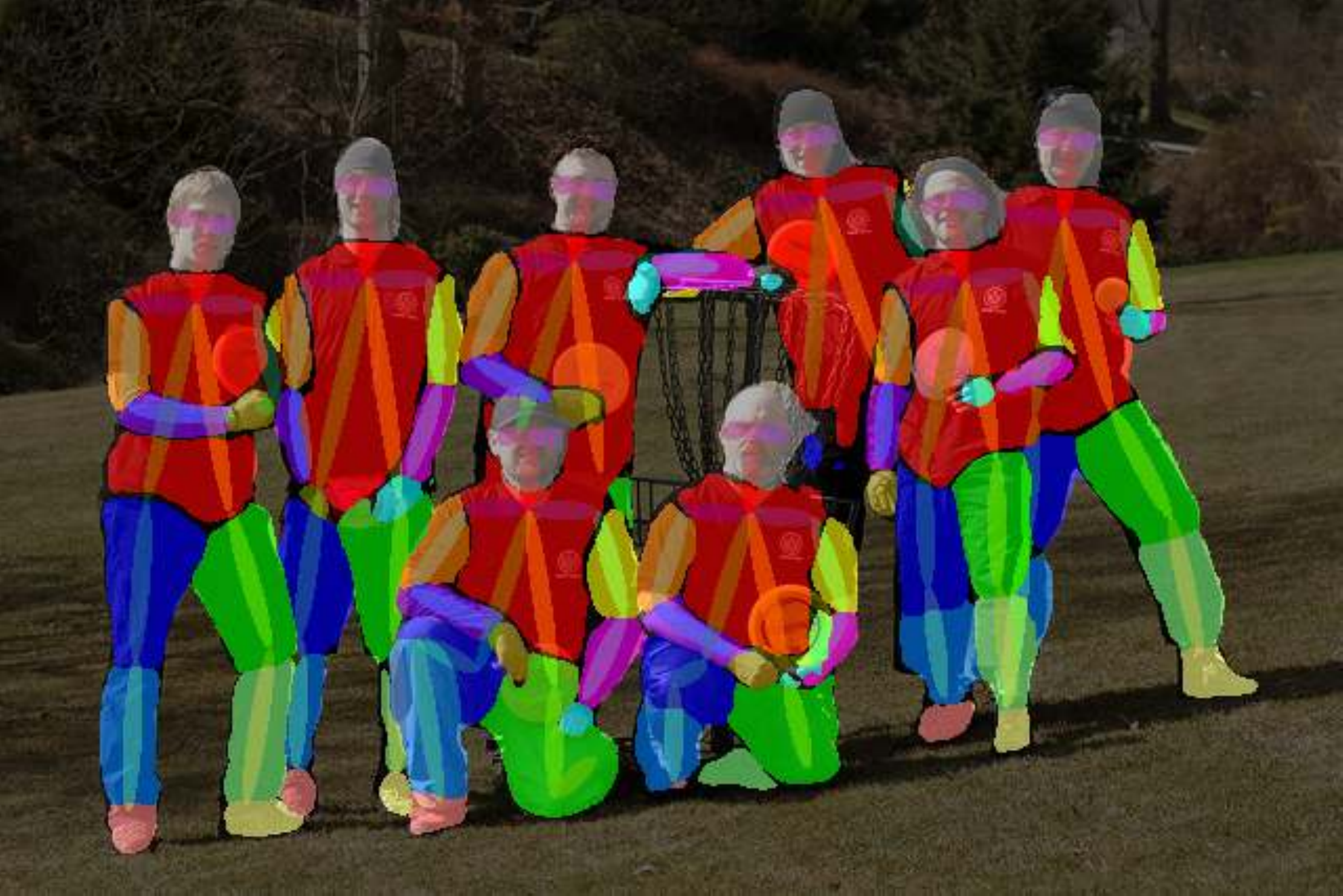}
 	\includegraphics[height=.1726\textwidth]{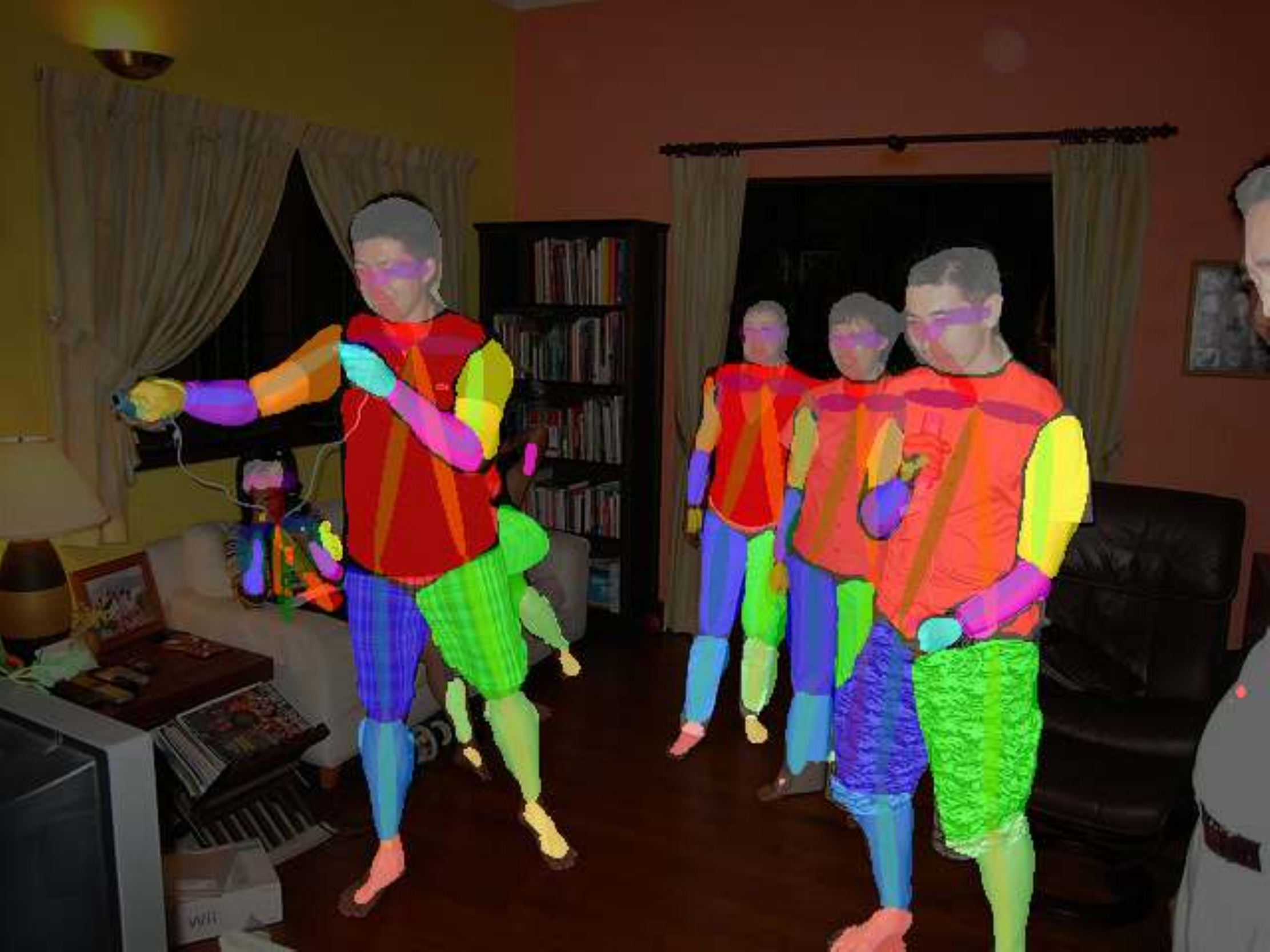}\\
 \caption{Qualitative results of the proposed method \textit{CDCL} on Pascal-Person-Parts and COCO validation images.}
\label{fig:seg-parts} 
\end{figure*}

\subsection{Training}
We initialized the backbone network using the pre-trained weights on ImageNet~\cite{krizhevsky2012imagenet}. The head networks are randomly initialized. During training, we randomly pick an equal number of real and synthetic images to form a mini-batch, and feed it to the network. Then, we compute the loss using Eq(\ref{eqn:overall-loss}), and update the network parameters via Adam optimizer with an initial learning rate $0.001$. The training batch size is set to $10$. Following the literature~\cite{papandreou2018personlab}, we set $\alpha=1.0$, $\beta = 1.0$, $\gamma=0.5$ to balance pose estimation loss and part segmentation loss. We refer the readers to Sec.\ref{sec:hyerparam} for further details on the hyperparameter analysis.

\subsection{Inference}
During testing, we only predict the part segmentation. Our model predicts $14$ body part score maps and one background score map. Following DeepLab~\cite{chen2018deeplab}, we run multi-scale inference and perform max-pooling to obtain the final part score maps. The part segmentation is derived by using the argmax value from the final part score maps. Given a fixed image size $368$x$654$, the average inference processing time is about 16 frames per second using a PC with a single Titan XP GPU.

\section{Experiments}\label{sec:results}

We train our model with COCO Keypoint dataset~\cite{lin2014microsoft} and our synthetic dataset. We then evaluated the performance of the resulting model on two public benchmarks, the Pascal-Person-Parts~\cite{chen_cvpr14}, and the COCO-DensePose~\cite{Guler2018DensePose}. 

 \subsection{Evaluation benchmarks}
 
\textbf{Pascal-Person-Parts~\cite{chen_cvpr14}} is a challenging dataset for multi-person body part segmentation. It consists of $1,716$ training and $1,817$ test images, where the human body is split into $6$ different parts including head, torso, upper and lower arms, as well as upper and lower legs.

\textbf{COCO-DensePose~\cite{Guler2018DensePose}} is a manually annotated dataset with the body part annotations. We evaluate multi-person body part segmentation on its body part annotations. The dataset contains $26,151$ training images, and the \textit{minival} has $1,508$ validation images.

\begin{table*}[t]
	\centering
	\setlength{\tabcolsep}{0pt}
	\caption{Performance comparison of human body part segmentation (mIOU, \%) on Pascal-Person-Parts dataset~\cite{chen_cvpr14}. Note that the symbol ``+'' indicates using additional real dataset with human-annotated segmentation labels.}
	\label{tbl:pascal}
	\begin{tabular*}{\textwidth}{@{}@{\extracolsep{\fill}}lcccccccccc@{}}
	\toprule
	Method & Real Seg. GT & Syn Seg. GT & \ \ Head \ \ & \ \ Torso \ \ & U-arms & L-arms & U-legs & L-legs & Bkg & Avg\\
	\midrule
	DeepLab-LFOV~\cite{chen2014semantic} & \cmark & \xmark & $78.09$ & $54.02$ & $37.29$ & $36.85$ & $33.73$ & $29.61$ & $92.85$ & $51.78$ \\
	HAZN~\cite{xia2016zoom}  & \cmark & \xmark & $80.79$ & $59.11$ & $43.05$ & $42.76$ & $38.99$ & $34.46$ & $93.59$ & $56.11$ \\
	Attention~\cite{chen2016attention}  & \cmark & \xmark & $81.47$ & $59.06$ & $44.15$ & $42.50$ & $38.28$ & $35.62$ & $93.65$ & $56.39$ \\
	LG-LSTM~\cite{liang2016semantic2}  & \cmark & \xmark & $82.72$ & $60.99$ & $45.40$ & $47.76$ & $42.33$ & $37.96$ & $88.63$ & $57.97$ \\
	LIP~\cite{gong2017look}  & \cmark & \xmark & $83.26$ & $62.40$ & $47.80$ & $45.58$ & $42.32$ & $39.48$ & $94.68$ & $59.36$ \\
	Graph LSTM~\cite{liang2016semantic} & \cmark & \xmark & $82.69$ & $62.68$ & $46.88$ & $47.71$ & $45.66$ & $40.93$ & $94.59$ & $60.16$ \\ 
	DeepLab v2~\cite{chen2018deeplab} & \cmark & \xmark & - & - & - & - & - & - & - & $64.94$ \\ 
	WSHP~\cite{fang2018weakly} & \cmark+ & \xmark & $87.15$ & $72.28$ & $57.07$ & $56.21$ & $52.43$ & $50.36$ & $97.72$ & $67.60$ \\
	CDCL  & \xmark & \cmark & $\textbf{75.53}$ & $\textbf{66.26}$ & $\textbf{63.28}$ & $\textbf{57.14}$ & $\textbf{47.75}$ & $\textbf{51.45}$ & $\textbf{93.72}$ & $\textbf{65.02}$  \\
    \midrule
    \midrule
	CDCL+Pascal  & \cmark & \cmark & $\textbf{86.39}$ & $\textbf{74.70}$ & $\textbf{68.32}$ & $\textbf{65.98}$ & $\textbf{59.86}$ & $\textbf{58.70}$ & $\textbf{95.79}$ & $\textbf{72.82}$ \\
	\bottomrule
	\end{tabular*}
\end{table*}
\begin{table*}[t]
	\centering
	\setlength{\tabcolsep}{0pt}
	\caption{Performance comparison of human body part segmentation (mIOU, \%) on COCO-DensePose human body masks~\cite{Guler2018DensePose}. 
	Note that the symbol ``+'' indicates using additional real dataset with human-annotated segmentation labels.}
	\label{tbl:densepose-coco}
	\begin{tabular*}{\textwidth}{@{}@{\extracolsep{\fill}}lccccccccccc@{}}
	\toprule
	Method & Real Seg. GT & Syn Seg. GT &Head & Torso & U-arms & L-arms & U-legs & L-legs & Bkg & Avg\\
	\midrule
	WSHP~\cite{fang2018weakly} & \cmark+ & \xmark &   $67.33$ & $62.22$ & $51.50$ & $55.66$ & $54.22$ & $53.11$ & $76.81$ & $60.12$ \\
	CDCL & \xmark & \cmark & $68.45$ & $66.21$ & $59.96$ & $51.72$ & $50.71$ & $50.57$ & $75.55$ & $60.45$ \\
	\midrule
	\midrule
	CDCL+Pascal & \cmark & \cmark & $66.16$ & $64.80$ & $60.33$ & $61.19$ & $55.97$ & $54.96$ & $92.03$ & $65.06$ \\
	CDCL+COCO & \cmark & \cmark &  $\textbf{73.15}$ & $\textbf{68.74}$ & $\textbf{63.79}$ & $\textbf{67.66}$ & $\textbf{63.39}$ & $\textbf{60.62}$ & $\textbf{93.55}$ & $\textbf{70.13}$ \\
	\bottomrule
	\end{tabular*}
\end{table*}

\subsection{Main results}
We compare our technique with several state-of-the-art supervised approaches, including HAZN~\cite{xia2016zoom}, Attention~\cite{chen2016attention}, LG-LSTM~\cite{liang2016semantic2}, LIP~\cite{gong2017look}, Graph LSTM~\cite{liang2016semantic}, DeepLab~\cite{chen2018deeplab,chen2014semantic}, and WSHP~\cite{fang2018weakly}.
Note that all these approaches use Pascal-Person-Parts dataset including the part segmentation labels as the training data while our network does not need to use any of the data from Pascal-Person-Parts at all. Following the settings of Pascal-Person-Parts~\cite{chen_cvpr14}, we predict $6$ body parts and measure the prediction results using the mean Intersection of Union (mIOU)~\cite{everingham2015pascal}.

Table~\ref{tbl:pascal} shows the performance comparison with different state-of-the-art methods, and Figure~\ref{fig:seg-parts} visualizes our prediction results. Without the segmentation training data provided by Pascal-Person-Parts, the proposed method~\textit{CDCL} achieves $65.02\%$ mIOU, which is comparable to or better than several state-of-the-art supervised approaches, such as DeepLab v2~\cite{chen2018deeplab} and Graph LSTM~\cite{liang2016semantic}. It is worth noting that the proposed \textit{CDCL} has better or similar performance compared to the other fully supervised methods, except for the head region. The main reason is that the head definition in our synthetic dataset does not match the head definition in Pascal-Person-Parts dataset. In our synthetic dataset, the head definition consistently includes the head and the neck. But in Pascal-Person-Parts, some of the ground truth head regions do not include the neck.

We further compare our method with the state-of-the-art approach~\cite{fang2018weakly} on COCO-DensePose. For a fair comparison, we follow the body part settings of WSHP~\cite{fang2018weakly}, and measure mIOU for the 6 different body parts and background. 
As shown on the second row \textit{CDCL} of Table~\ref{tbl:densepose-coco}, our result is slightly better than WSHP~\cite{fang2018weakly} which used real segmentation training data from both Pascal-Person-Parts and AIC~\cite{aic}. 

\subsection{Adding real data with part segmentation labels}

\begin{figure}[t]
	\centering
\includegraphics[width=.99\columnwidth]{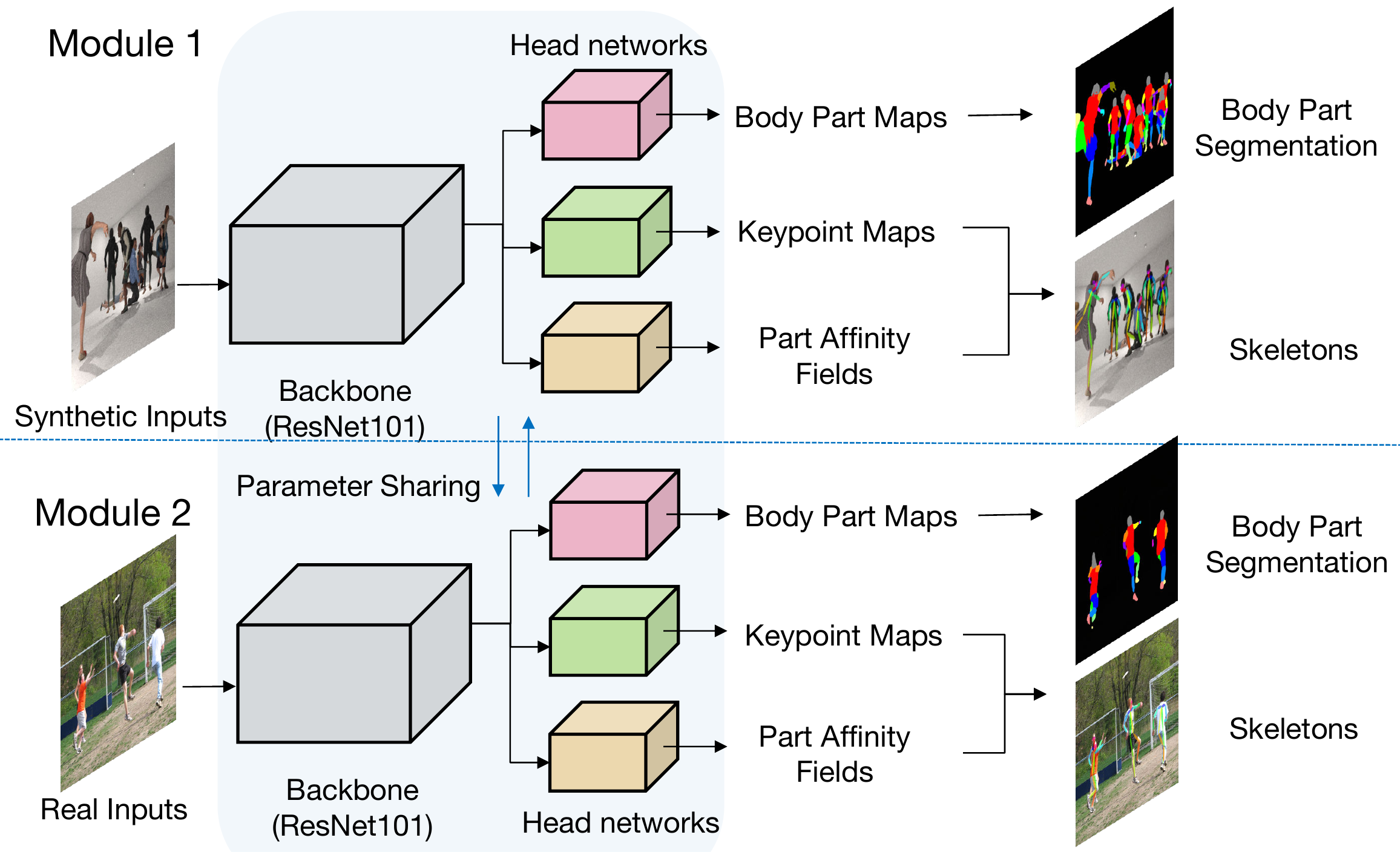}
	\caption{An extension of our training framework when real part segmentation labels are available for training.}
	\label{fig:network-extension}
\end{figure}

To obtain the performance \textit{upper bound} of our technique, we evaluate our method when real data with part segmentation labels are used during training. As shown in Figure~\ref{fig:network-extension}, we share the parameters of all the modules and train the network using real and synthetic datasets. The bottom row \textit{CDCL+Pascal} of Table~\ref{tbl:pascal} shows the result on Pascal-Person-Parts where Pascal-Person-Parts training data is used. Our method outperforms WSHP by a large margin.

The same model is evaluated on the COCO-DensePose test data and the result is shown on the third row \textit{CDCL+Pascal} of Table~\ref{tbl:densepose-coco}. Again it outperforms WSHP by a large margin.

If we use COCO-DensePose training data instead, and evaluate on COCO-DensePose test data, we obtain an additional gain and the result is shown on the fourth row \textit{CDCL+COCO} of Table~\ref{tbl:densepose-coco}.

\subsection{Comparison with adversarial learning}
\label{sec:adversarial}
Recent studies~\cite{ren2017cross,Tsai_adaptseg_2018,zhang2018fully} used adversarial training to align the feature spaces of the synthetic and real images. Thus, we compare the performance of our method with the adversarial training strategy. 
Since the model presented in~\cite{ren2017cross} cannot be directly used for part segmentation, we implemented our own network similar to~\cite{ren2017cross}. Our network has a backbone (ResNet101) and two head networks, one for the part segmentation head and the other for the discriminator.

Table~\ref{tbl:compare-adv} shows the performance comparison on two datasets. We can see that adversarial training (\textit{ADV}) achieves better performance than that of training with synthetic data only without adversarial training (\textit{SYN}), but it does not perform as well as our complementary learning technique.

\begin{table}[t]
\centering
\caption{Performance comparison of human body part segmentation (mIOU, \%) of different methods.}
\label{tbl:compare-adv}
\begin{tabular}{lcc}
    \toprule
	Method  & Pascal-Person-Parts & COCO-DensePose\\
	\midrule
	SYN & $10.18$ & $10.12$ \\
	ADV & $16.42$ & $19.24$ \\
	CDCL & $\textbf{65.02}$ &  $\textbf{60.45}$\\
	\bottomrule
\end{tabular}
\end{table} 

\begin{figure}[t]
 \centering
 \setlength{\tabcolsep}{1pt}
 \begin{tabular}{cccc}
 	\includegraphics[height=.095\textwidth]{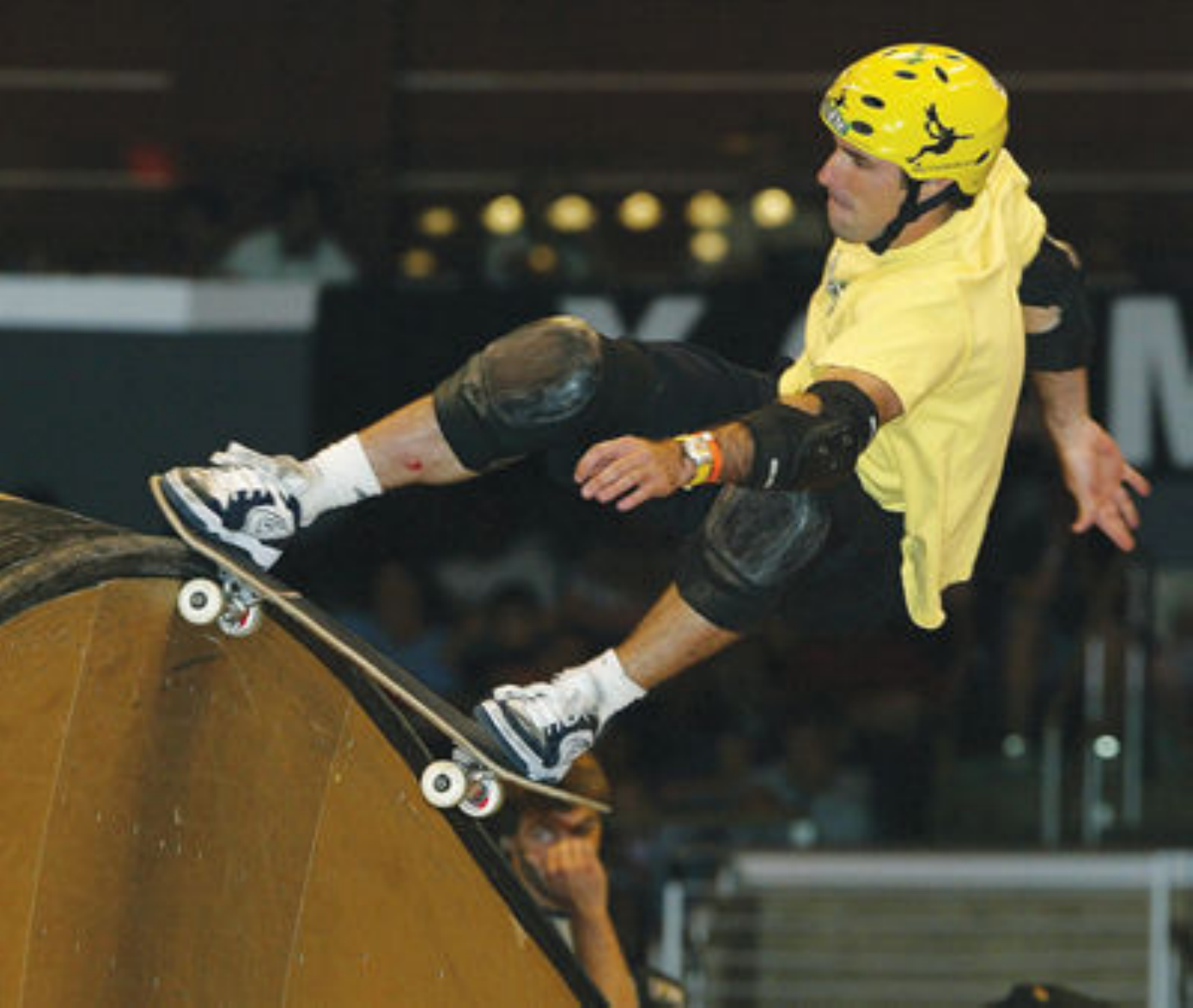}
 	&\includegraphics[height=.095\textwidth]{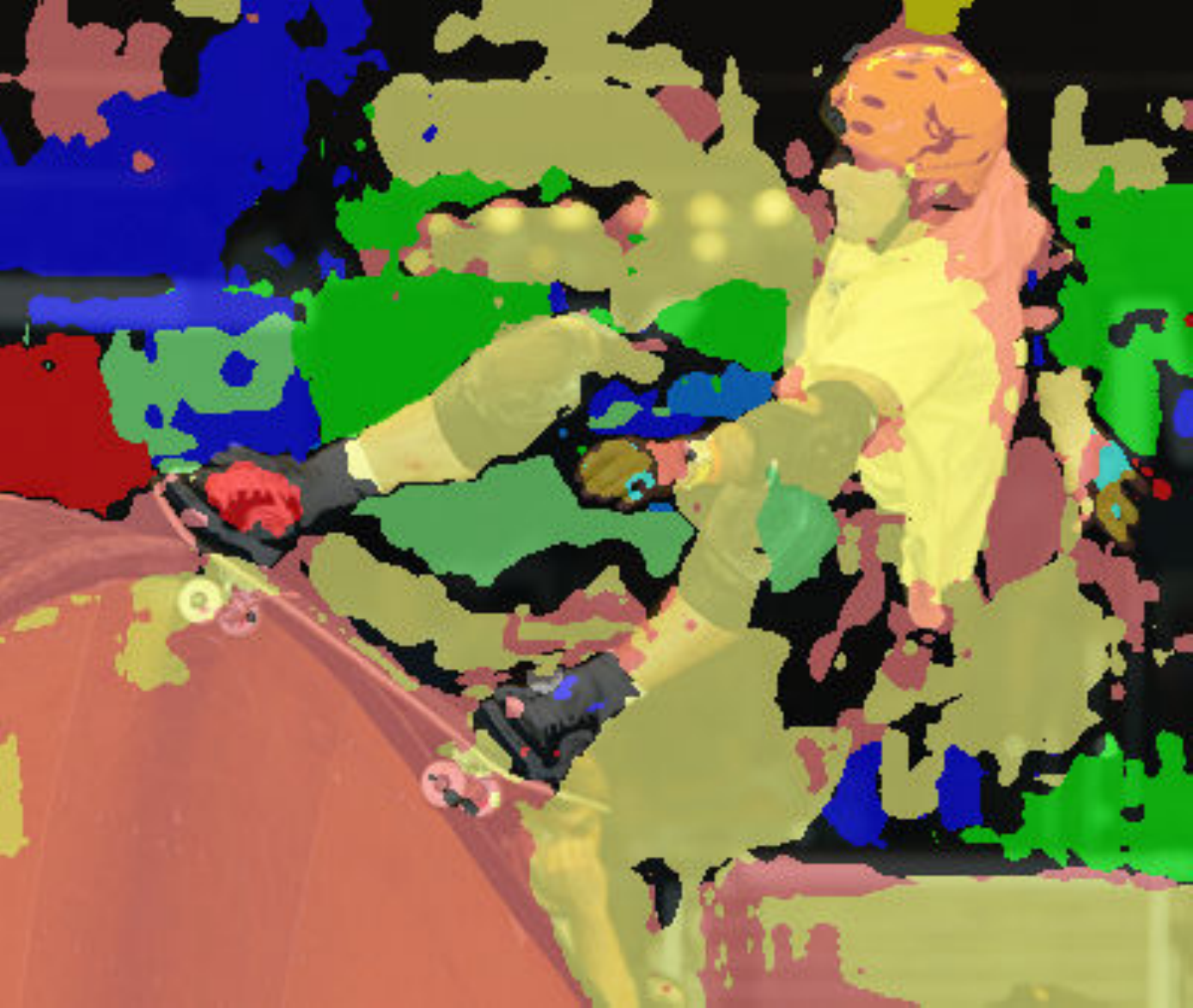}
 	&\includegraphics[height=.095\textwidth]{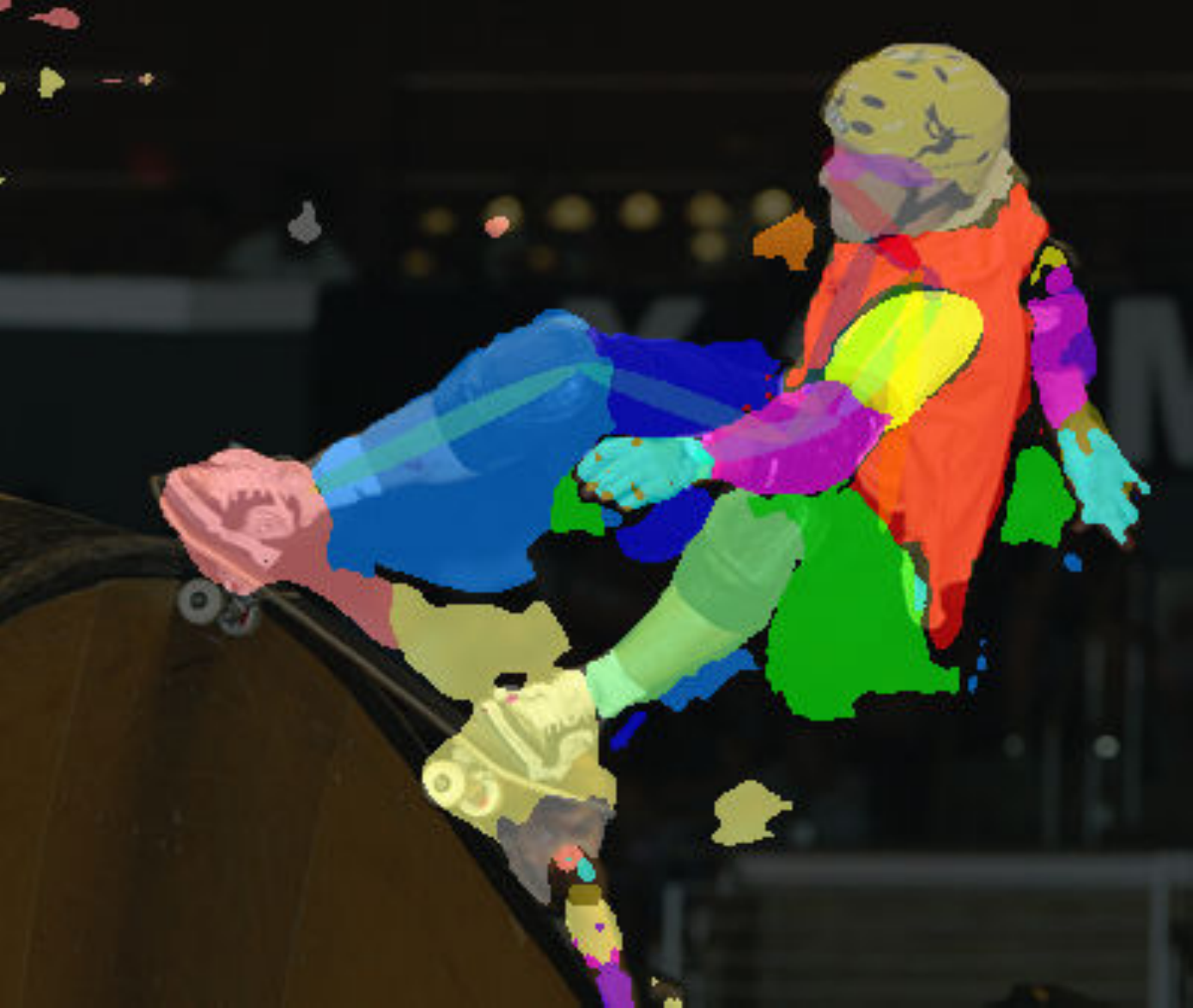}
 	&\includegraphics[height=.095\textwidth]{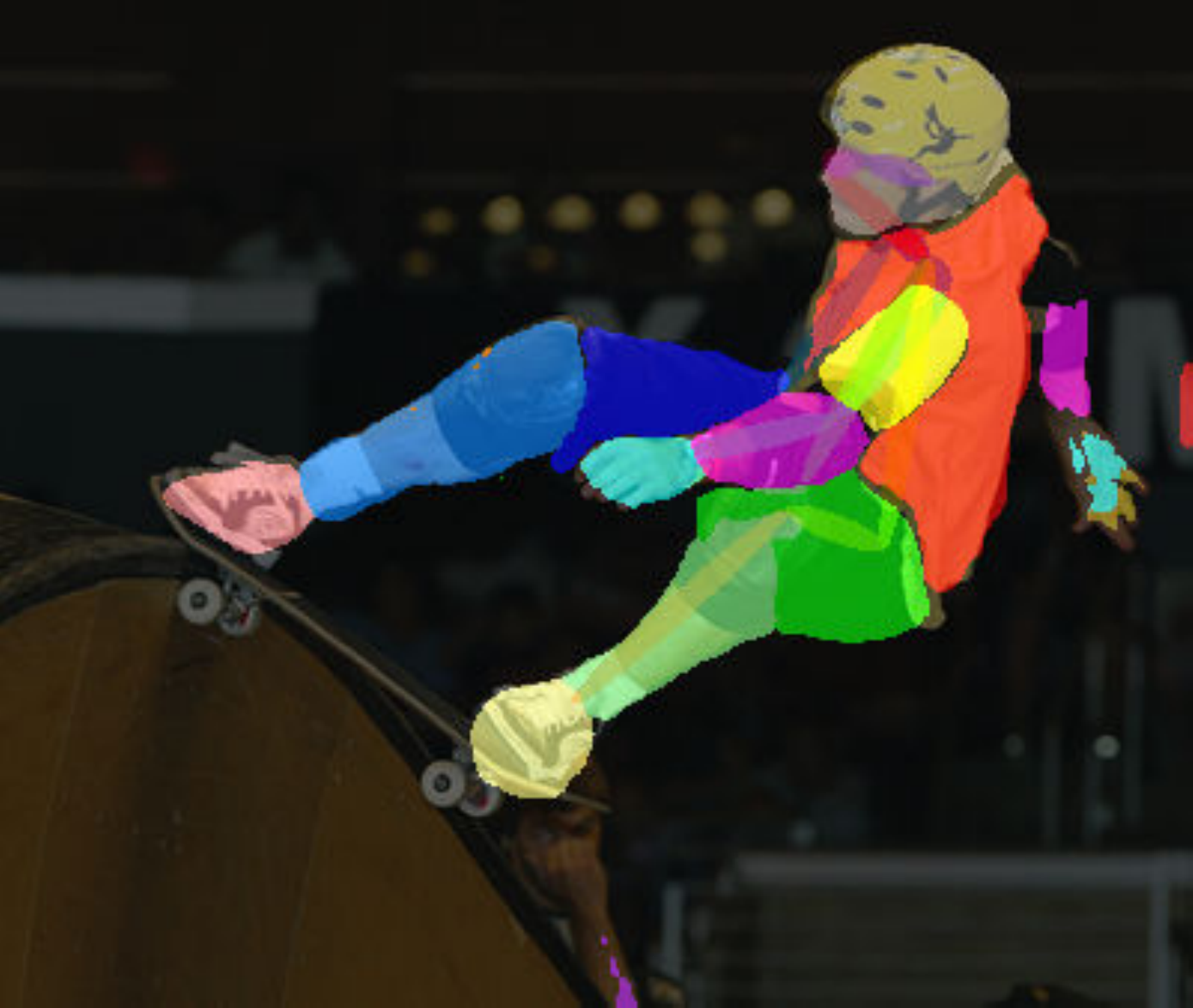}\\
 	\small{Input image} & \small{SYN} & \small{NO-SP} & \small{CDCL}\\
    \end{tabular}
 \caption{Qualitative comparison of the proposed method with different training strategies.}
\label{fig:ablation} 
\end{figure}

\begin{table}
\centering
\caption{Ablations of training with different types of data.}
\label{tbl:abalation-pascal}
\begin{tabular}{lccccc}
    \toprule
	Method  & Syn. & Syn. & Real & Pascal & COCO\\
	 & Parts & Poses & Poses & mIOU & mIOU\\
	\midrule
	SYN & \cmark & \cmark & \xmark  & $10.18$ & $10.12$ \\
	NO-SP & \cmark & \xmark & \cmark  & $49.71$ & $50.66$ \\
	CDCL & \cmark & \cmark & \cmark  & $\textbf{65.02}$ & $\textbf{60.45}$ \\
	\bottomrule
\end{tabular}
\end{table} 

\textcolor{blue}{
\begin{table}
\centering
\caption{Performance comparison (mIOU, \%) with the fully supervised baseline.}
\label{tbl:abalation-baseline}
\begin{tabular}{lcc}
    \toprule
	Method  & Pascal-Person-Parts & COCO-DensePose\\
	\midrule
	CDCL & $65.02$ & $60.45$ \\
	Fully-supervised & $65.40$ & $61.12$ \\
	\bottomrule
\end{tabular}
\end{table}
}
\subsection{Ablation study}
\label{sec:ablation}
\subsubsection{\textbf{Synthetic pose labels}}
Since our approach uses both synthetic poses and real poses, one interesting question is whether the synthetic pose is useful. To answer this question, we have trained our network without the synthetic poses (i.e. with synthetic parts and real poses). This configuration is denoted as \textit{NO-SP}, and the results on Pascal-Person-Parts and COCO-DensePose are shown in Table~\ref{tbl:abalation-pascal}. For completeness, we also show the results of \textit{SYN} (synthetic parts + synthetic poses), and \textit{CDCL} (synthetic parts + synthetic poses + real poses). We can see that \textit{NO-SP} outperforms \textit{SYN} by a large margin thanks to the knowledge learned from the real data, and adding synthetic poses further boosts the performance. Figure~\ref{fig:ablation} shows a qualitative comparison of the three configurations. \textit{SYN} has trouble handling the background, \textit{NO-SP} performs much better, and \textit{CDCL} further improves upon \textit{NO-SP}.

\subsubsection{\textbf{Fully supervised baseline}} We study a fully supervised baseline by removing the synthetic training data, and train a model using real part segmentation labels only. This configuration is denoted as \textit{Fully-supervised}, and the results are shown in Table~\ref{tbl:abalation-baseline}. We see that \textit{CDCL} performs comparably to \textit{Fully-supervised} because \textit{CDCL} effectively reduces the domain gap.

\begin{figure}[t]
	\setlength{\tabcolsep}{0.5pt}
	\renewcommand{\arraystretch}{2}
	\begin{tabular}{lll}
	\rotatebox[origin=c]{90}{\small{Left Elbow}} &
	\raisebox{-.5\height}{\includegraphics[height=.35\columnwidth]{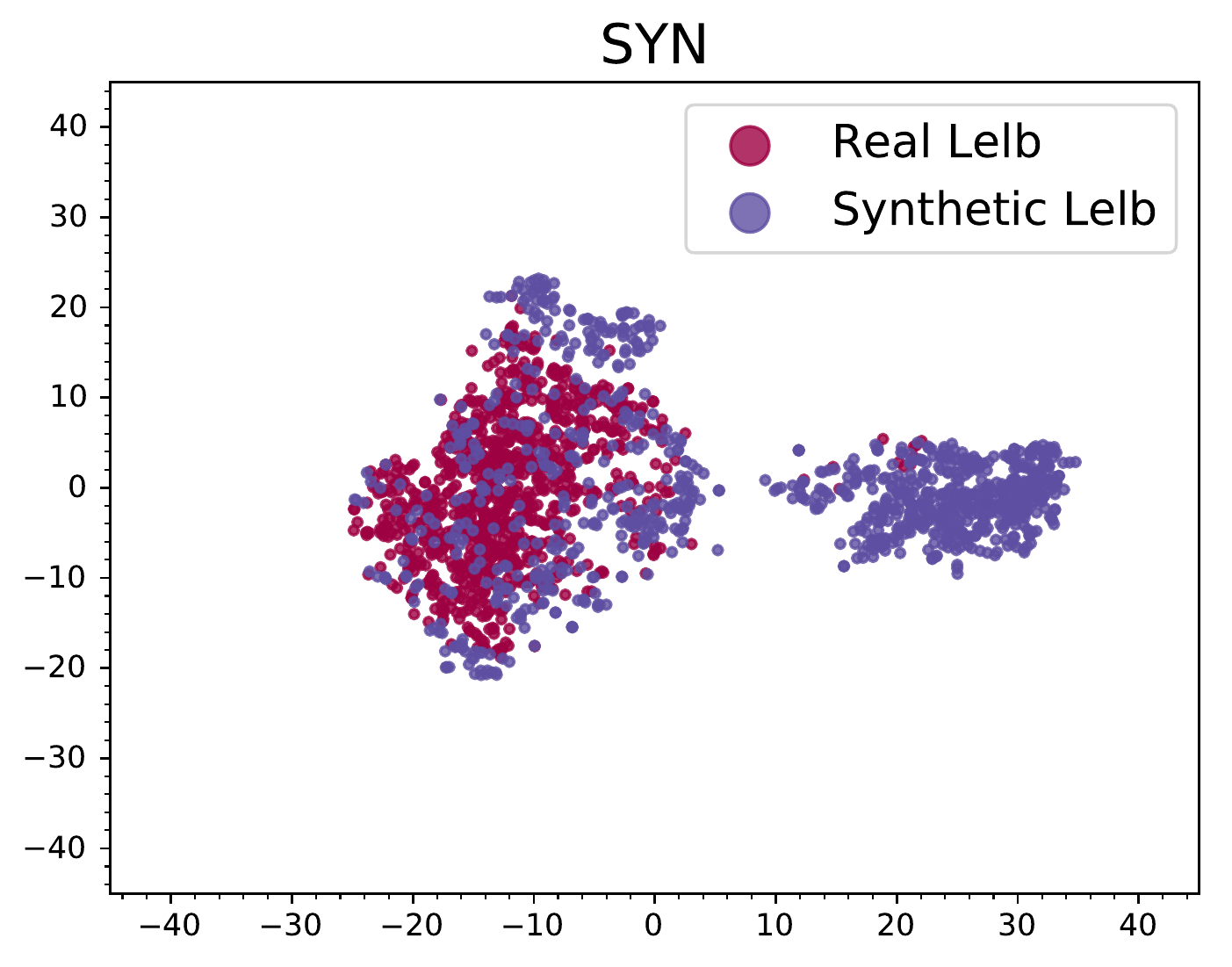}} &
	\raisebox{-.5\height}{\includegraphics[height=.35\columnwidth]{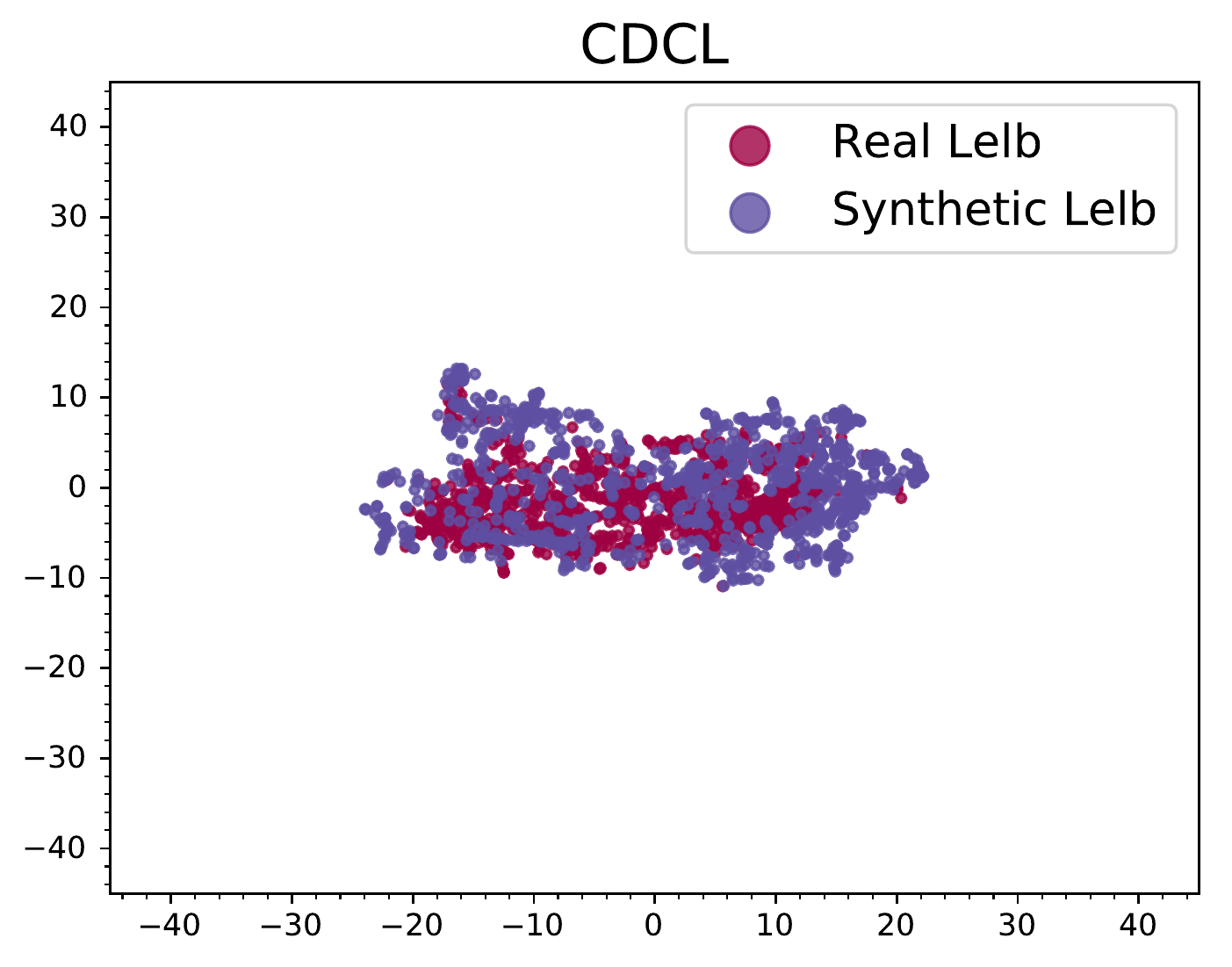}}\\[0.9cm]
	\rotatebox[origin=c]{90}{\small{Right Knee}} &
	\raisebox{-.5\height}{\includegraphics[height=.35\columnwidth]{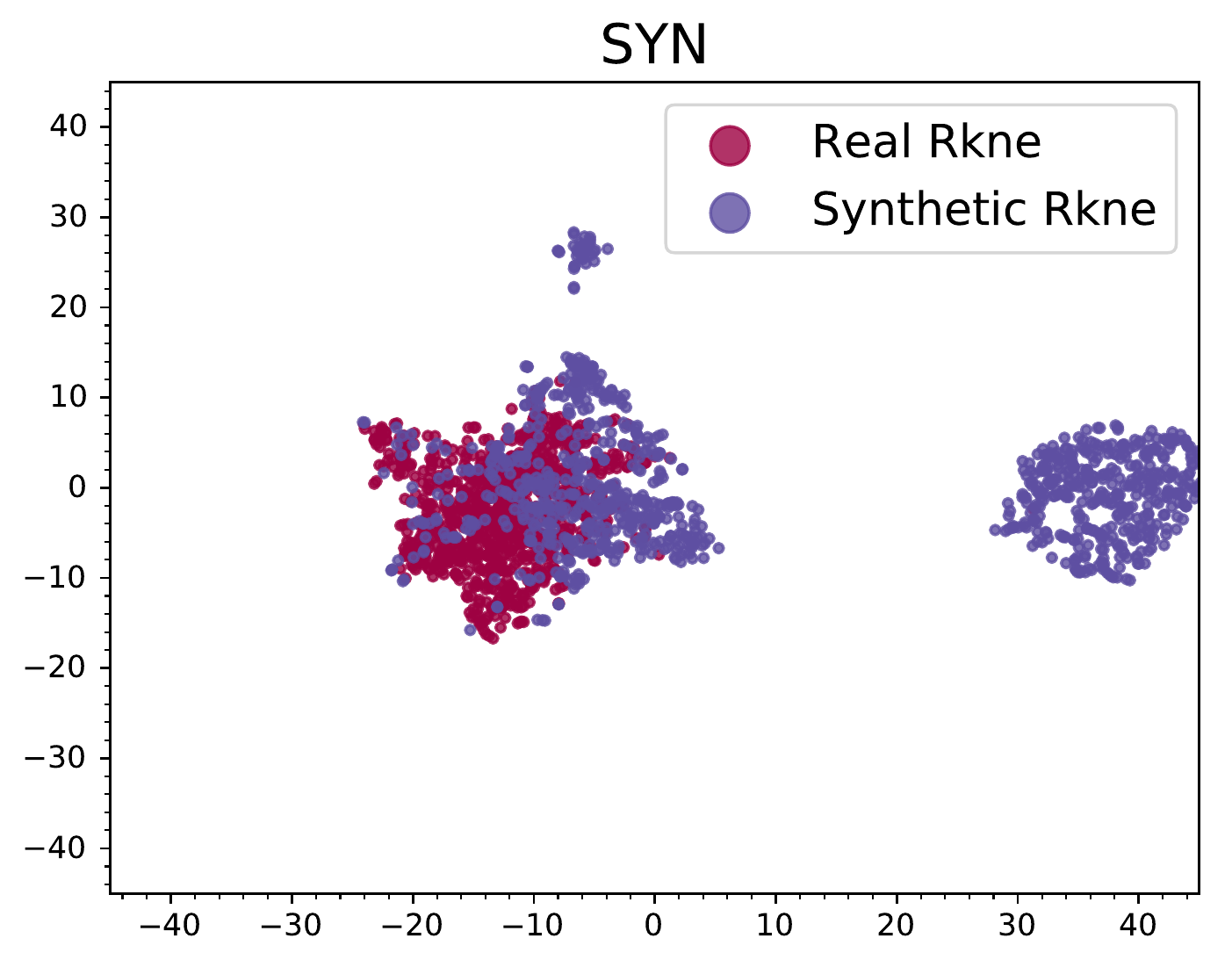}} &
	\raisebox{-.5\height}{\includegraphics[height=.35\columnwidth]{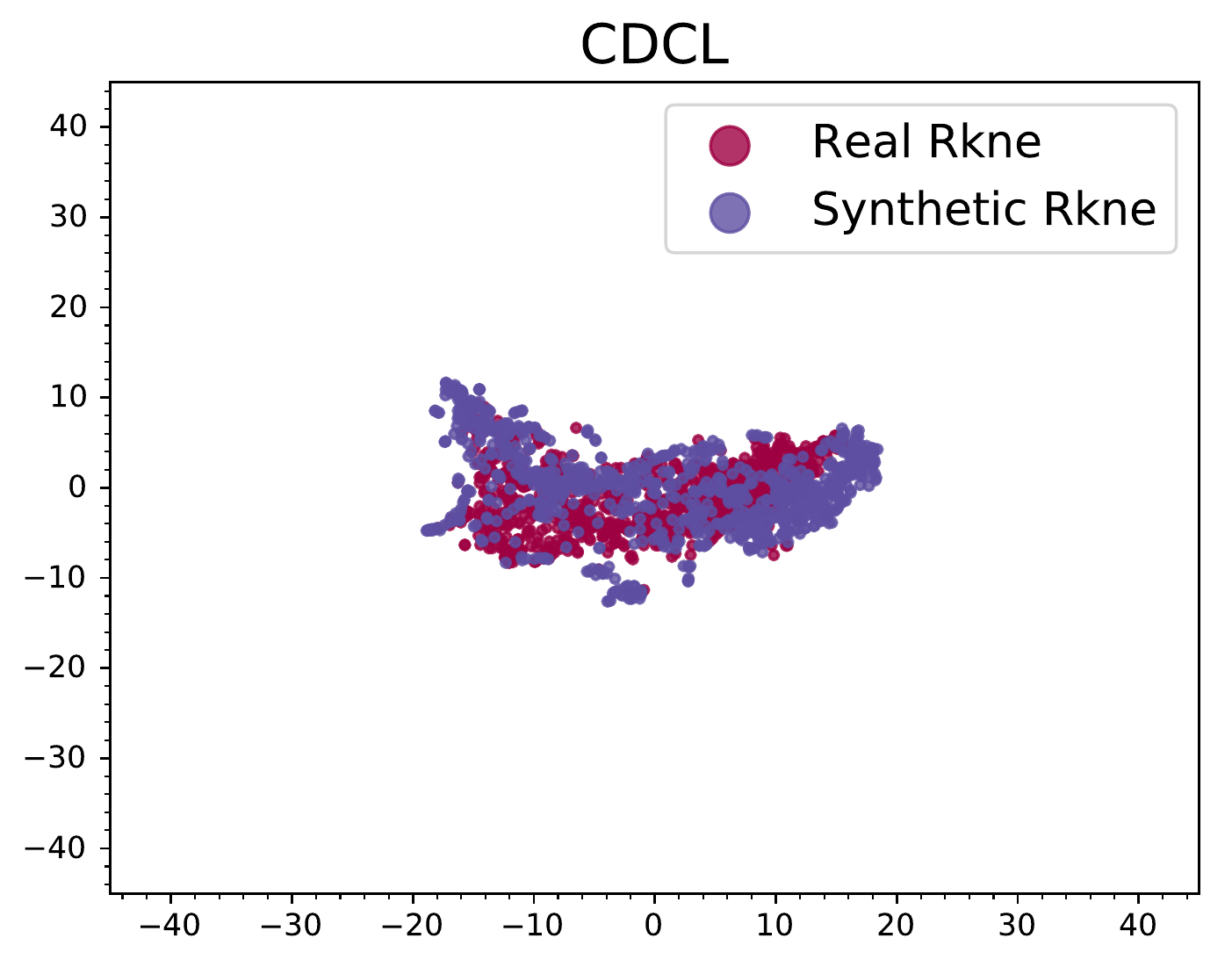}}
	\end{tabular}
	\caption{t-SNE visualization~\cite{maaten2008visualizing} of the feature spaces of the real and synthetic body parts.}
	\label{fig:feat-visual}
\end{figure}

\label{sec:feat-visual}
\subsubsection{\textbf{Feature space visualization}}
 
We visualize the features of two different models (\textit{SYN} and \textit{CDCL}) from the real and synthetic images using the t-SNE visualization technique~\cite{maaten2008visualizing}. In Figure~\ref{fig:feat-visual}, the left column shows the features extracted with the model \textit{SYN} (trained with synthetic data only), and the right column are from the model \textit{CDCL}. The first row shows the features extracted at the left elbow position, and the second row shows the features extracted at the right knee position. In each plot, the red dots indicate the real data while the purple dots indicate the synthetic data. We can see that the red and purple dots in the right column are aligned very well, but they do not align well in the left column. This indicates that our complementary learning technique is effective at aligning the feature space of the real data with that of the synthetic data.

\begin{figure}[b]
 \centering
 \setlength{\tabcolsep}{1pt}
 \begin{tabular}{cccc}
 	\includegraphics[width=.24\columnwidth]{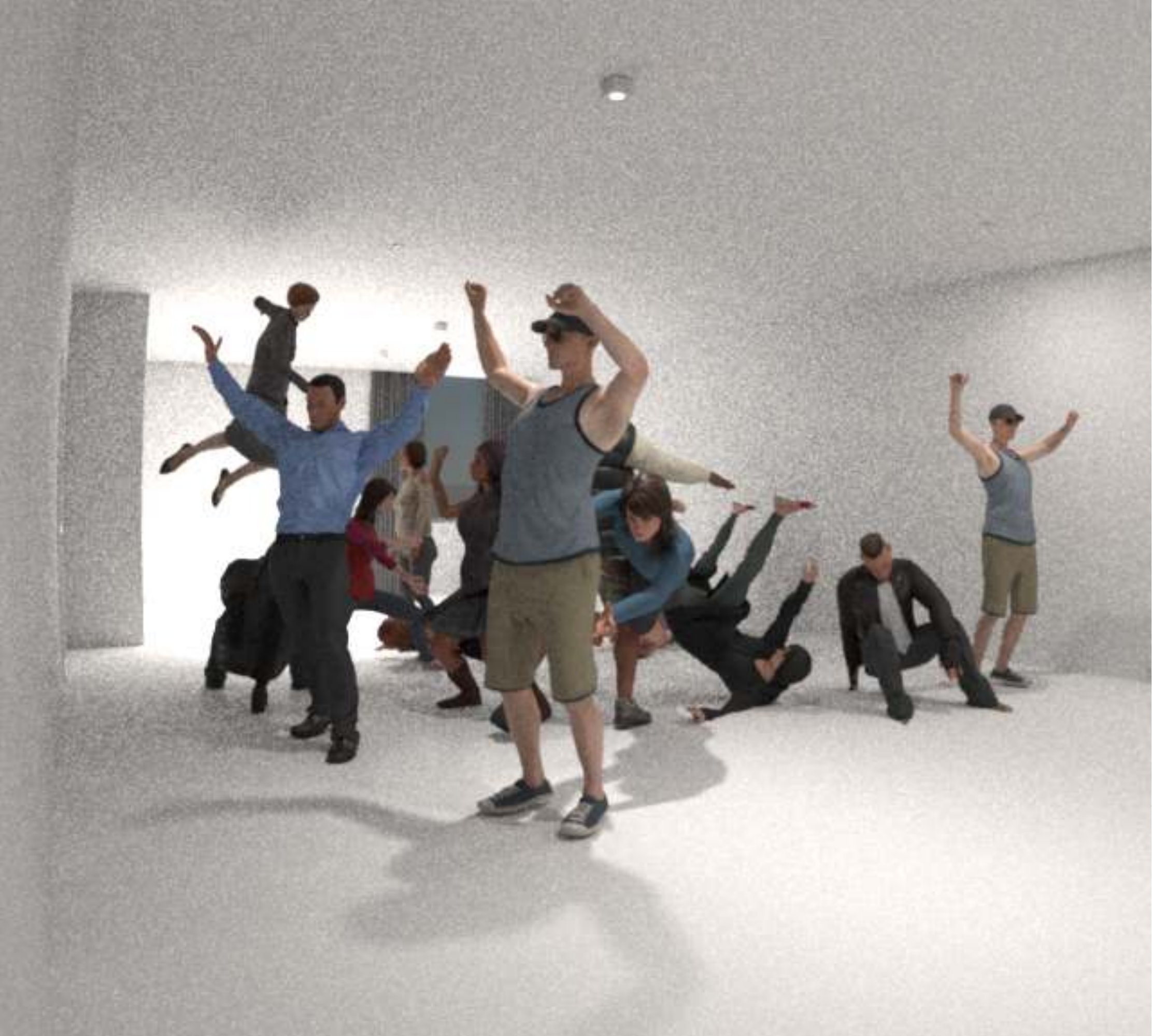}
 	&\includegraphics[width=.24\columnwidth]{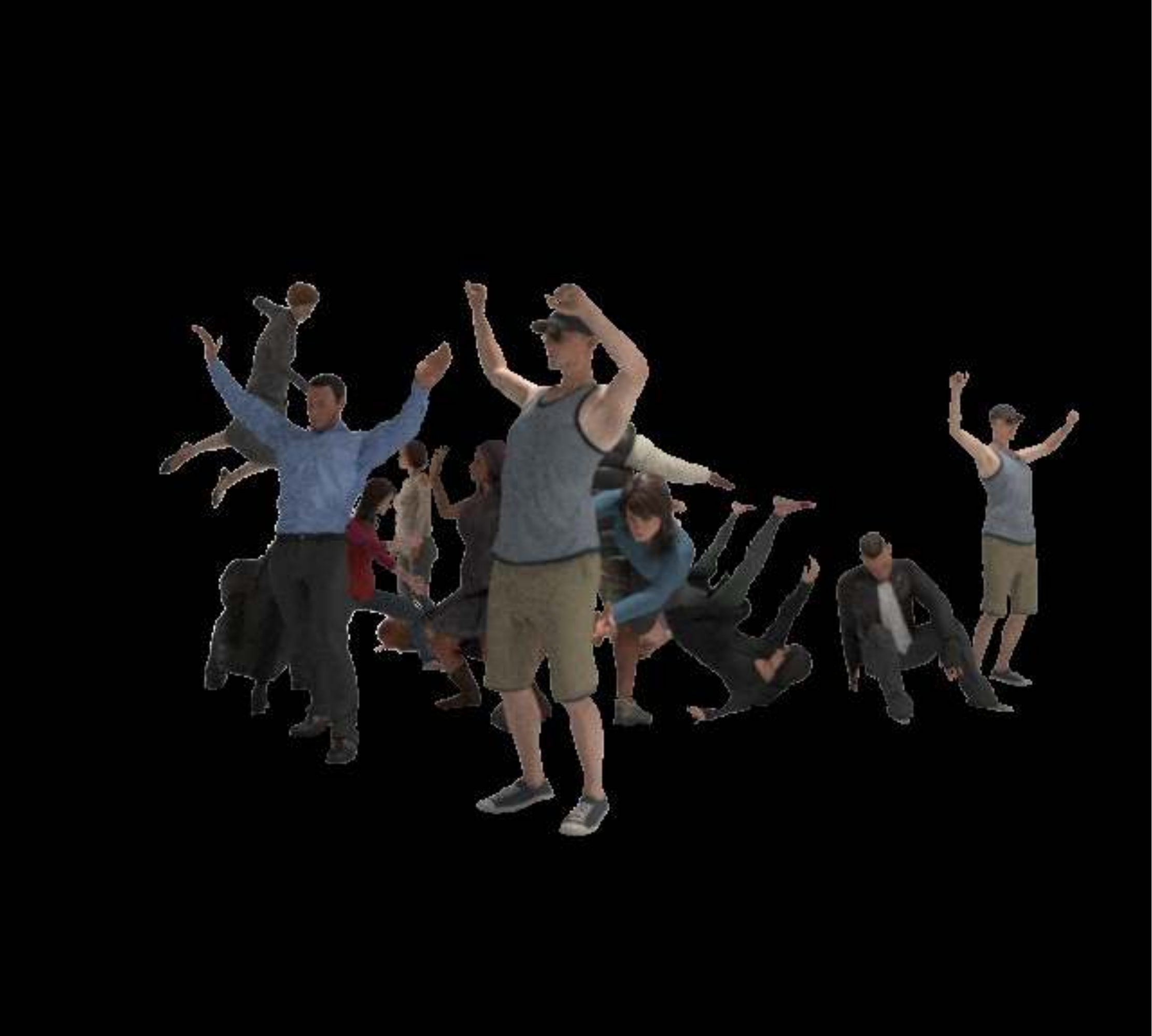} &
    \includegraphics[width=.24\columnwidth]{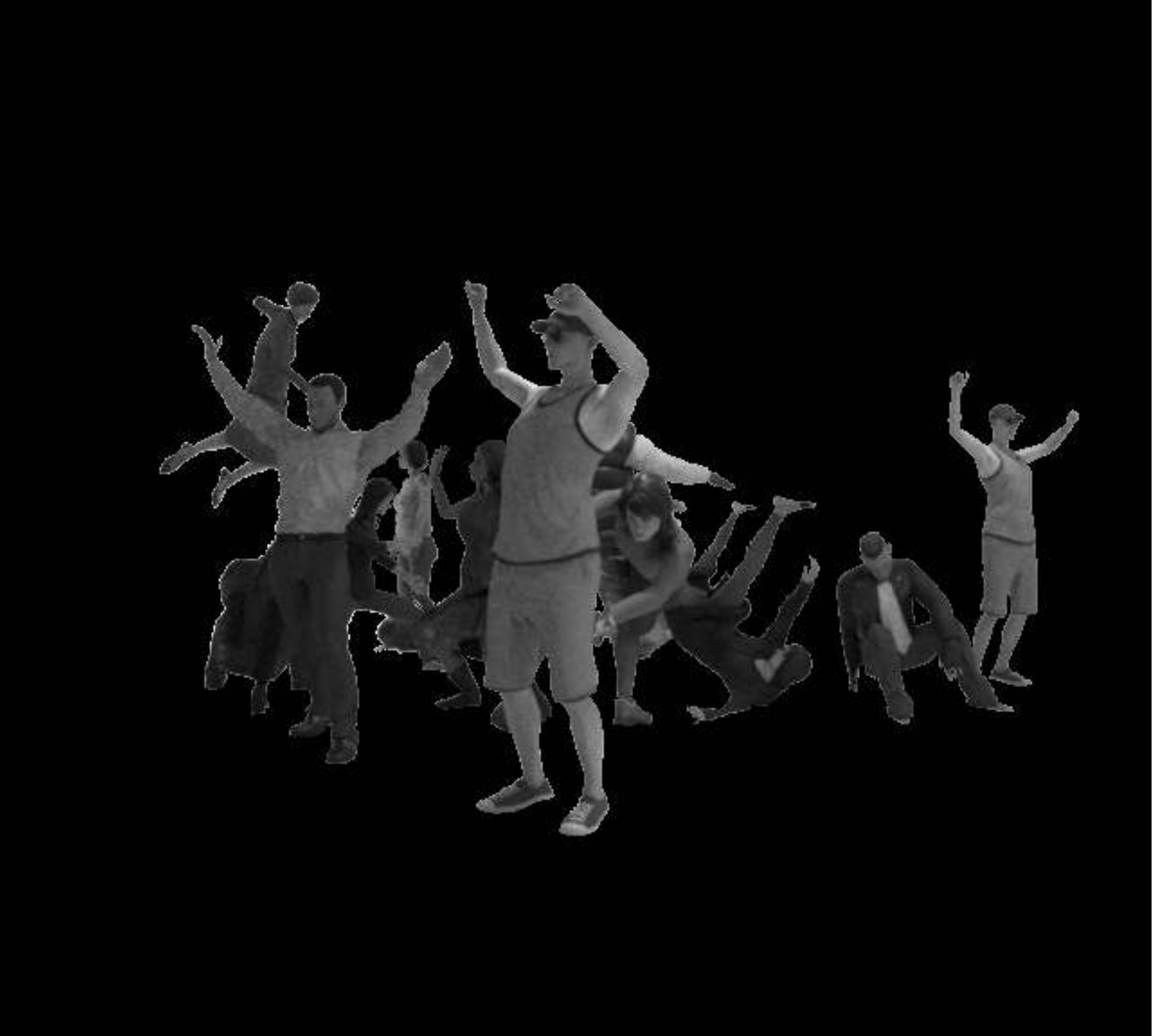}
 	&\includegraphics[width=.24\columnwidth]{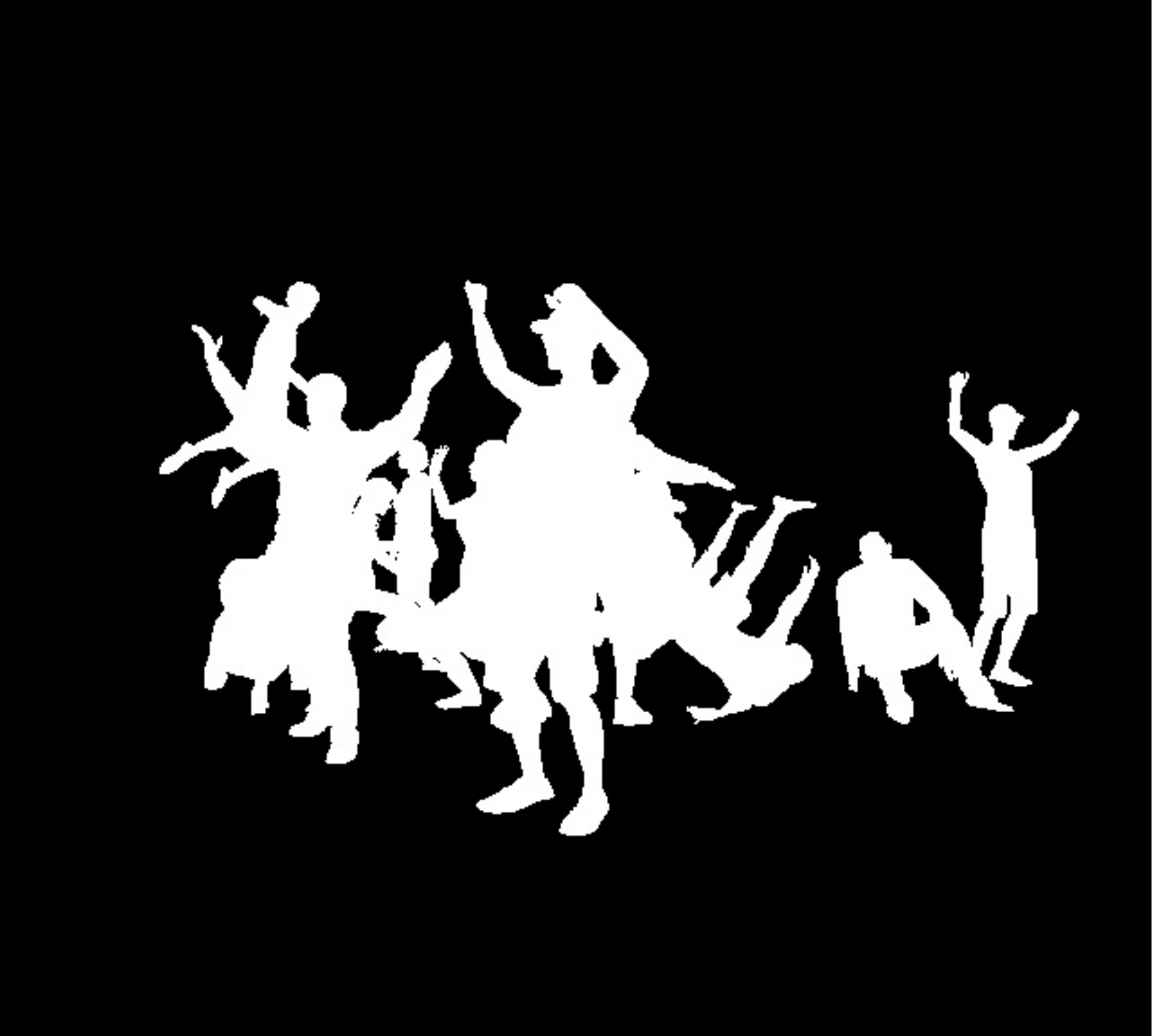}\\
    \small{Original} & \small{No Background} & \small{Gray-scale} & \small{Binary Mask}\\
    \end{tabular}
 \caption{Different configurations of the synthetic data.}
\label{fig:ablation-syn} 
\end{figure}
\begin{table}[b]
	\centering
	\caption{Performance comparison (mIOU, \%) of our method using different synthetic training data.}
	\label{tbl:syn-study}
	\resizebox{1.\columnwidth}{!}{
    \begin{tabular}{lcccc}
	\toprule
	 & Original & No Background & Gray-scale & Binary Mask\\
	 \midrule
	 Pascal-Person-Parts & $65.02$ & $63.96$  & $62.78$ & $43.38$\\
	 COCO-DensePose & $60.45$ & $59.39$ & $58.34$ & $40.77$\\
	\bottomrule
	\end{tabular}
	}
\end{table}

\subsubsection{\textbf{Synthetic training data analysis}}
\label{synthetic-data-analysis}
Since our method learns part segmentation from synthetic data, one may wonder what elements of the synthetic data are essential to be rendered. To answer the question, we ablate our synthetic training data by gradually removing the background, colors, and the human texture, and train our model with these configurations, respectively. Figure~\ref{fig:ablation-syn} shows the examples of different configurations of the synthetic training data, and Table~\ref{tbl:syn-study} shows the performance comparison on Pascal-Person-Parts and COCO-DensePose datasets. Firstly, we observe that removing the background from the synthetic data causes only a small drop on the segmentation performance. This is an indication that our framework is learning the background from the real data. Secondly, after we further remove the color of the synthetic data (Gray-scale), we again only see a small drop on the performance. Finally, when we degrade our synthetic data to the extreme by just using binary masks, our framework still works reasonably well. These studies indicate that our framework mainly requires the pose variations in the foreground data and the rendering quality is not as critical compared to the conventional approach of directly training from synthetic data.

\begin{table}
	\centering
	\caption{Ablation study (mIOU, \%) of our method using different number of synthetic human models for training.}
	\label{tbl:different-human}
	\resizebox{1.\columnwidth}{!}{
    \begin{tabular}{lccccc}
	\toprule
	 Number of Human Models & 1 & 5 & 10 & 15 & 20\\
	 \midrule
	 Pascal-Person-Parts & $25.20$ & $52.12$  & $64.91$ & $64.78$ & $65.02$\\
	 COCO-DensePose & $22.12$ & $51.65$  & $60.22$ & $60.41$ & $60.45$\\
	\bottomrule
	\end{tabular}
	}
\end{table}

\begin{table}
	\centering
	\caption{Ablation study (mIOU, \%) of our method when compositing synthetic humans with different number of backgrounds for training.}
	\label{tbl:different-back}
    \begin{tabular}{lccc}
	\toprule
	 Number of Backgrounds & 1 & 100 & 1000\\
	 \midrule
	 Pascal-Person-Parts & $16.23$ & $49.81$  & $50.19$\\
	 COCO-DensePose & $14.50$ & $46.33$  & $48.65$ \\
	\bottomrule
	\end{tabular}
\end{table}
\begin{figure}[t]
 \centering
 \includegraphics[width=.99\columnwidth]{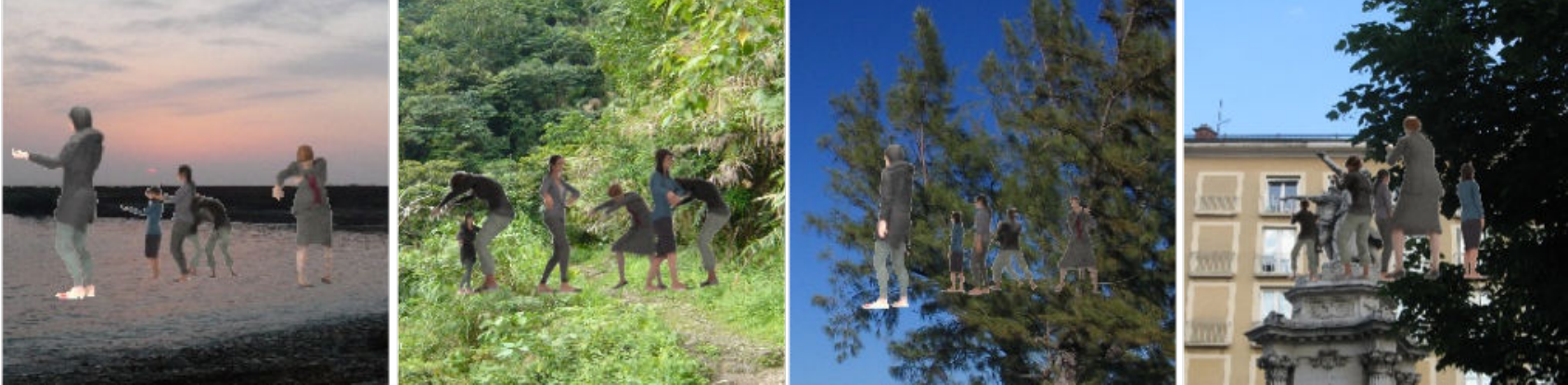}
 \caption{Examples of compositing synthetic humans with a variety of real-world scenery images.}
\label{fig:real-bk} 
\end{figure}  
 
\subsubsection{\textbf{Influence of the synthetic human geometry}}
Since our synthetic humans each have their own geometry of body shape and clothing, we study the influence of the model geometry by training with different number of synthetic human models.
Table~\ref{tbl:different-human} shows such results on Pascal-Person-Parts and COCO-DensePose datasets. We see that as we increase the number of synthetic human models for training, it improves the performance for part segmentation. However, the impact becomes less prominent if we use more than $10$ synthetic human models for training.

\subsubsection{\textbf{Compositing synthetic humans with different backgrounds}}
Since our synthetic dataset uses a single background of empty room, one may wonder what if we use more variety of backgrounds for training. Because it is time consuming to create a large variety of synthetic 3D background models, we composite the synthetic humans with a variety of real-world scenery images. We randomly select $1000$ scenery images from the Holidays dataset~\cite{jegou2008hamming} for data generation. Figure~\ref{fig:real-bk} shows a few examples of the composited images and Table~\ref{tbl:different-back} shows the results on Pascal-Person-Parts and COCO-DensePose datasets when we composite synthetic humans with different number of backgrounds. We can see that increasing the number of backgrounds improves the performance of part segmentation, but it does not work as well as using a simple background such as an empty room or a blank background. This is probably because the synthetic humans are not placed in realistic positions in the scene and there are lighting inconsistency between the synthetic humans and background. 
 
\begin{figure}
	 \centering
 \setlength{\tabcolsep}{1pt}
 \begin{tabular}{c}
 	\includegraphics[width=.6\columnwidth]{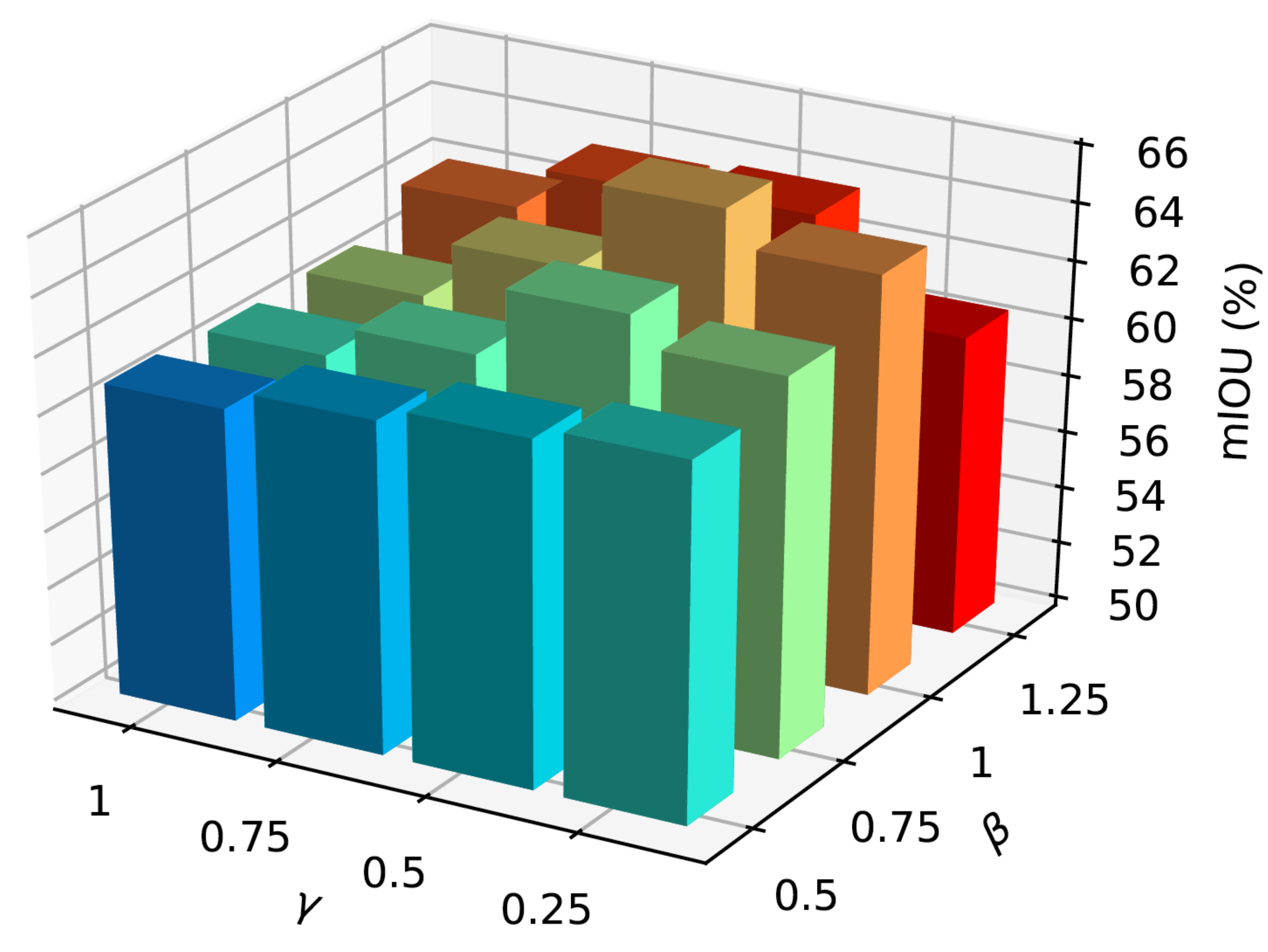}
    \end{tabular}
	\caption{Performance comparison (mIOU, \%) of CDCL with different combinations of hyperparameters on Pascal-Person-Parts dataset.}
	\label{fig:param-search}
\end{figure}

\subsubsection{\textbf{Influence of different losses}}
\label{sec:hyerparam}
We study the influence of the three terms in our learning objective (in Eq.(1)). The first two terms learn pose estimation from real and synthetic data, respectively. The third term learns part segmentation from synthetic data. We study the influence of the three terms by fixing the first weight $\alpha=1.0$ and iterating different combinations of $\beta$ and $\gamma$. The reason we set $\alpha=1.0$ is that we can rewrite Eq.(1) as $L = \alpha\Bigl( L_{pose}(D_r^{pose}) +  \frac{\beta}{\alpha} L_{pose}(D_s^{pose})+ \frac{\gamma}{\alpha} L_{part}(D_s^{part})\Bigr)$. Thus, we can omit the scaling factor by setting $\alpha=1.0$ and vary $\beta$ and $\gamma$. As shown in Figure~\ref{fig:param-search}, our method performs more favorably when $\gamma \simeq 0.5\times \beta$. Our method achieves the best performance when $\alpha$, $\beta$, $\gamma$ are set to $1.0$, $1.0$ and $0.5$, respectively. This indicates that the first two terms are equally important. We also observed that part segmentation loss is greater than pose estimation loss, thus the losses are better balanced when $\gamma$ is smaller than $\beta$. It is worth noting that the hyperparameters (i.e., $\alpha$, $\beta$, $\gamma$) are used to control the quality of the learning process. Our design principle of the hyperparameters is to ensure the three losses should have a similar scale, so that the three loss terms can be balanced and contribute to the learning process.

\begin{figure*}[tb]
 \centering
    \includegraphics[height=.146\textwidth]{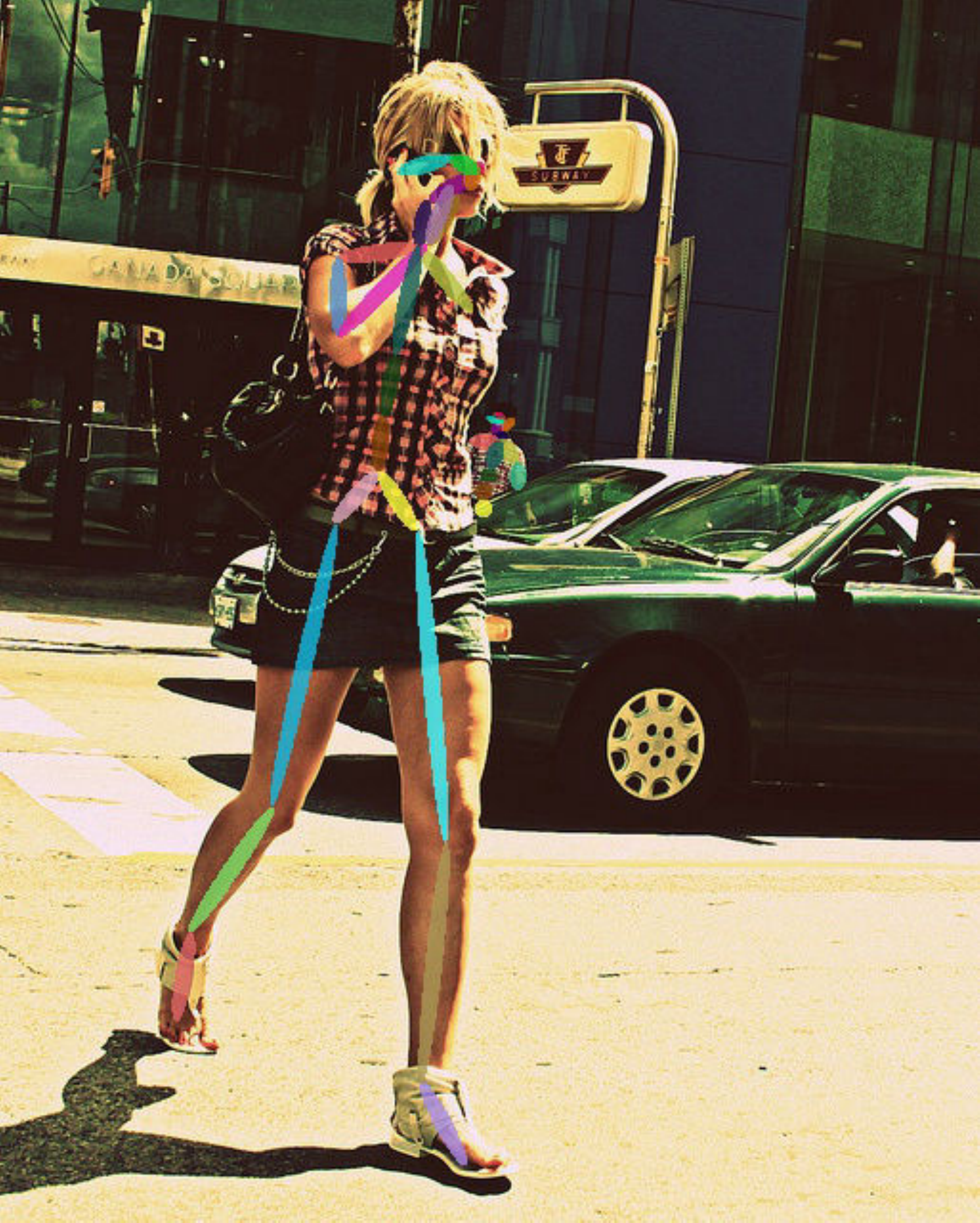}
 	\includegraphics[height=.146\textwidth]{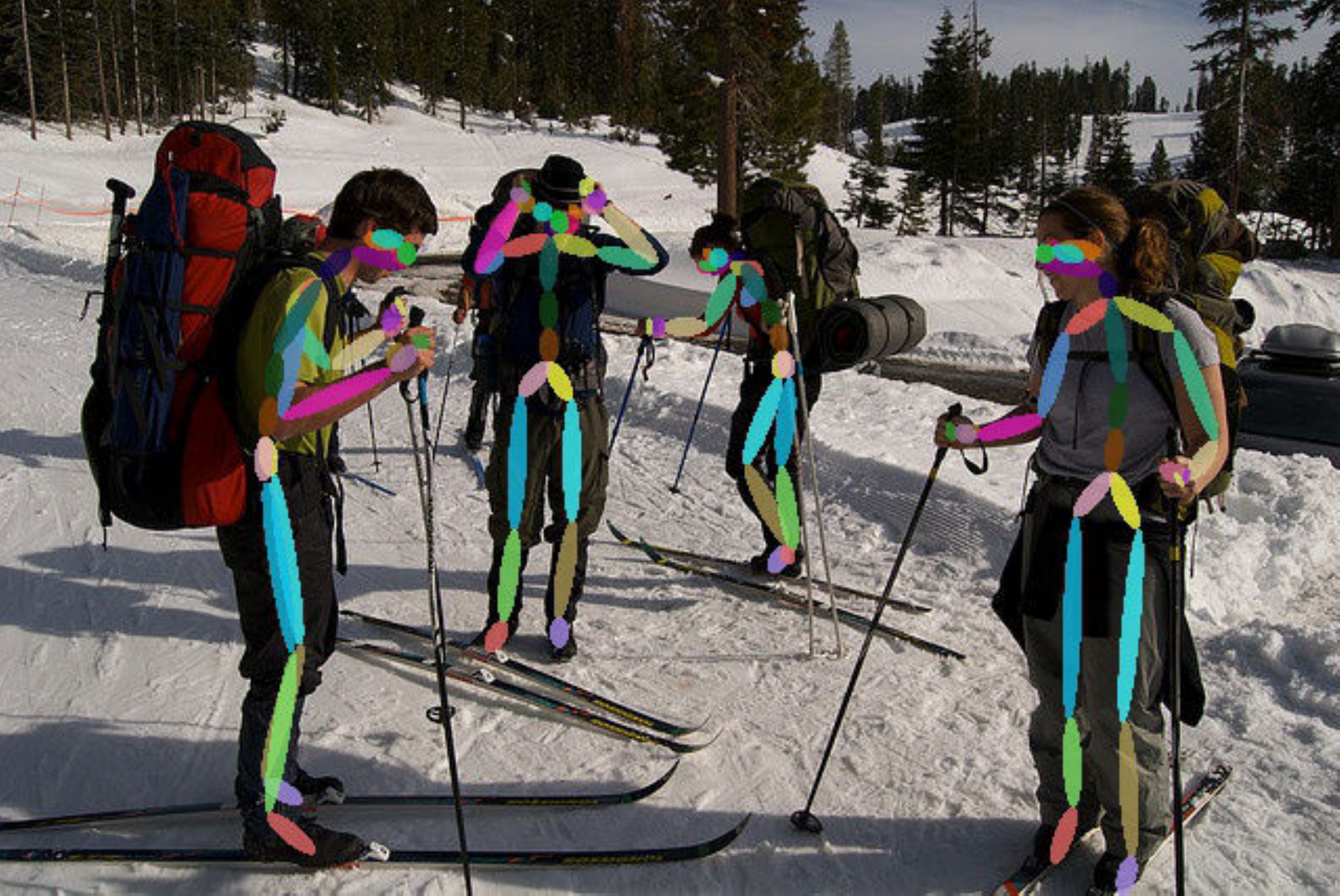}
 	\includegraphics[height=.146\textwidth]{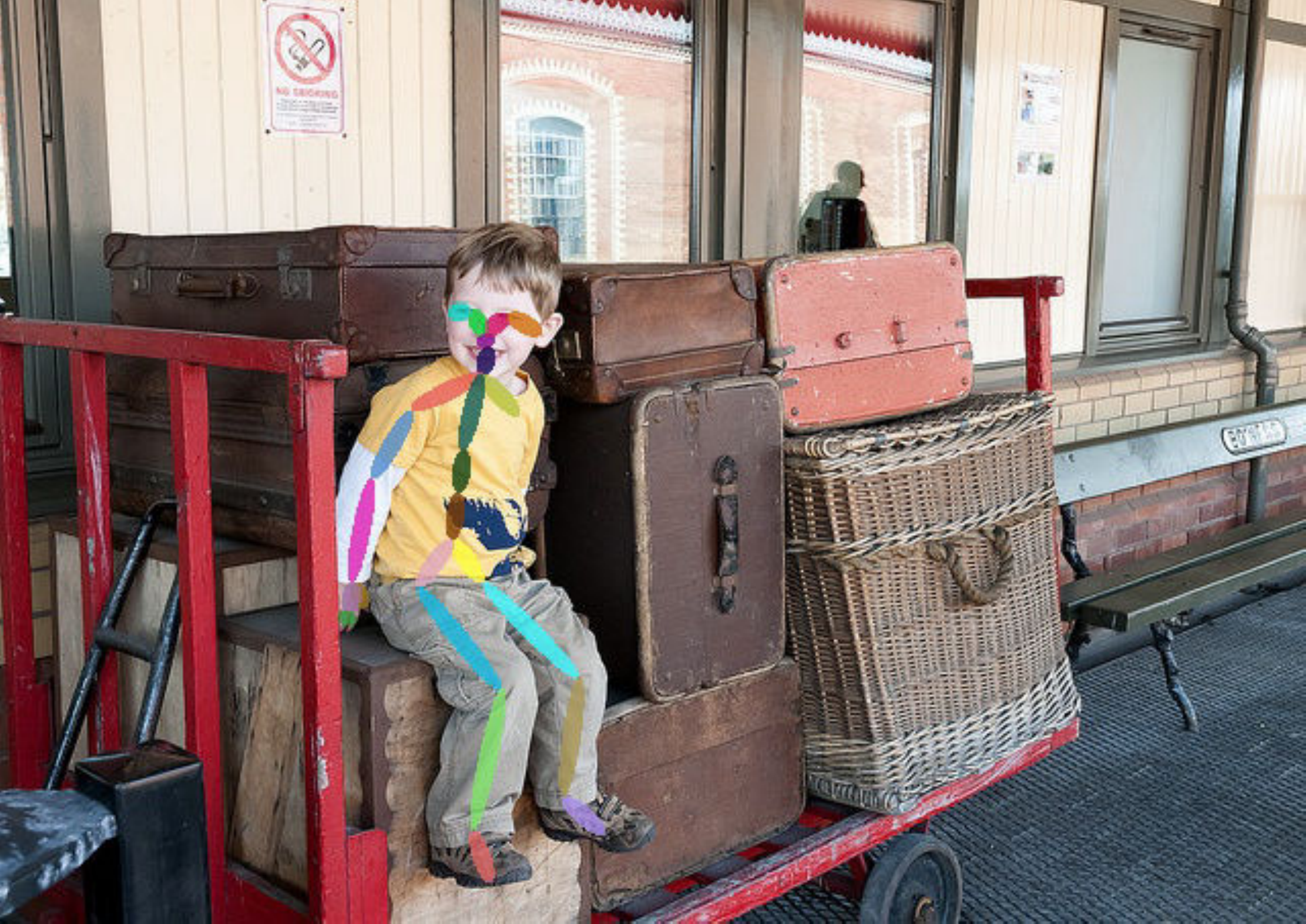}
 	\includegraphics[height=.146\textwidth]{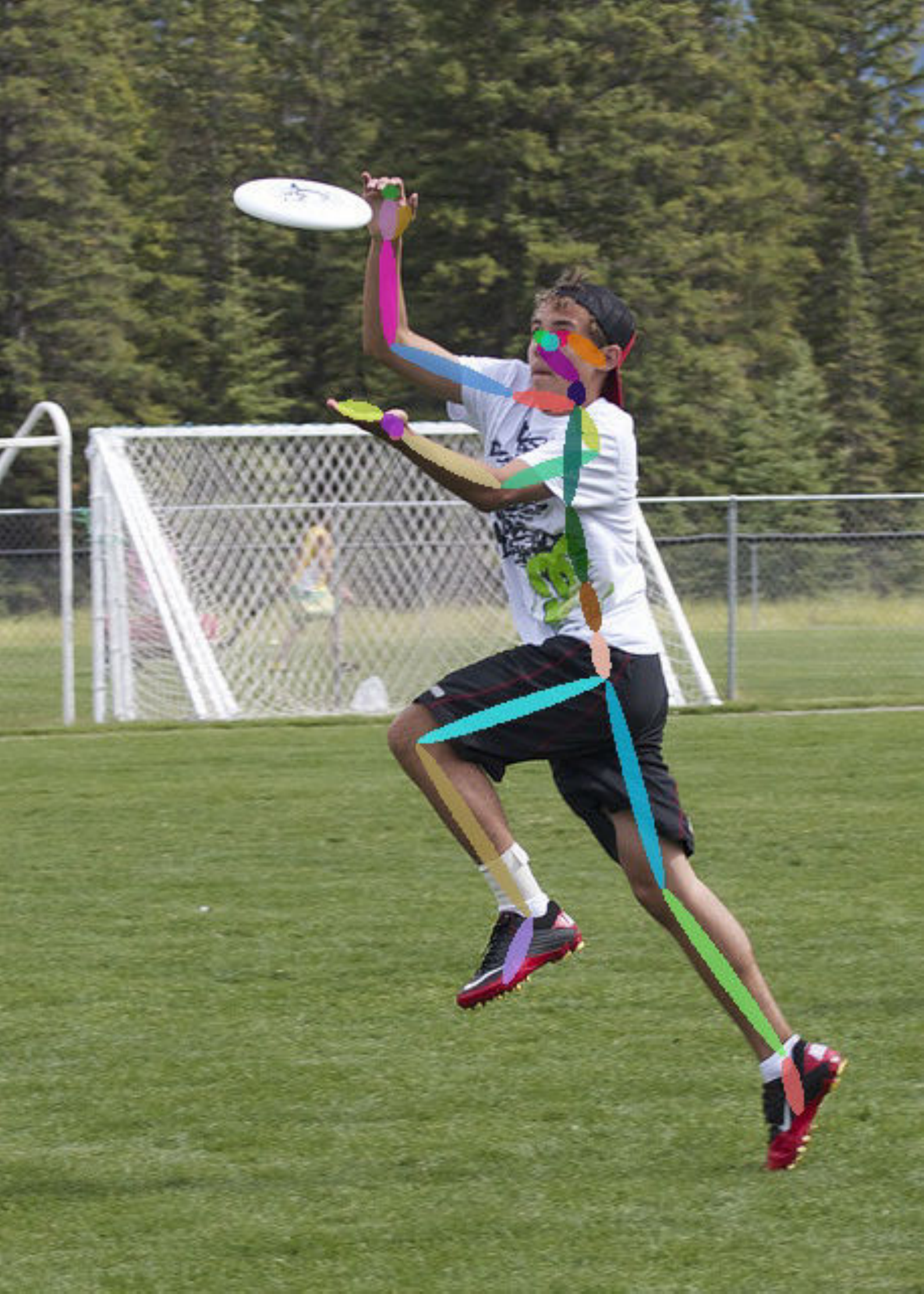}
 	\includegraphics[height=.146\textwidth]{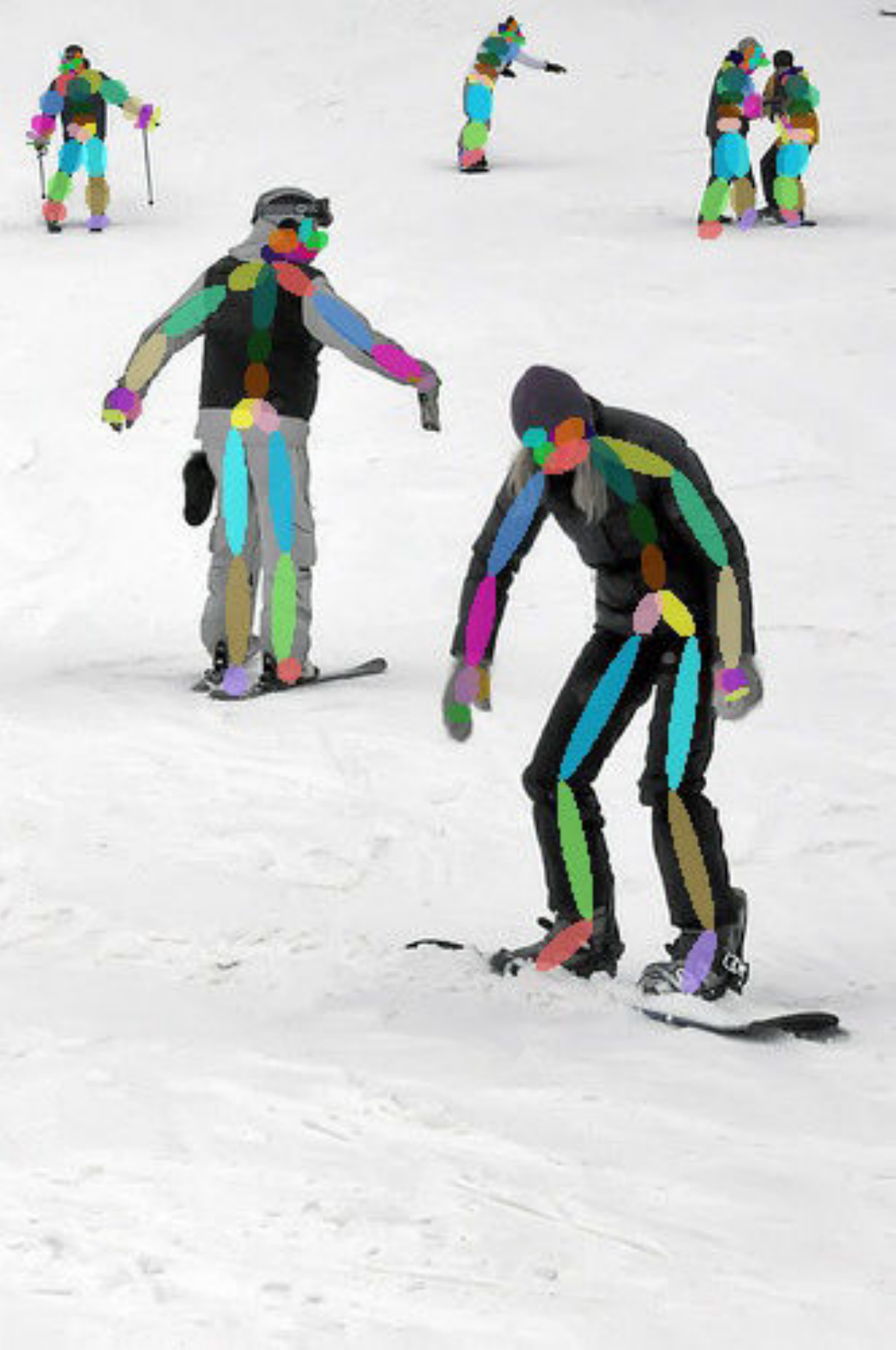}
 	\includegraphics[height=.146\textwidth]{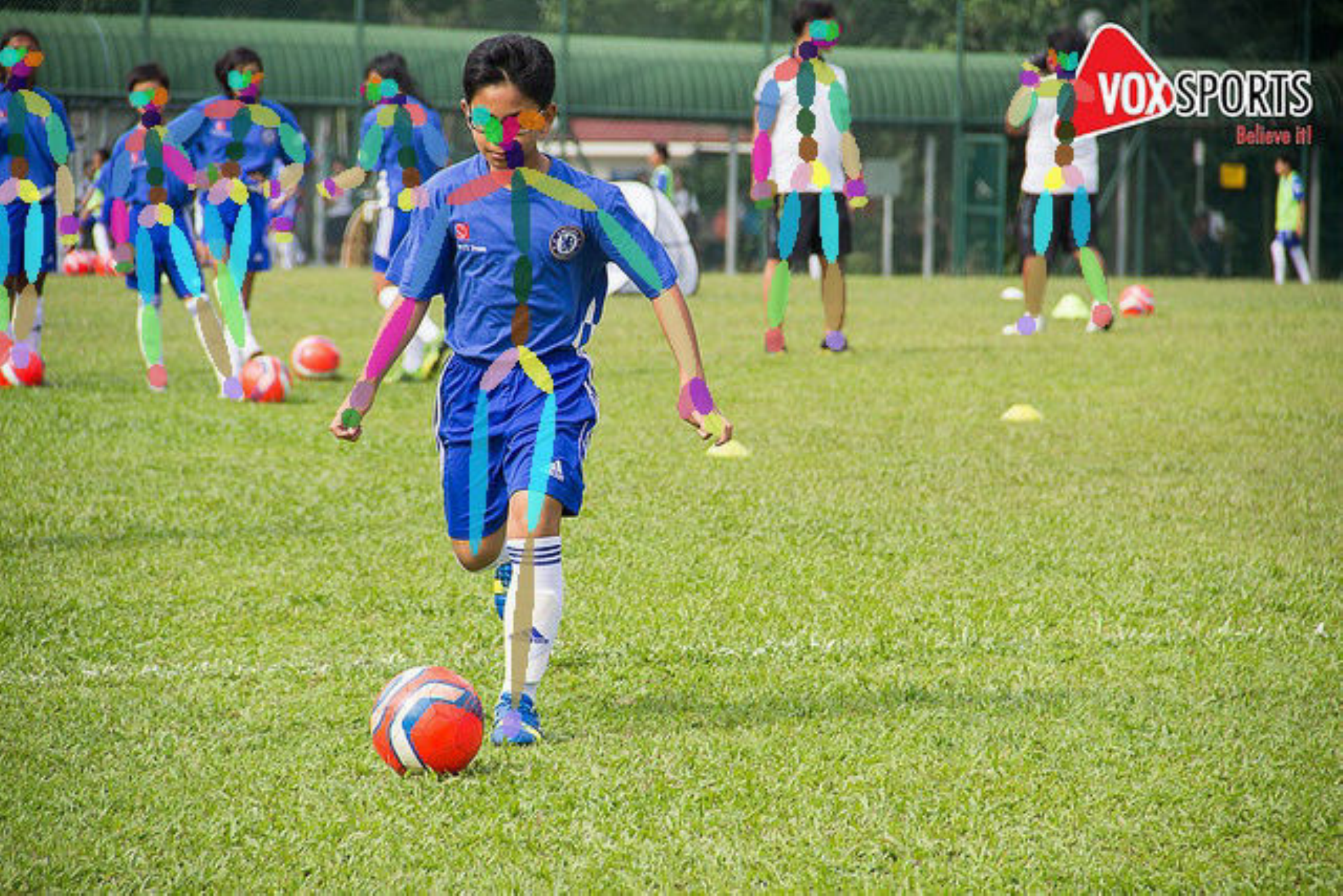}\\
 	\includegraphics[height=.148\textwidth]{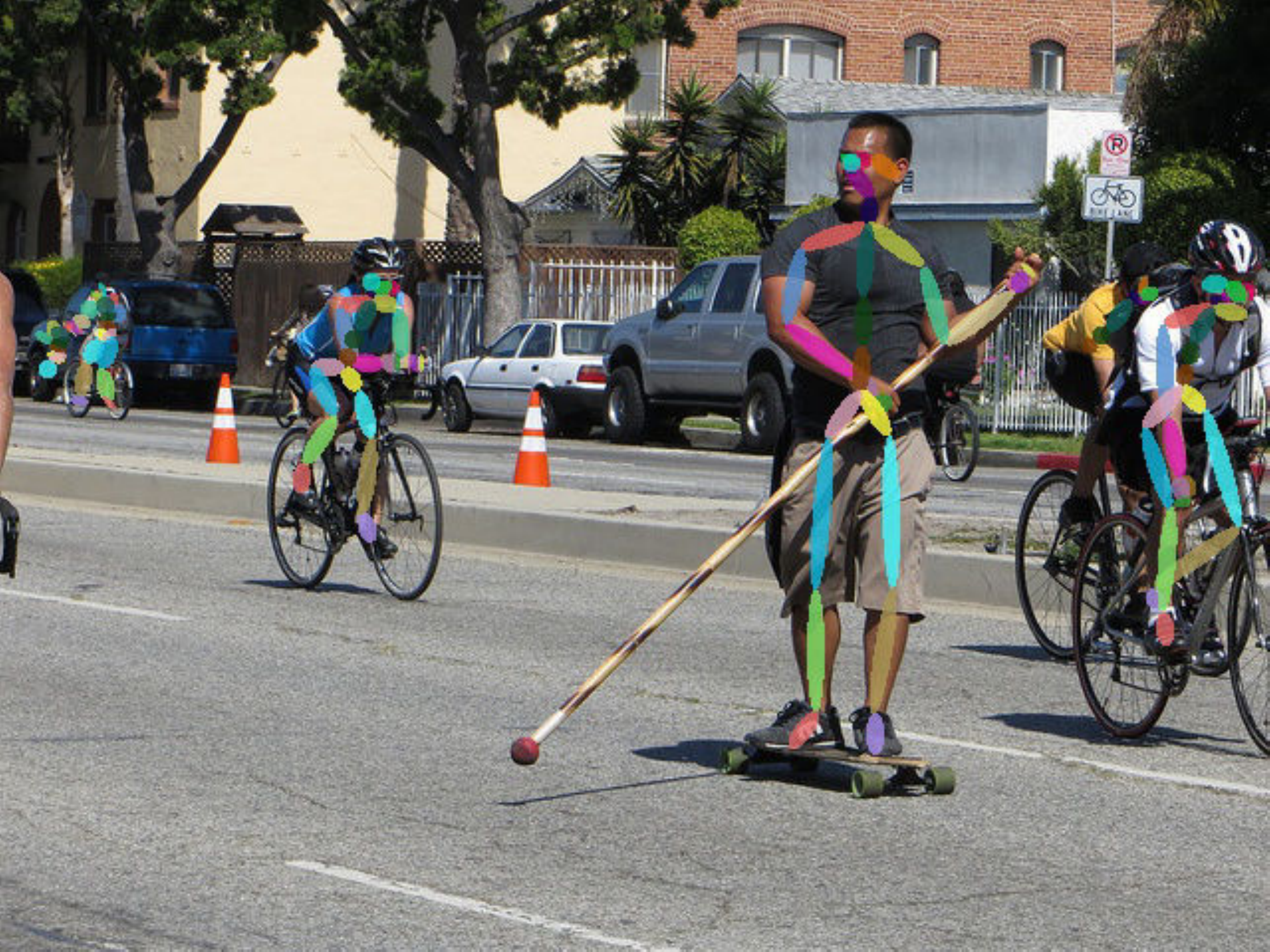}
 	\includegraphics[height=.148\textwidth]{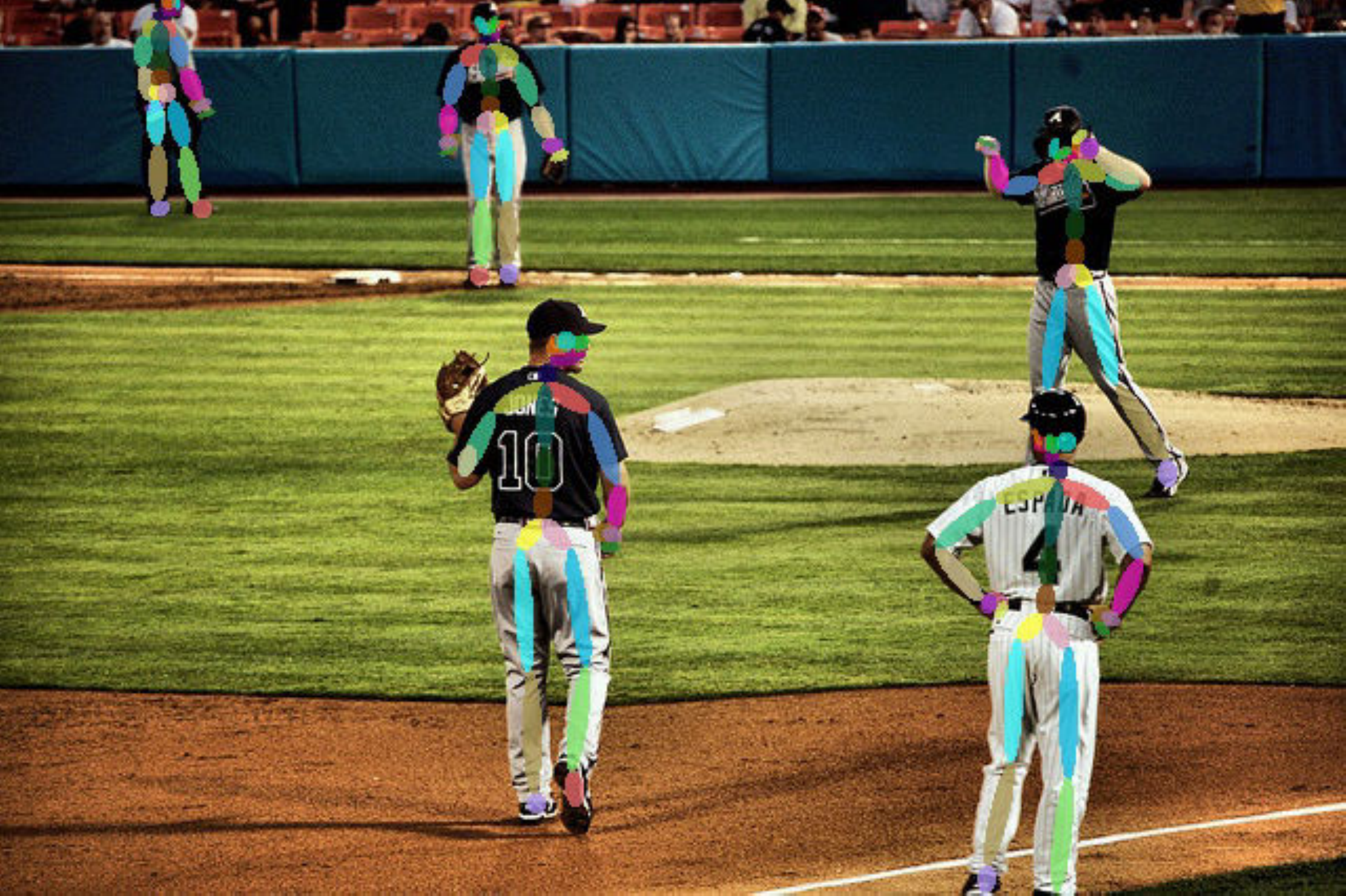}
 	\includegraphics[height=.148\textwidth]{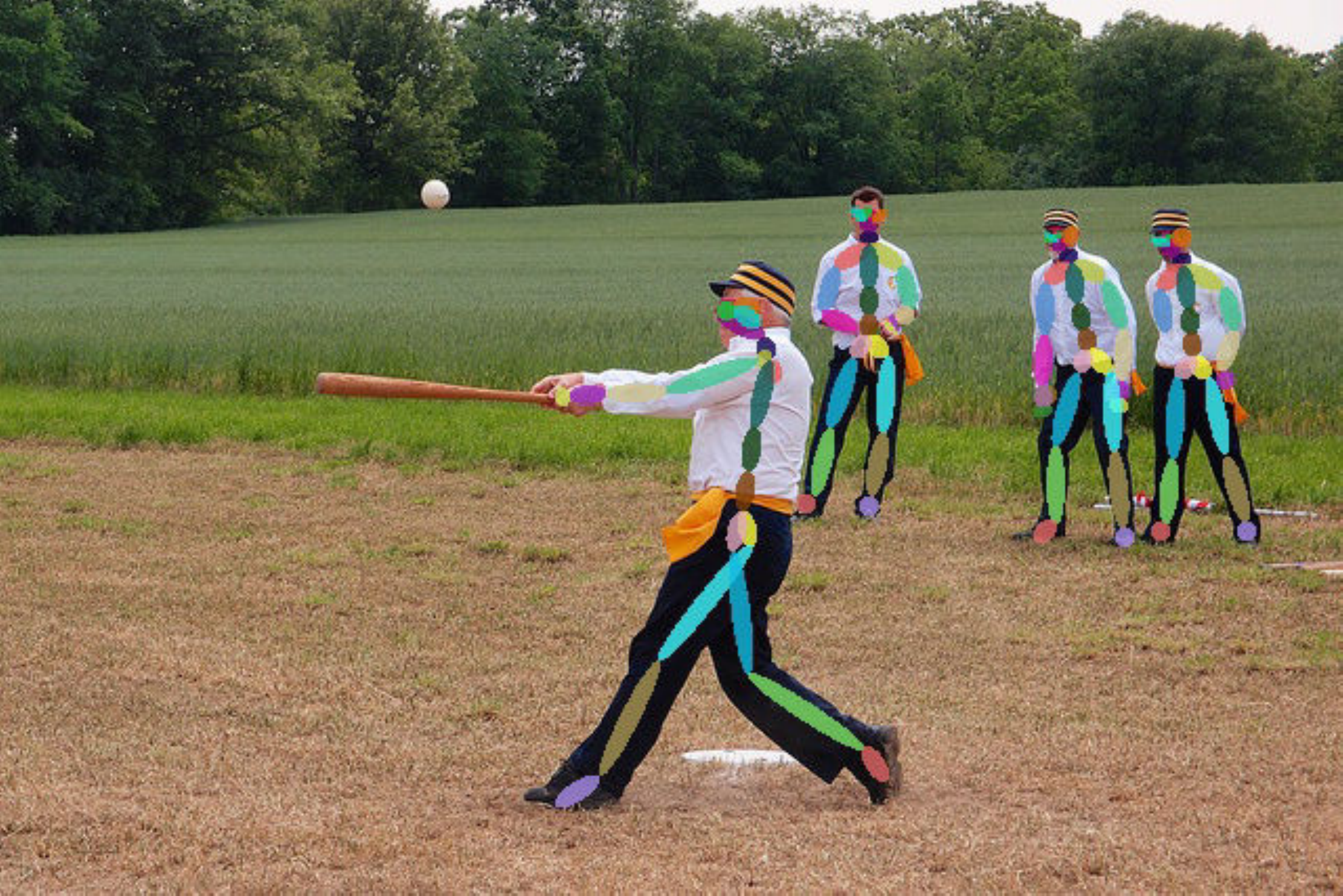}
 	\includegraphics[height=.148\textwidth]{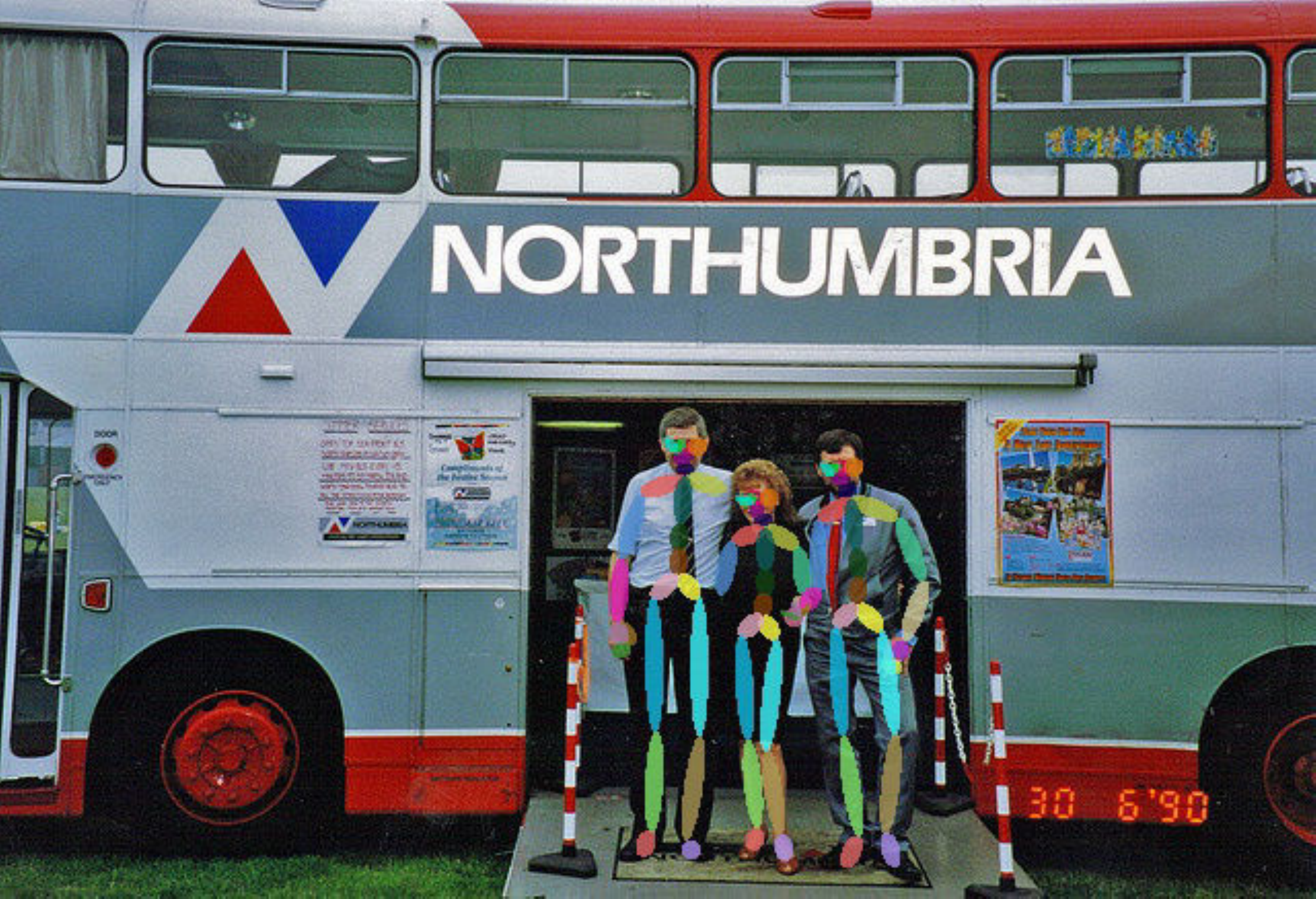}
 	\includegraphics[height=.148\textwidth]{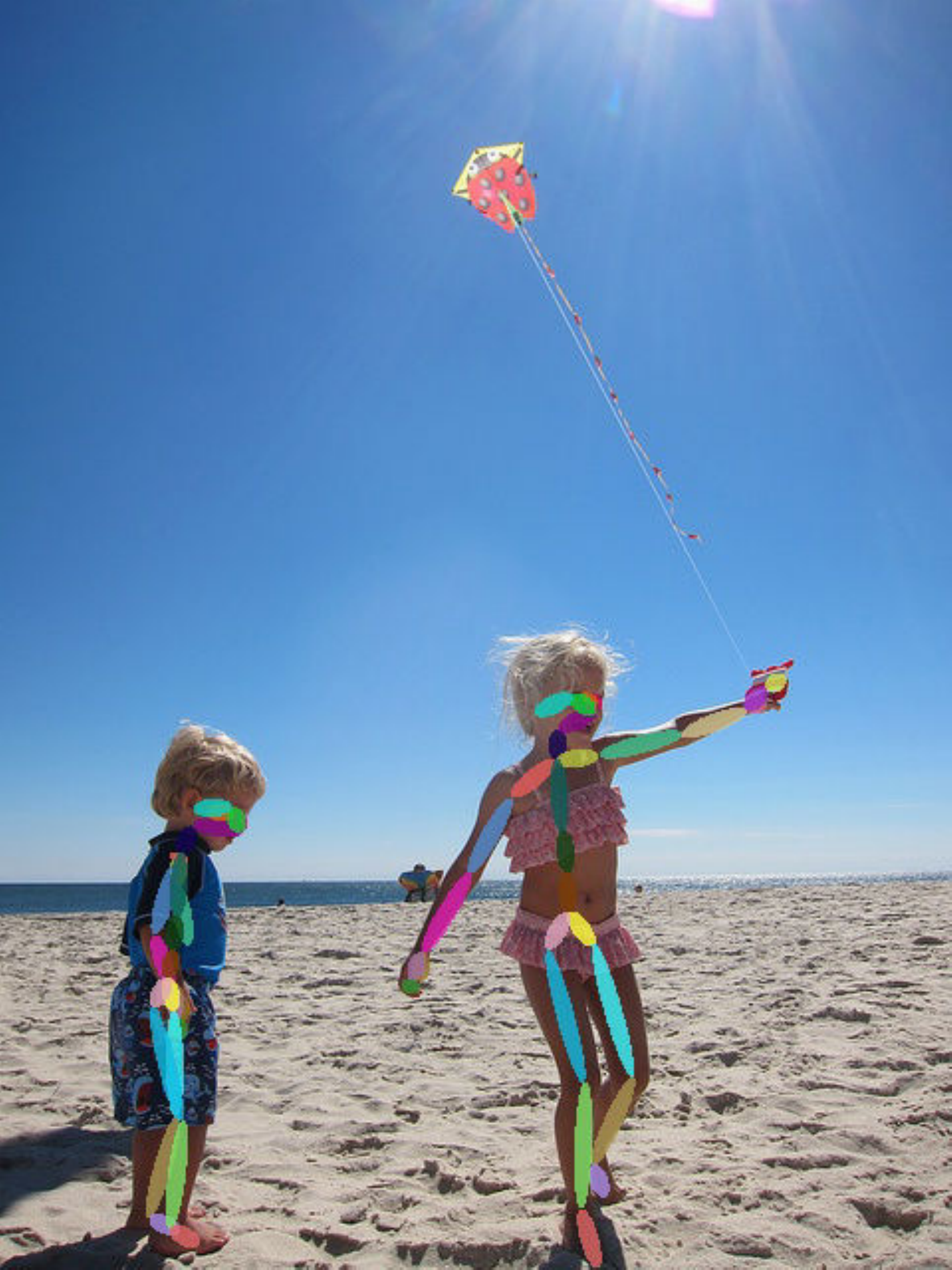}\\
 	\includegraphics[height=.151\textwidth]{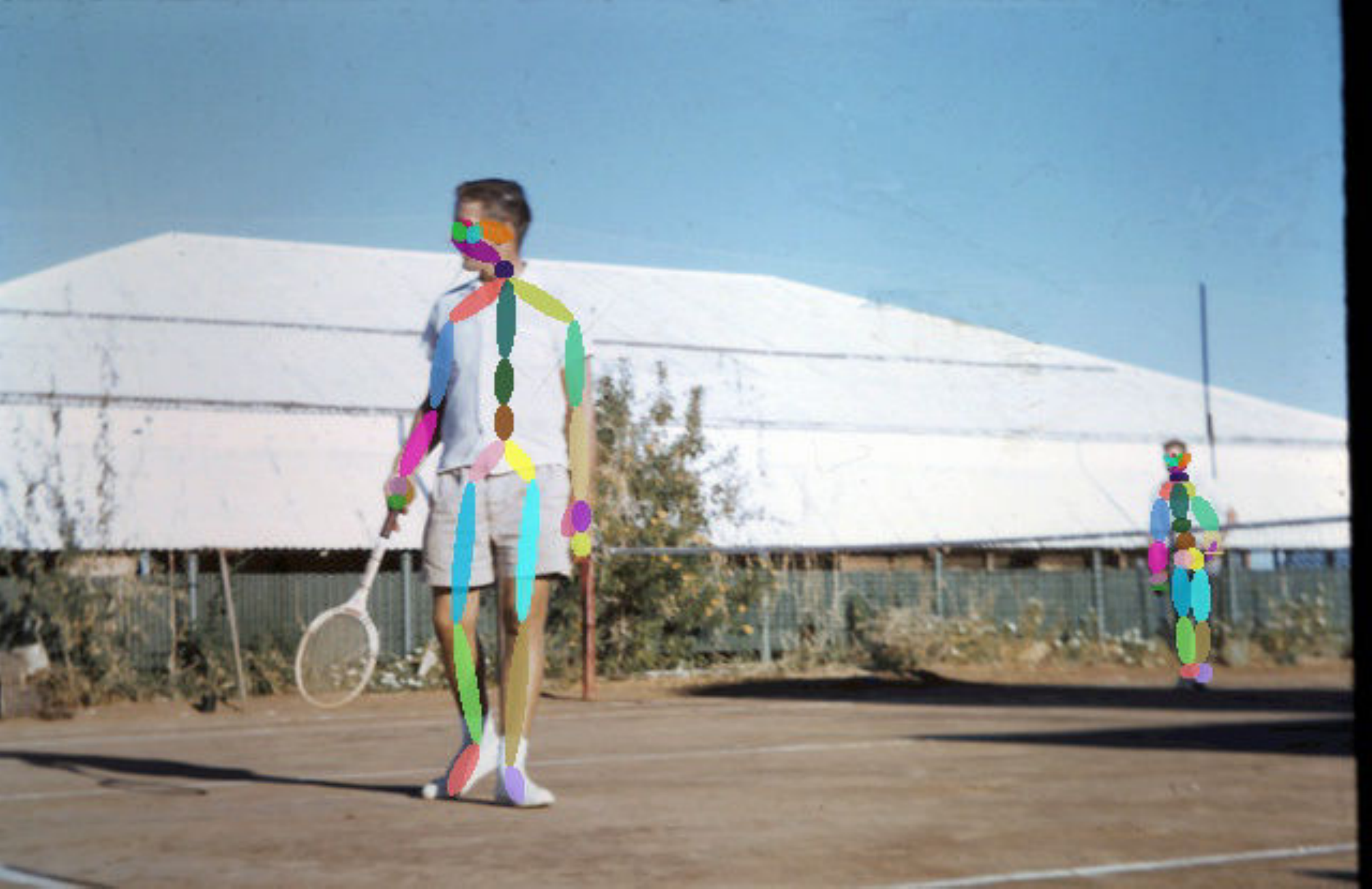}
    \includegraphics[height=.151\textwidth]{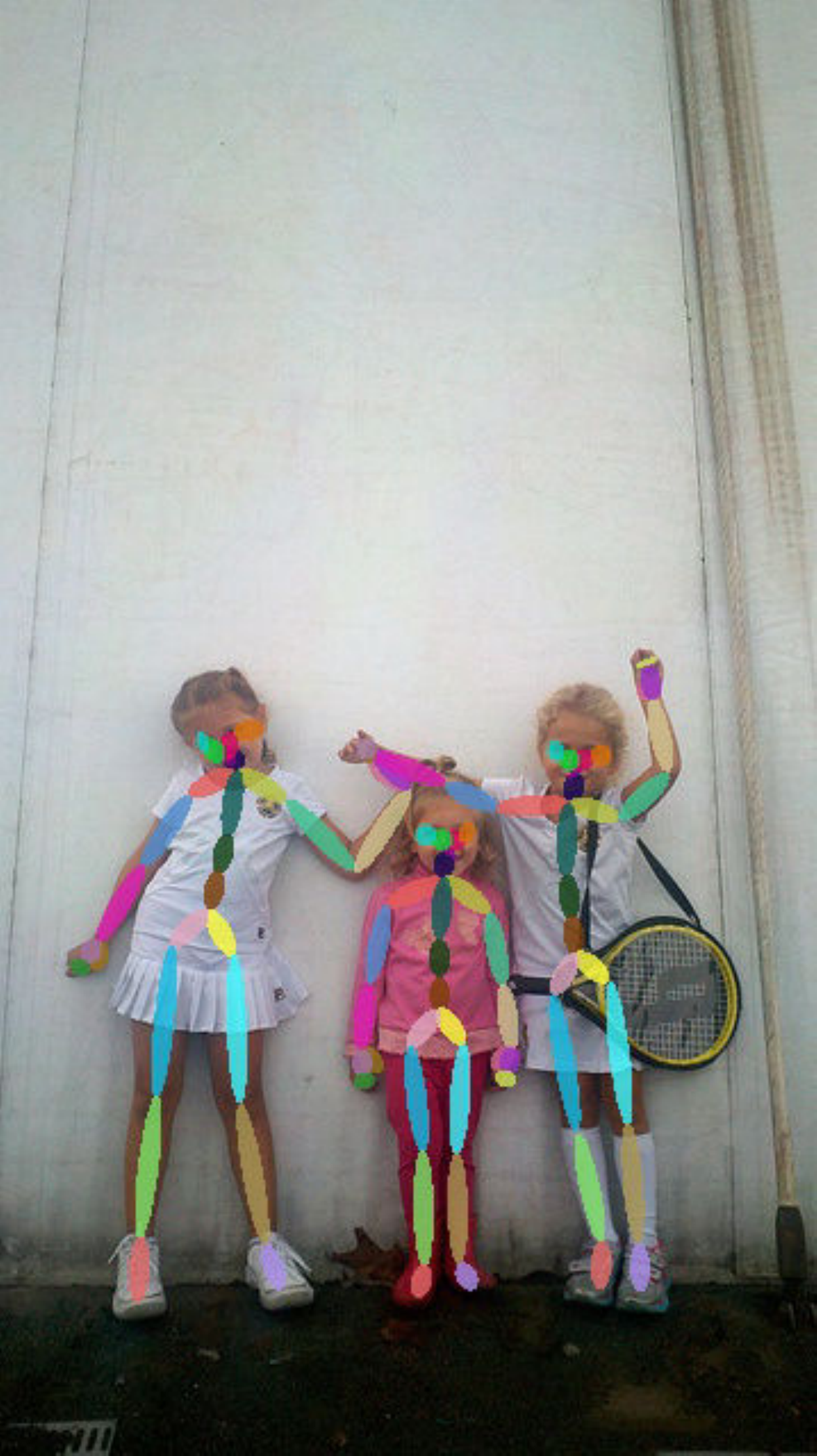}
 	\includegraphics[height=.151\textwidth]{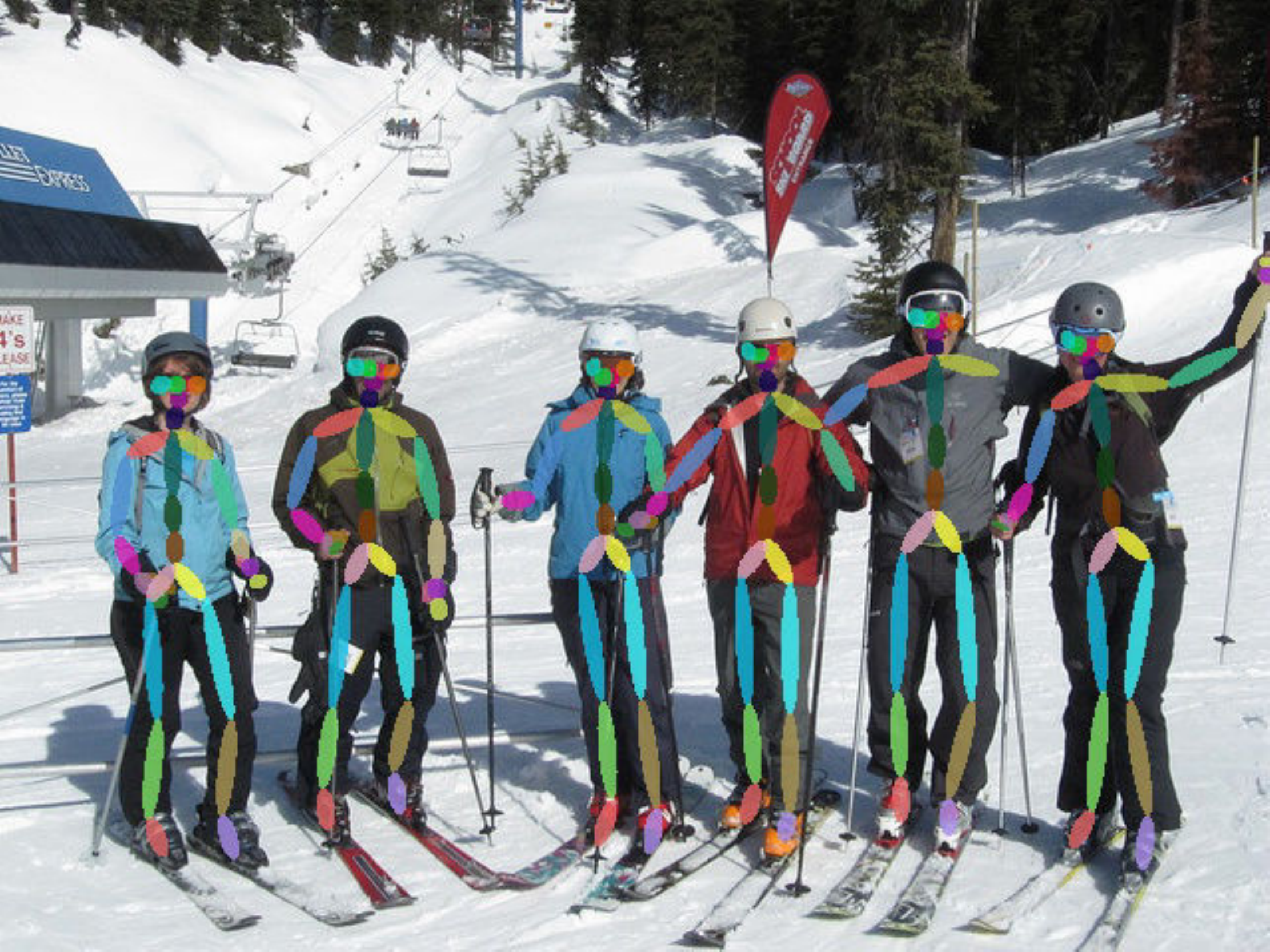}
 	\includegraphics[height=.151\textwidth]{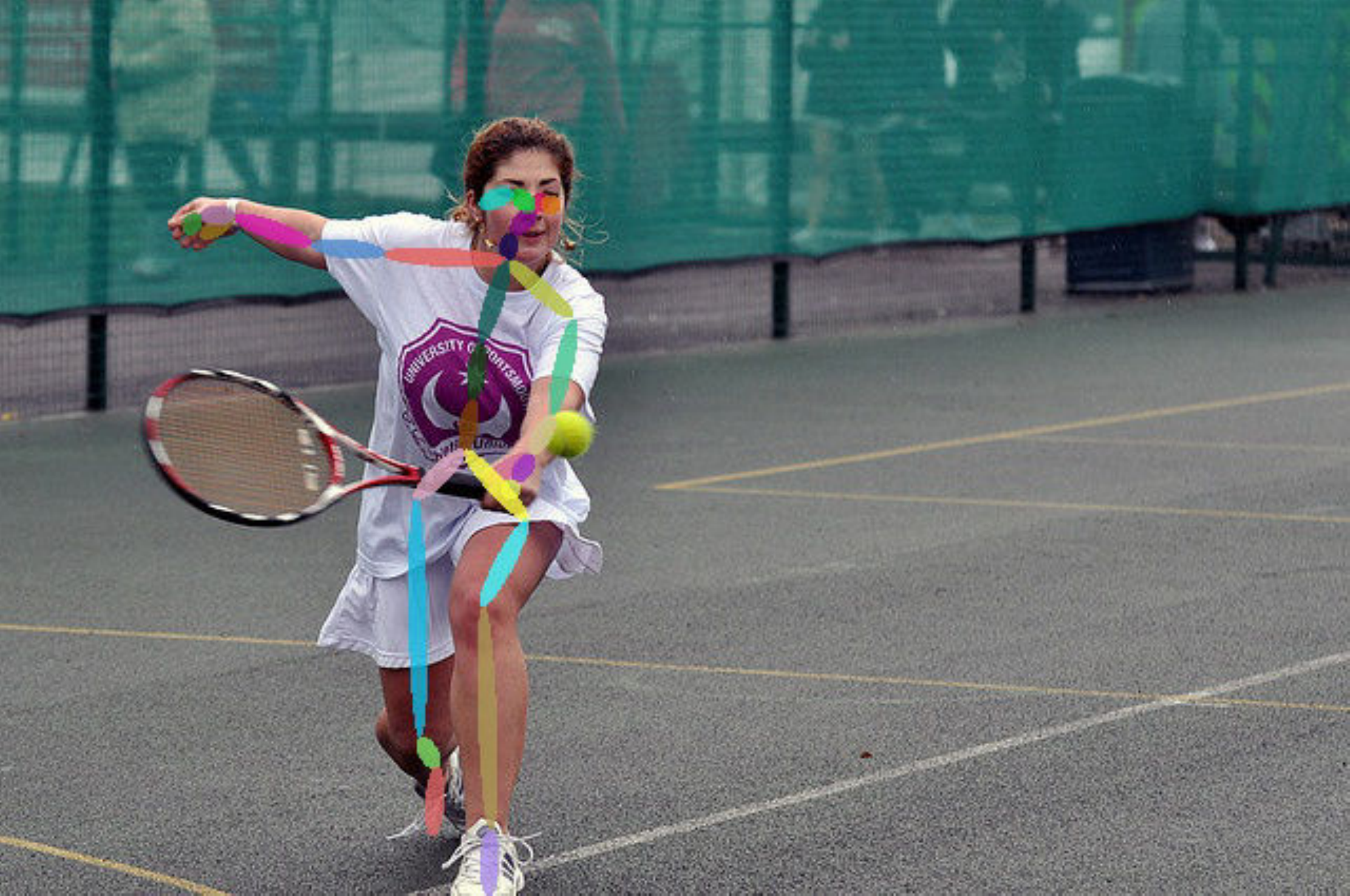}
 	\includegraphics[height=.151\textwidth]{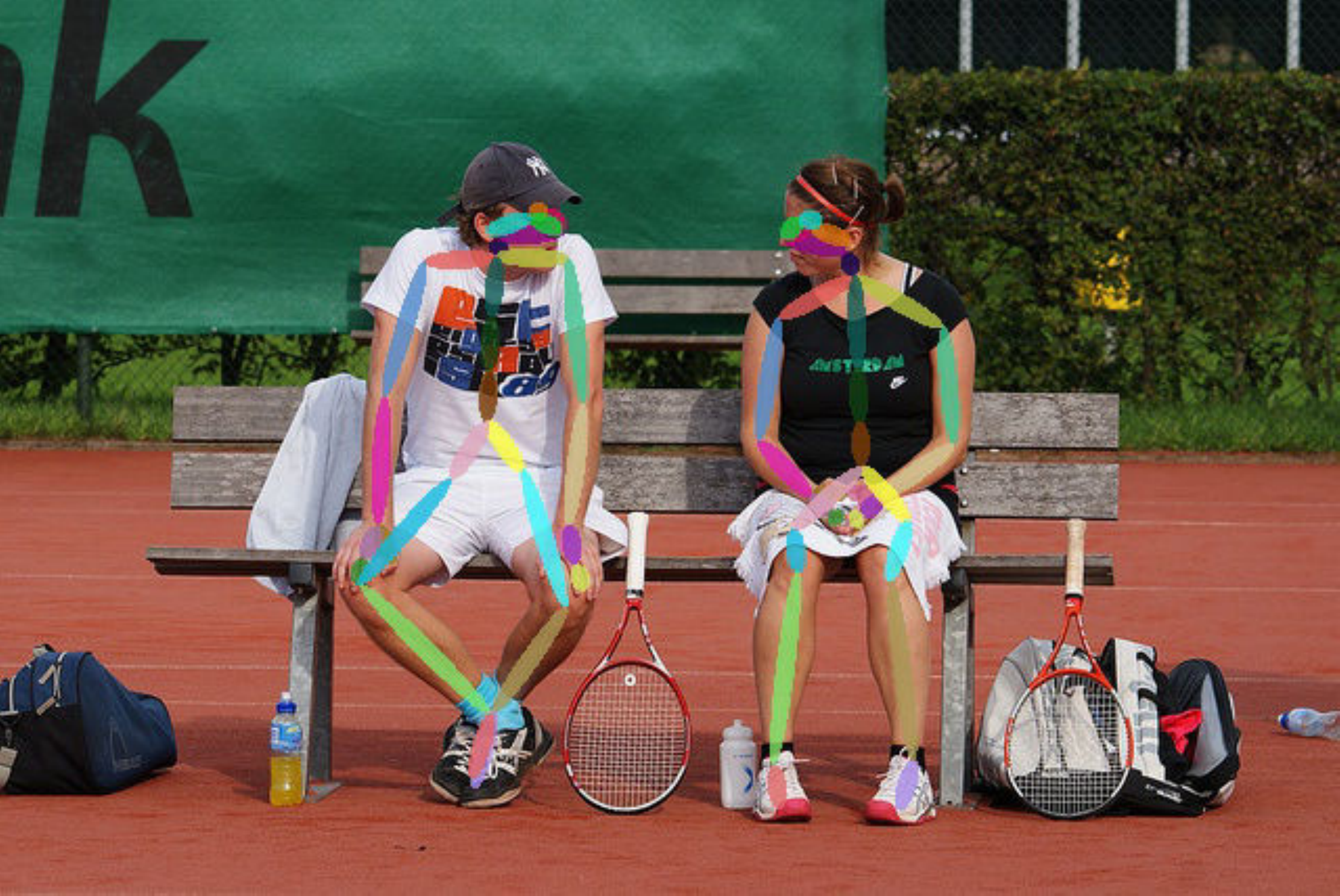}\\
    %
 \caption{Novel keypoint detection results on COCO validation images. Without any human labeling effort, our method learns to predict a new set of keypoints including those on the hands and feet.}
\label{fig:addition-kpts} 
\end{figure*}

\subsection{Novel keypoint detection}
\label{sec:novelkey}
Since our approach can easily create arbitrary annotations on synthetic data and transfer the knowledge to real domain, our method is highly scalable and flexible to users needs. For example, suppose we want to predict a new set of keypoints including hands and feet, it would be difficult to re-label the entire COCO dataset. With our technique, we can simply generate new labels on the synthetic data. We have performed an experiment to demonstrate this capability.

\begin{figure}
	\centering
	\includegraphics[height=.27\textwidth]{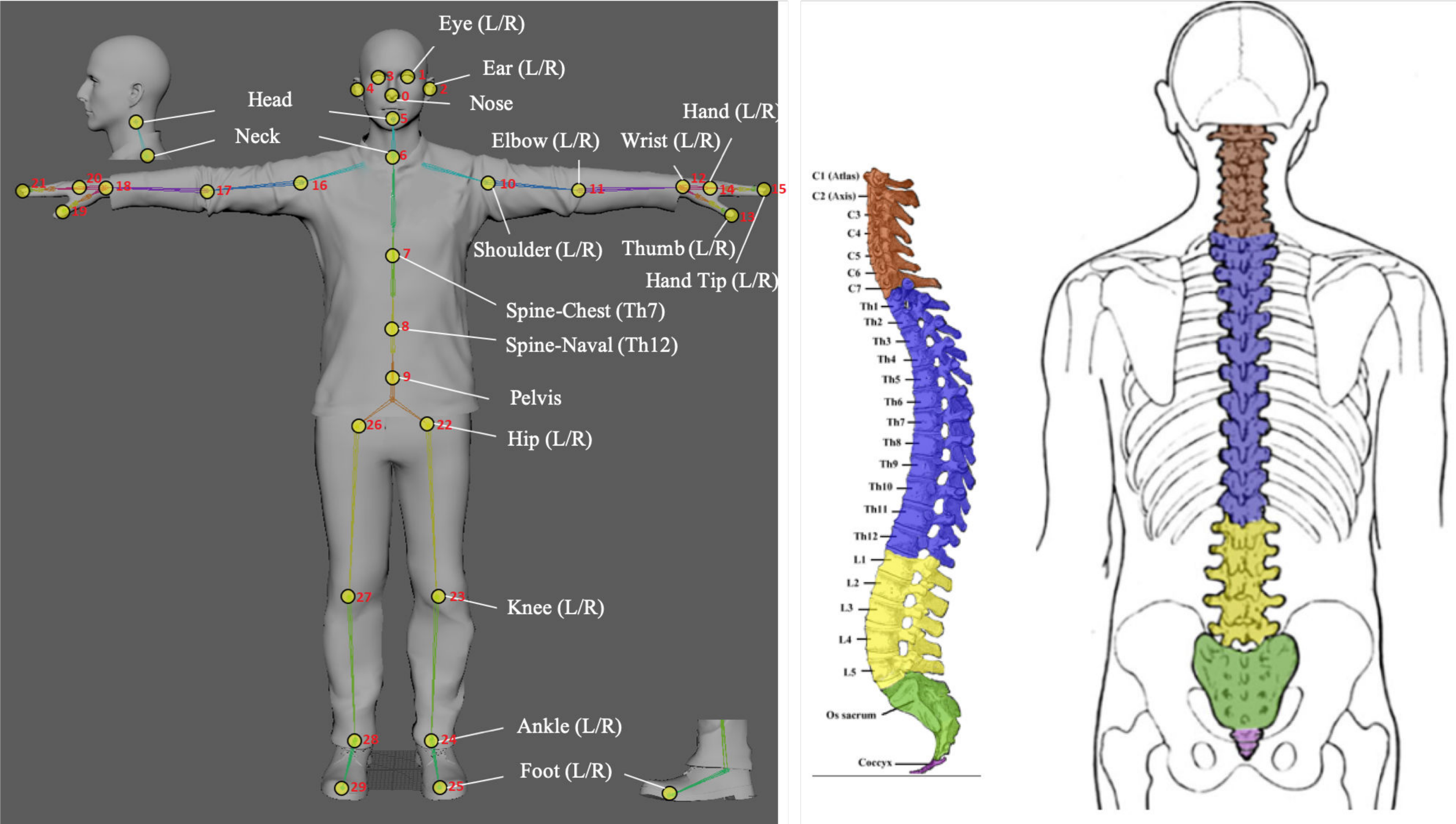}
	\caption{The definition of our created novel keypoints. There is a total of $30$ keypoints and $29$ part associations for constructing fine-grained human skeleton.}
	\label{fig:joint-define}
\end{figure}

\begin{figure*}[h]
	 \centering
 \setlength{\tabcolsep}{4pt}
 \begin{tabular}{cc}
    \includegraphics[width=1\columnwidth]{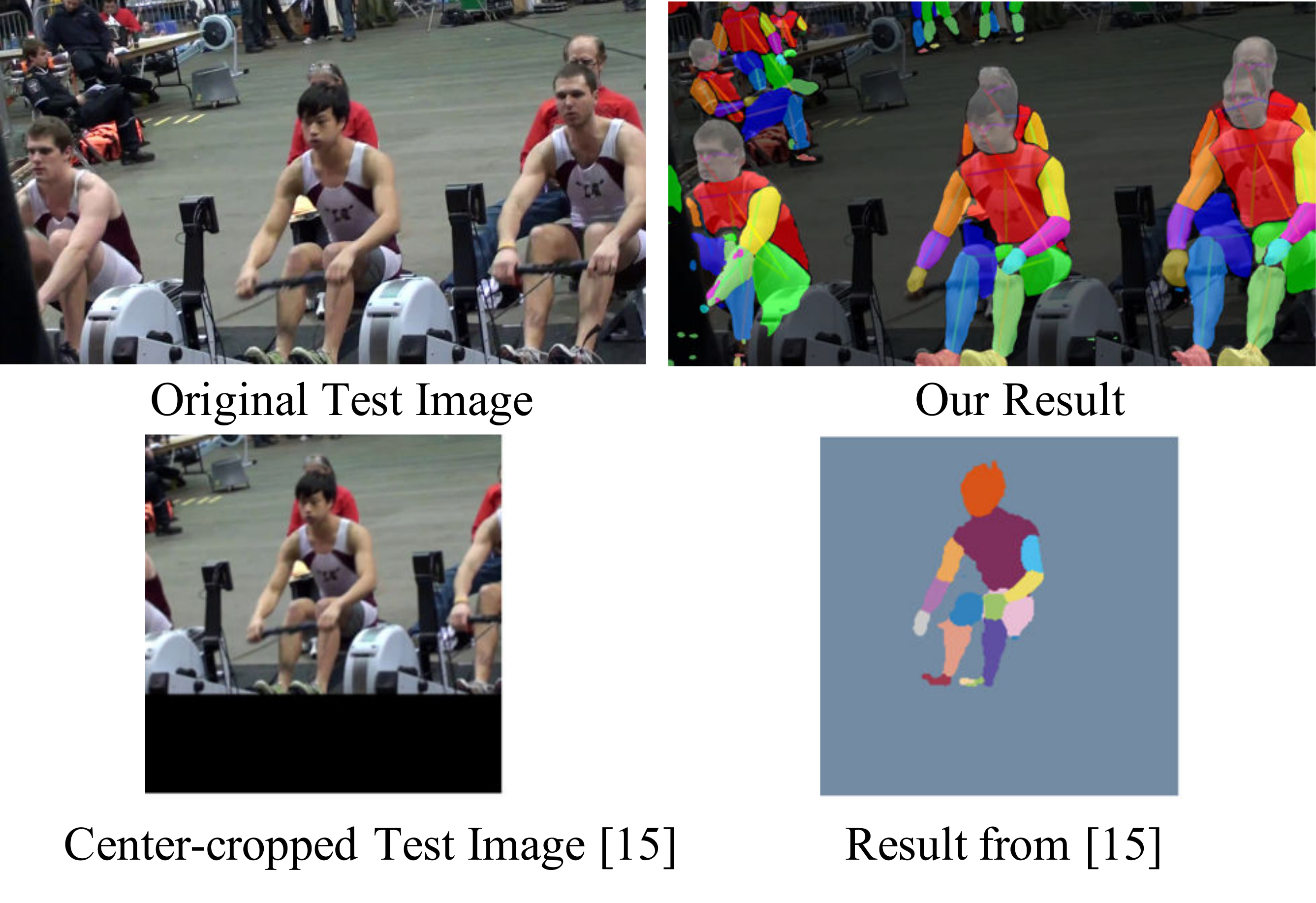}
    &\includegraphics[width=1\columnwidth]{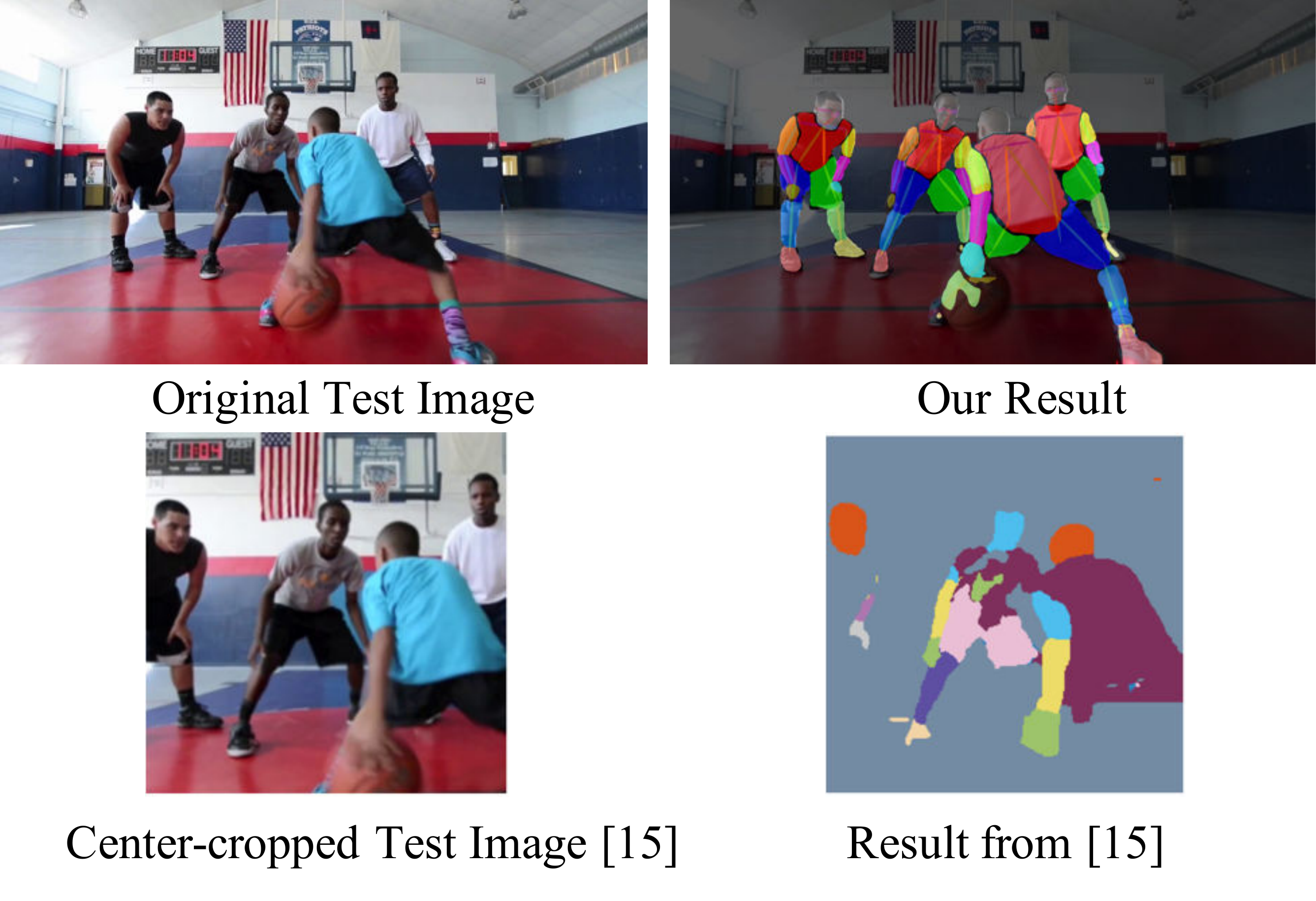}\\
 	(a) & (b)\\
    \end{tabular}
	\caption{Qualitative comparison with the state-of-the-art approach~\cite{varol2017learning}. Previous method required additional preprocessing to normalize the input image, and failed to predict part segmentation for all the people due to occlusions. In contrast, our method does not require any preprocessing, and generated correct part segmentation for all the people even though some of them are heavily occluded by others.}
	\label{fig:demo}
\end{figure*}

\begin{figure*}
 \centering
 \setlength{\tabcolsep}{0.5pt}
 \begin{tabular}{cccc}
  	\includegraphics[width=.245\textwidth]{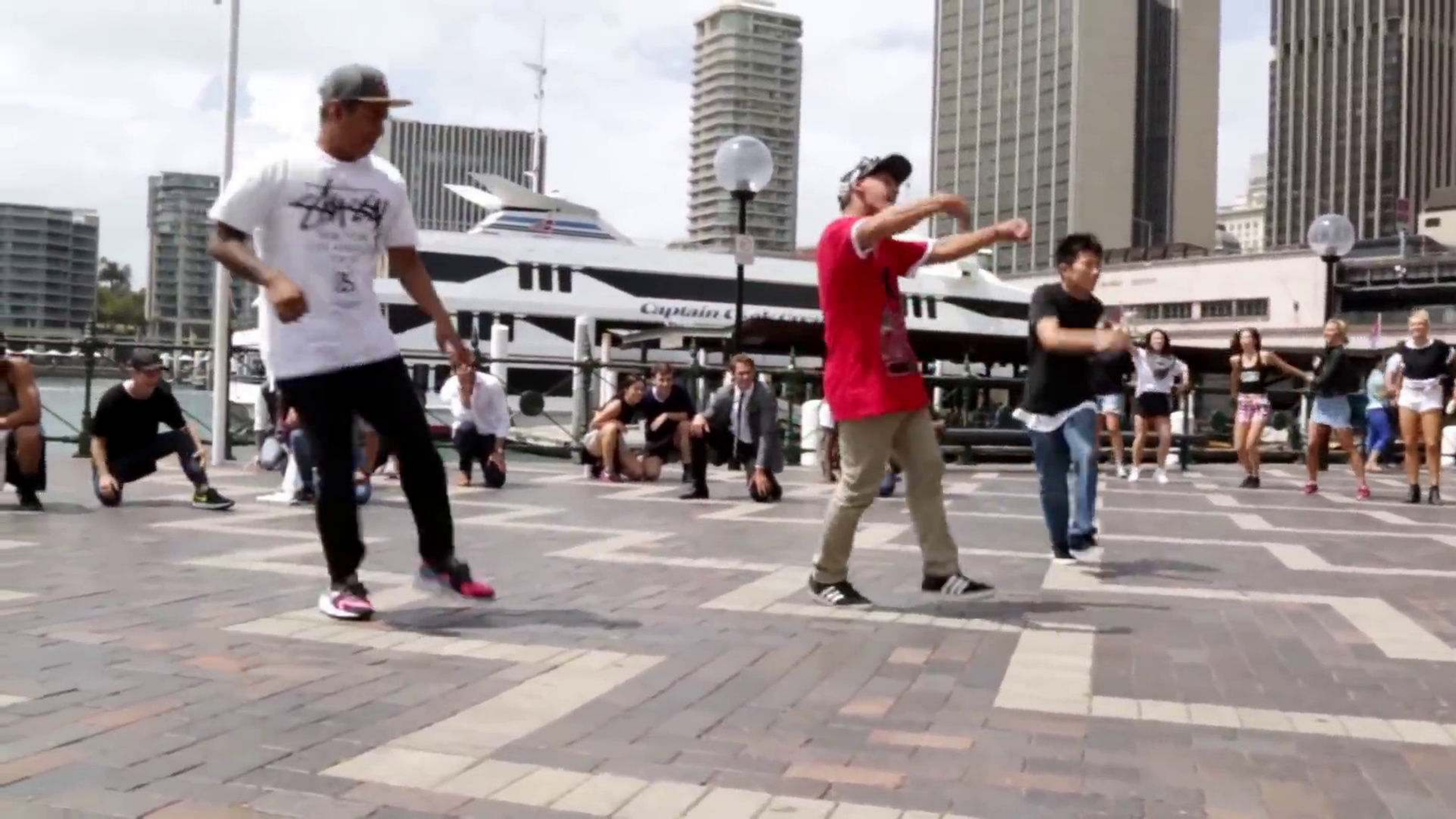}
 	&\includegraphics[width=.245\textwidth]{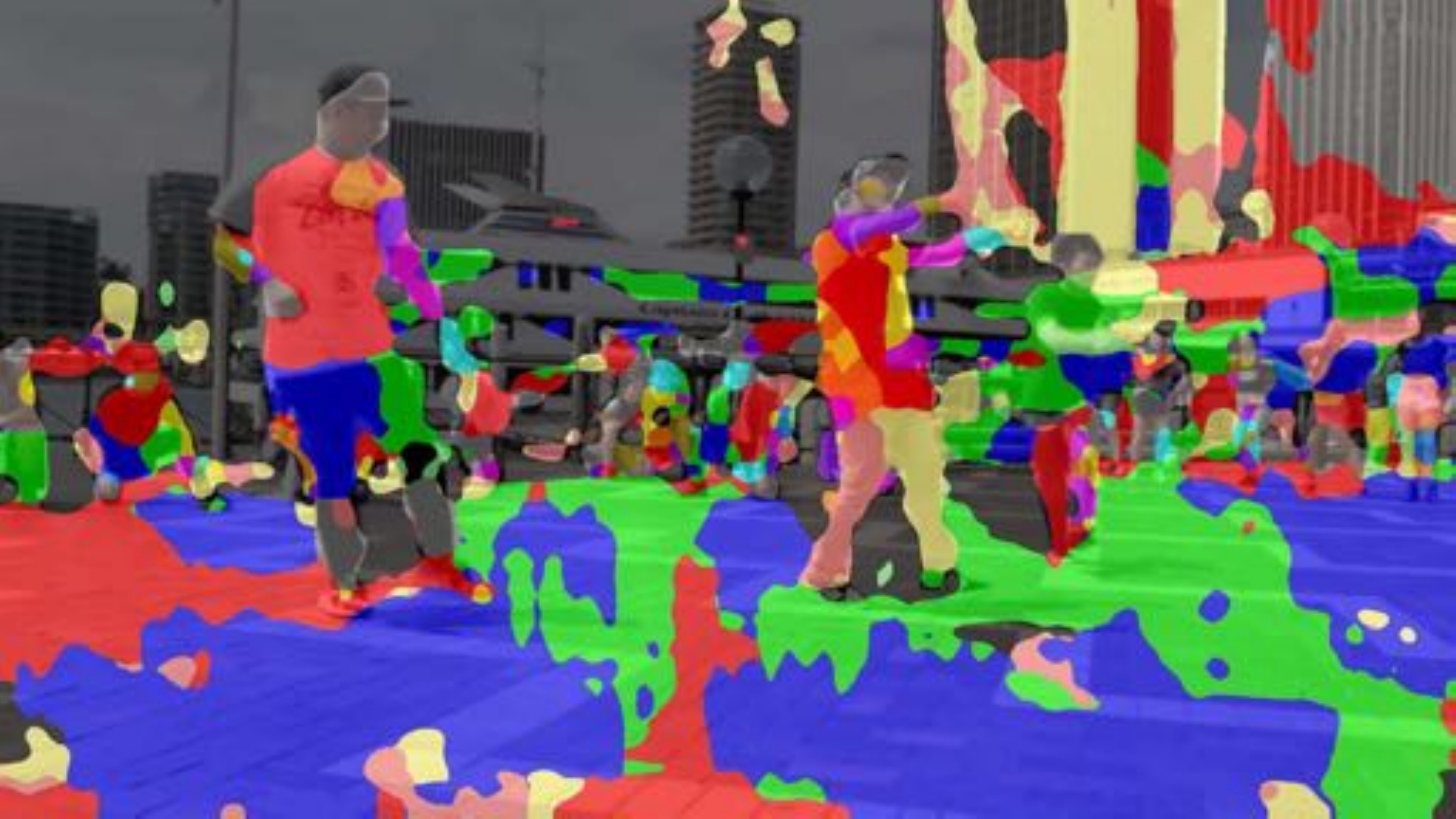}
    &\includegraphics[width=.245\textwidth]{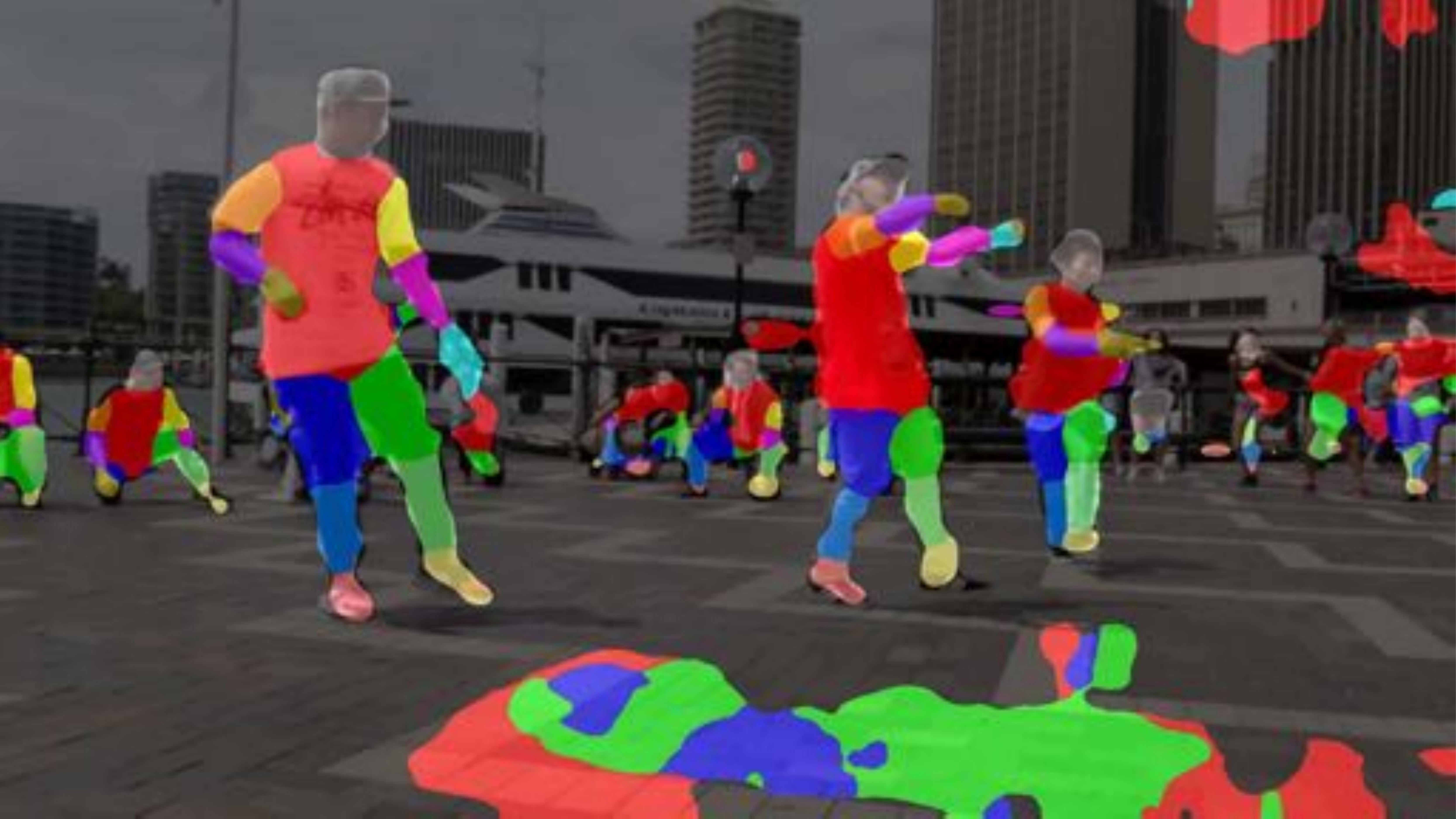}
 	&\includegraphics[width=.245\textwidth]{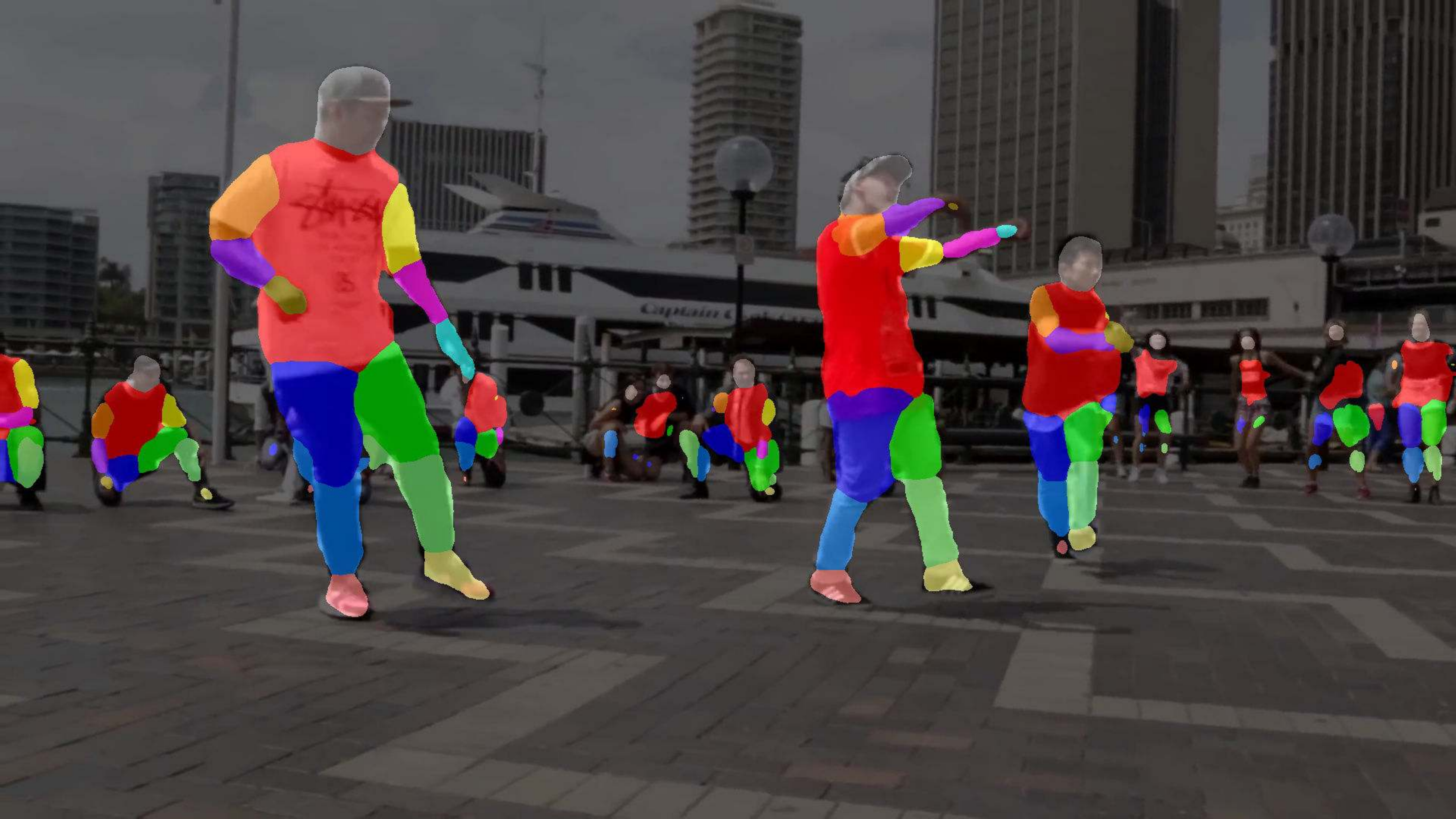}\\
  	\includegraphics[width=.245\textwidth]{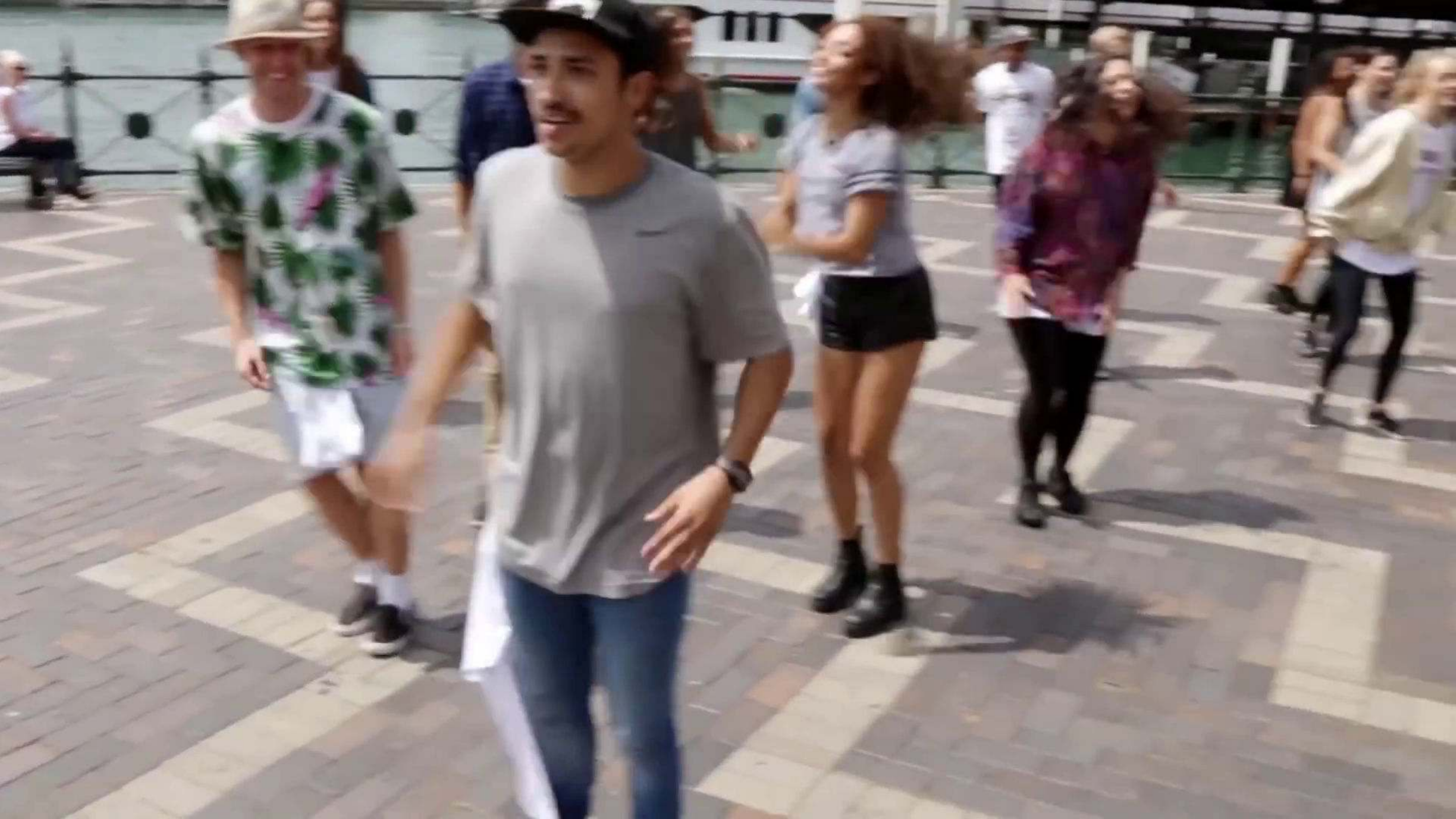}
 	&\includegraphics[width=.245\textwidth]{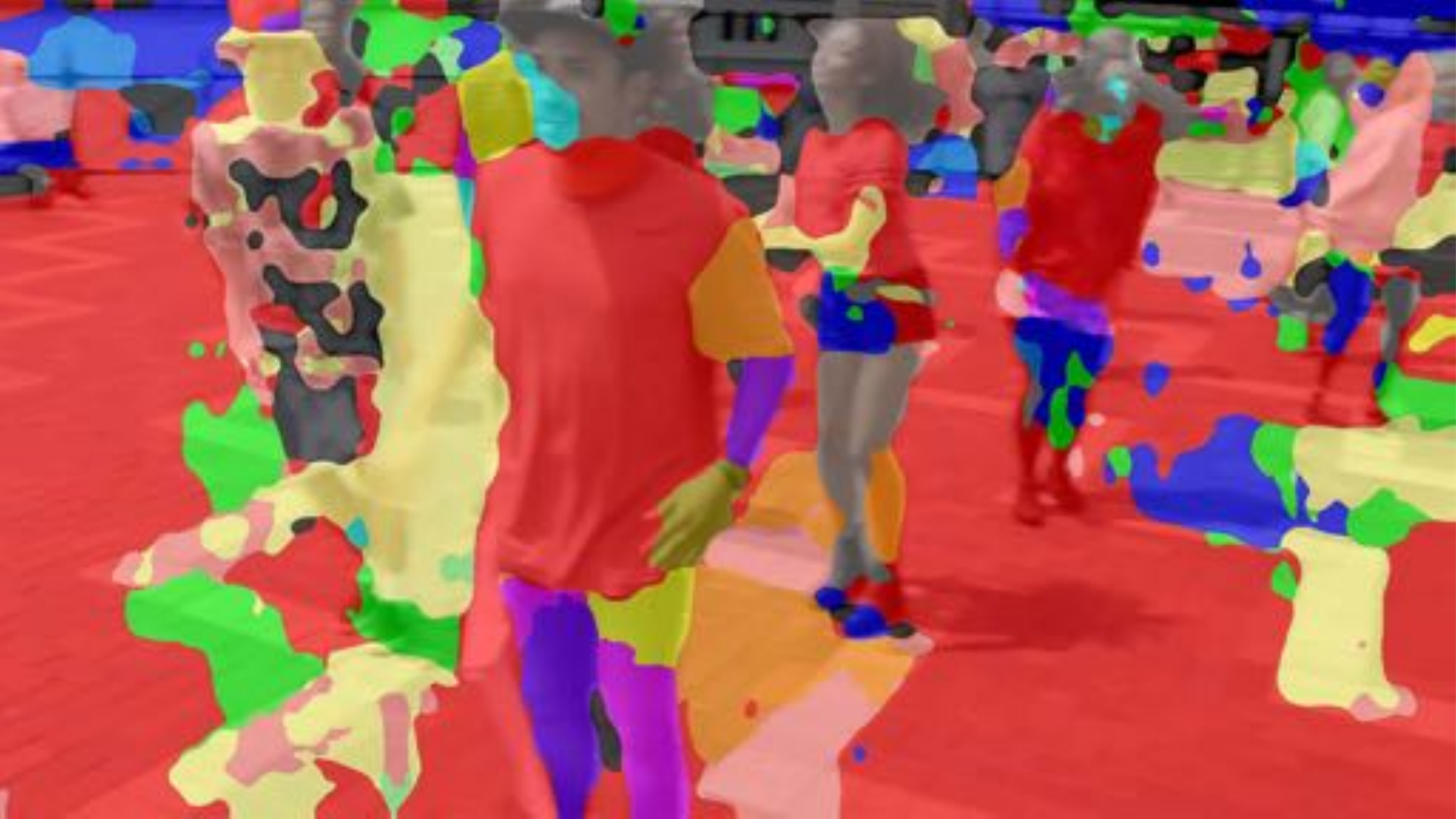}
    &\includegraphics[width=.245\textwidth]{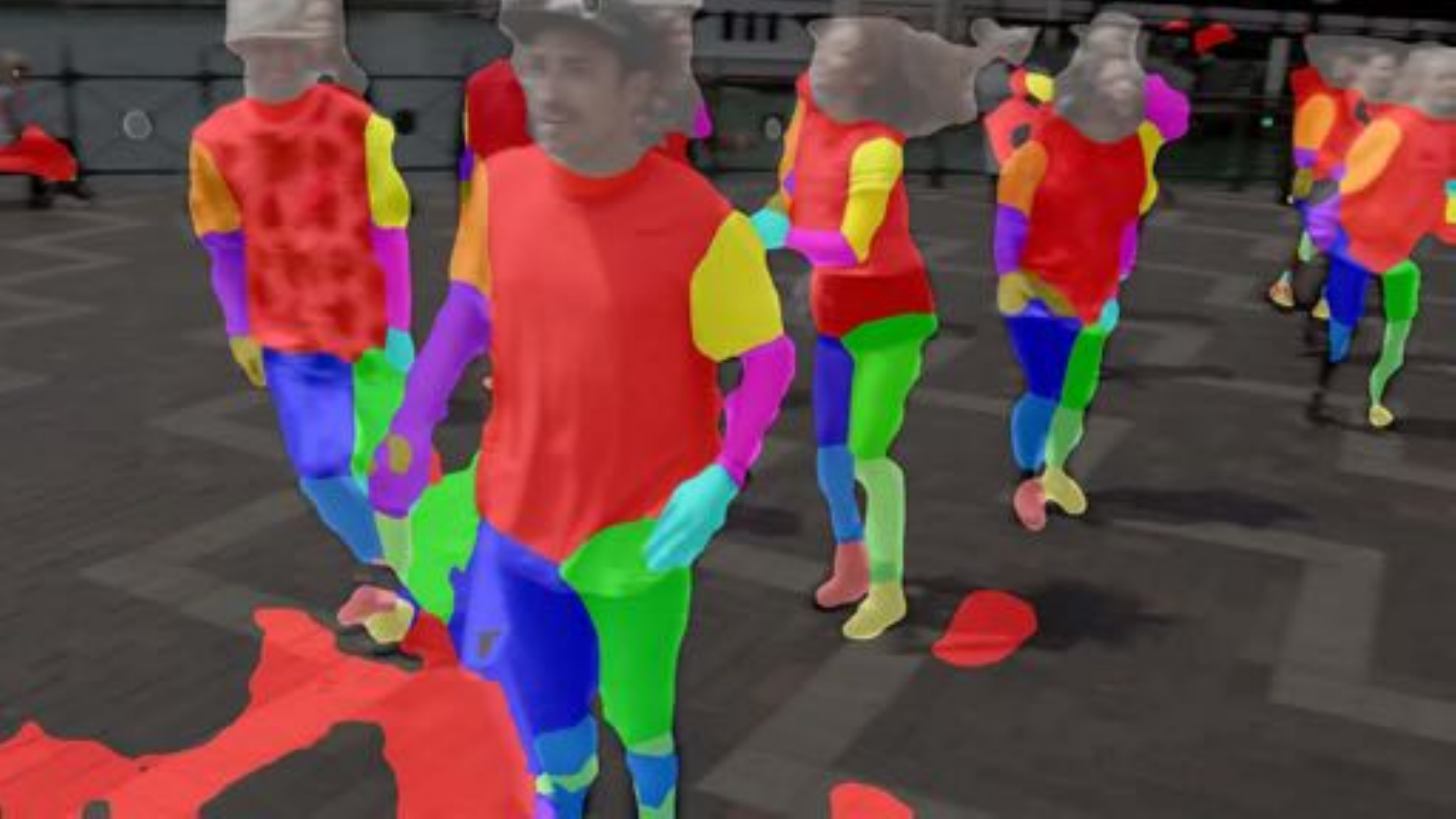}
 	&\includegraphics[width=.245\textwidth]{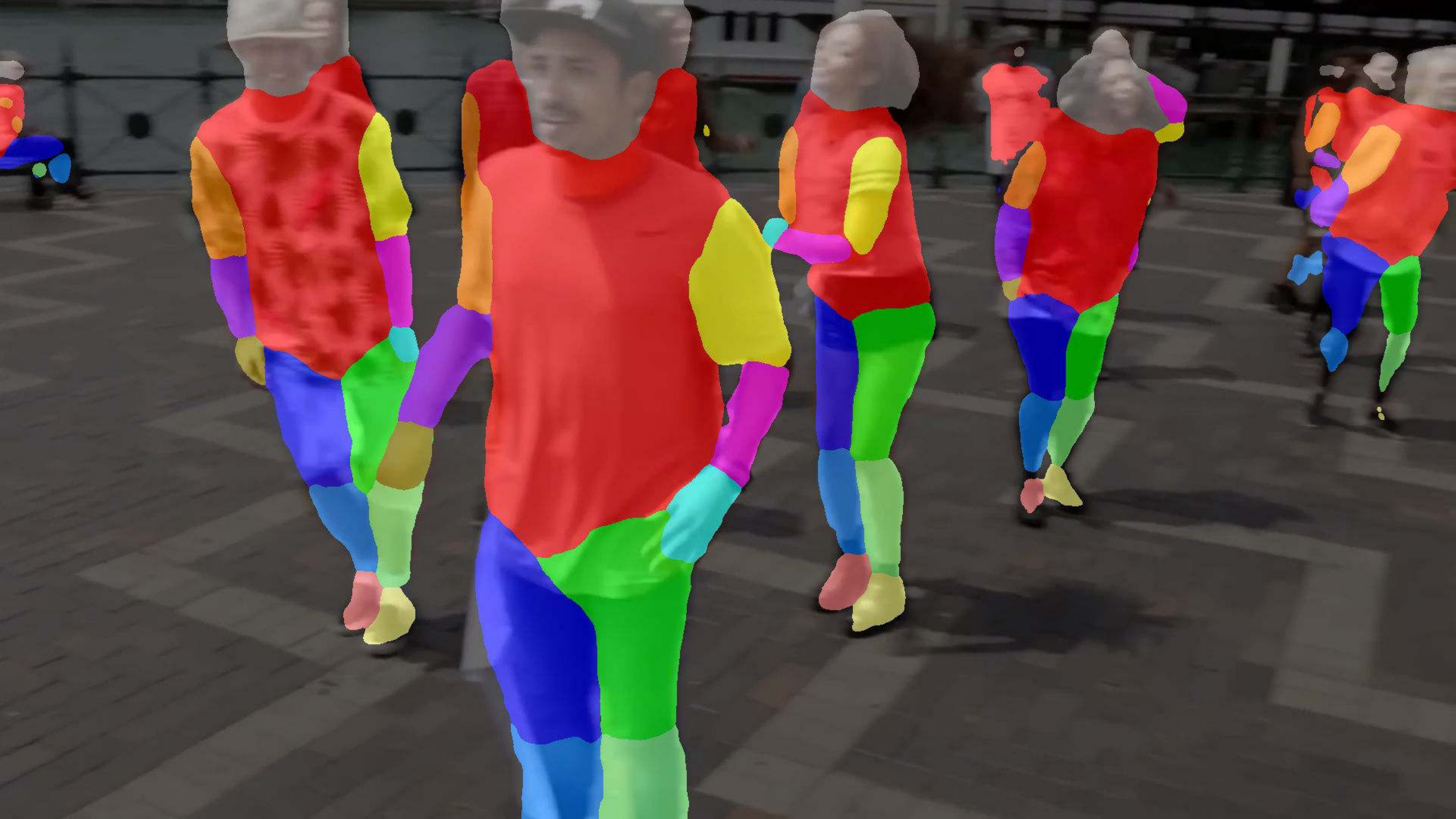}\\
  	\includegraphics[width=.245\textwidth]{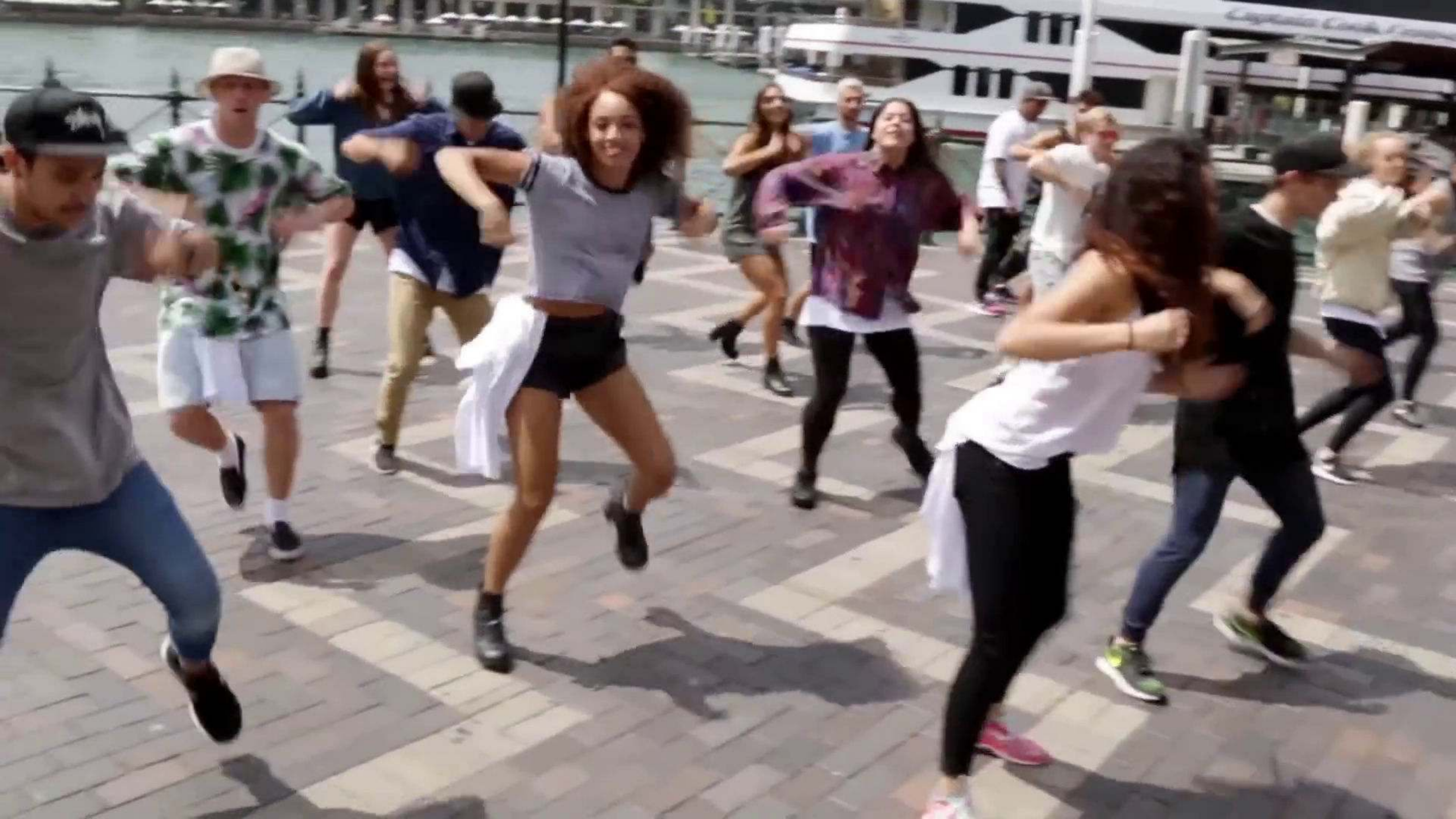}
 	&\includegraphics[width=.245\textwidth]{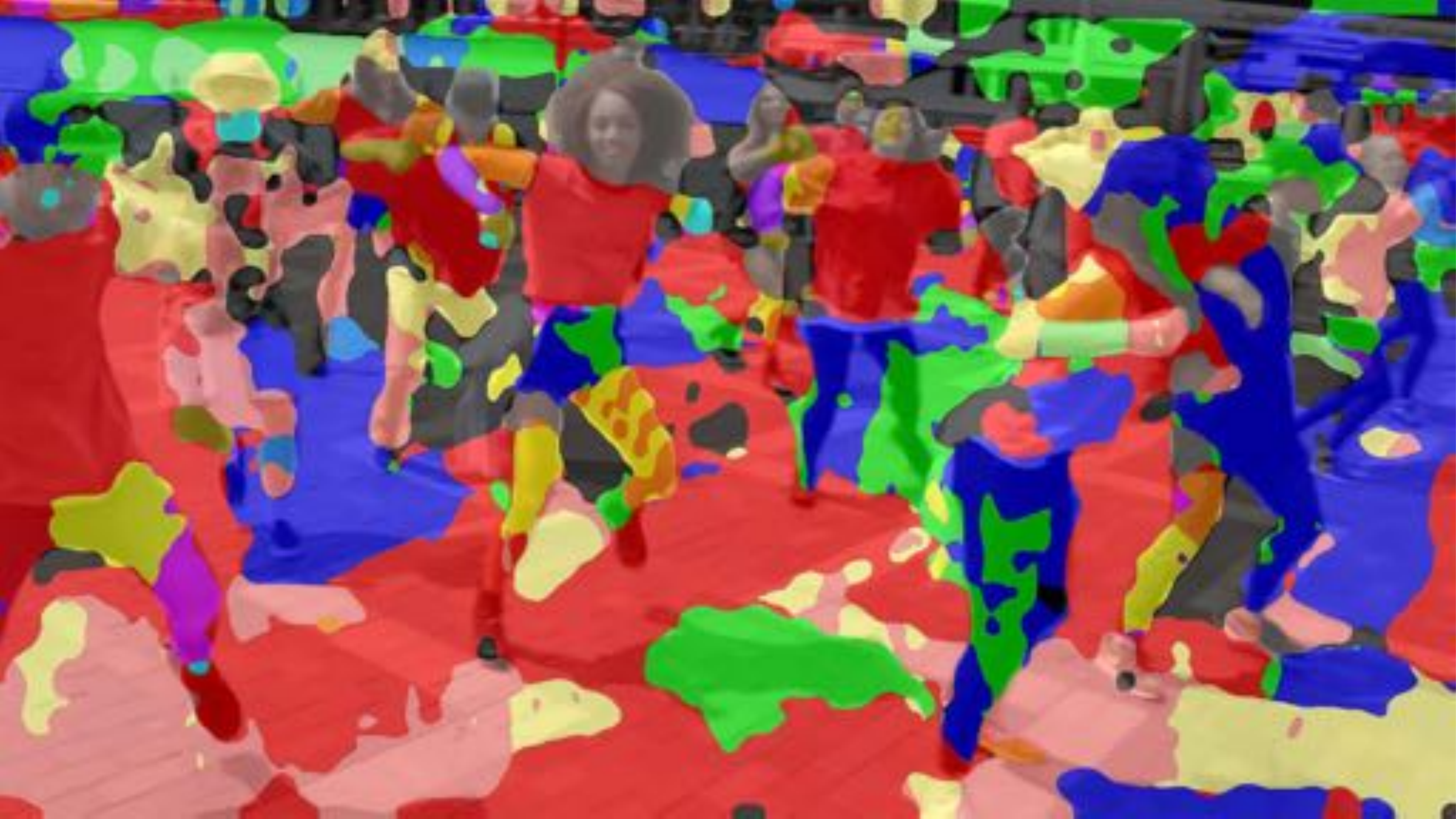}
    &\includegraphics[width=.245\textwidth]{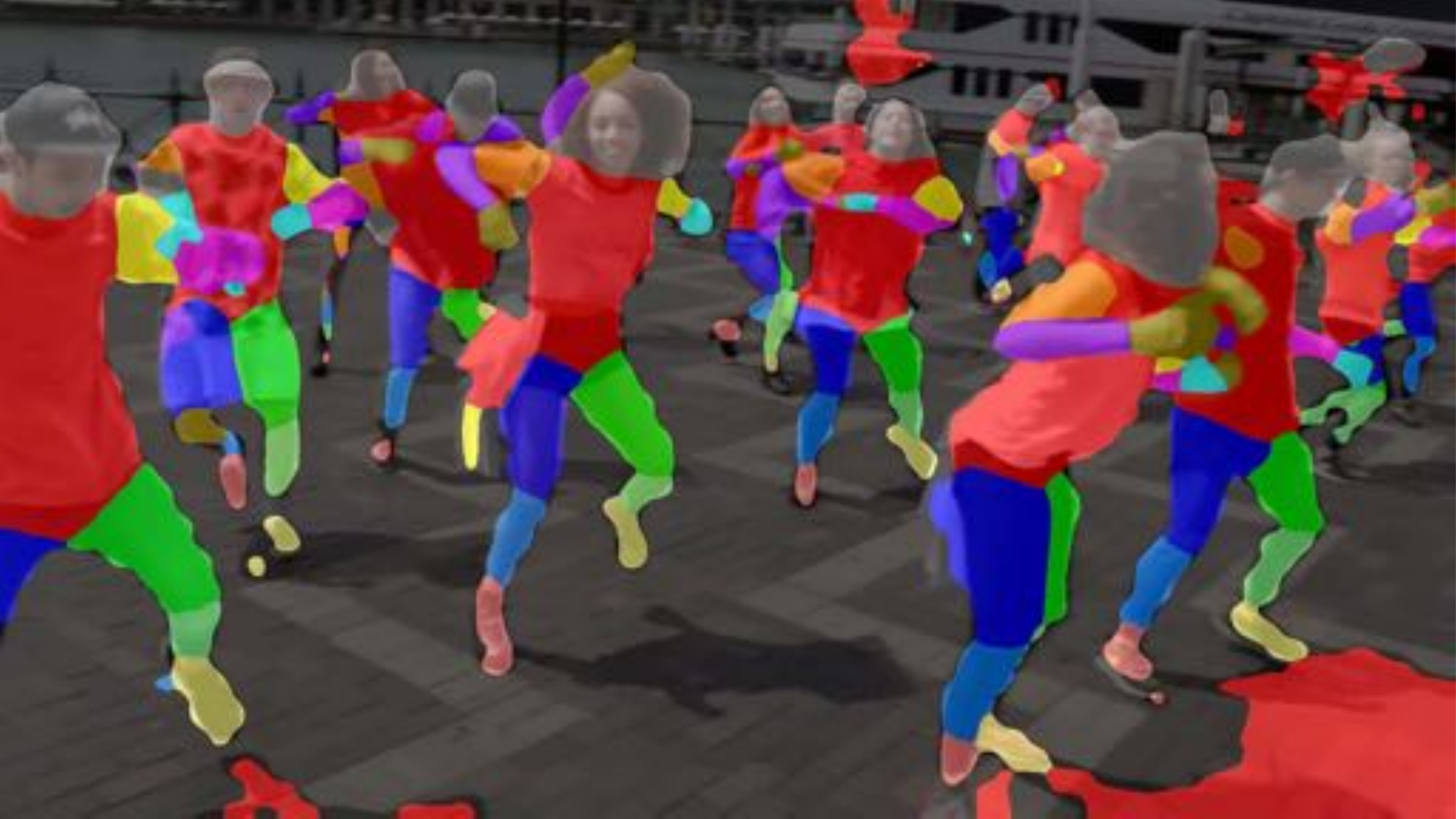}
 	&\includegraphics[width=.245\textwidth]{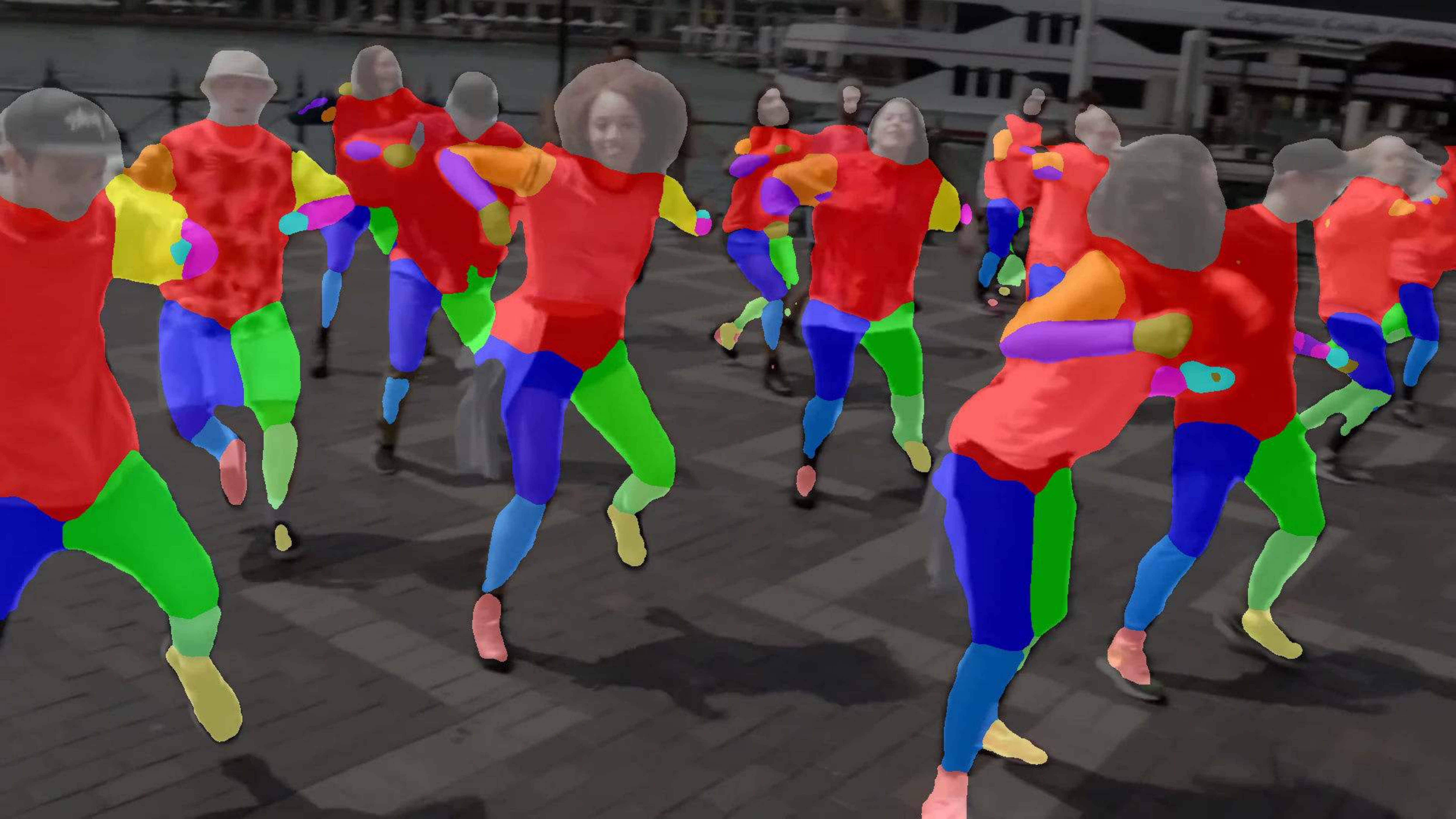}\\
   	\includegraphics[width=.245\textwidth]{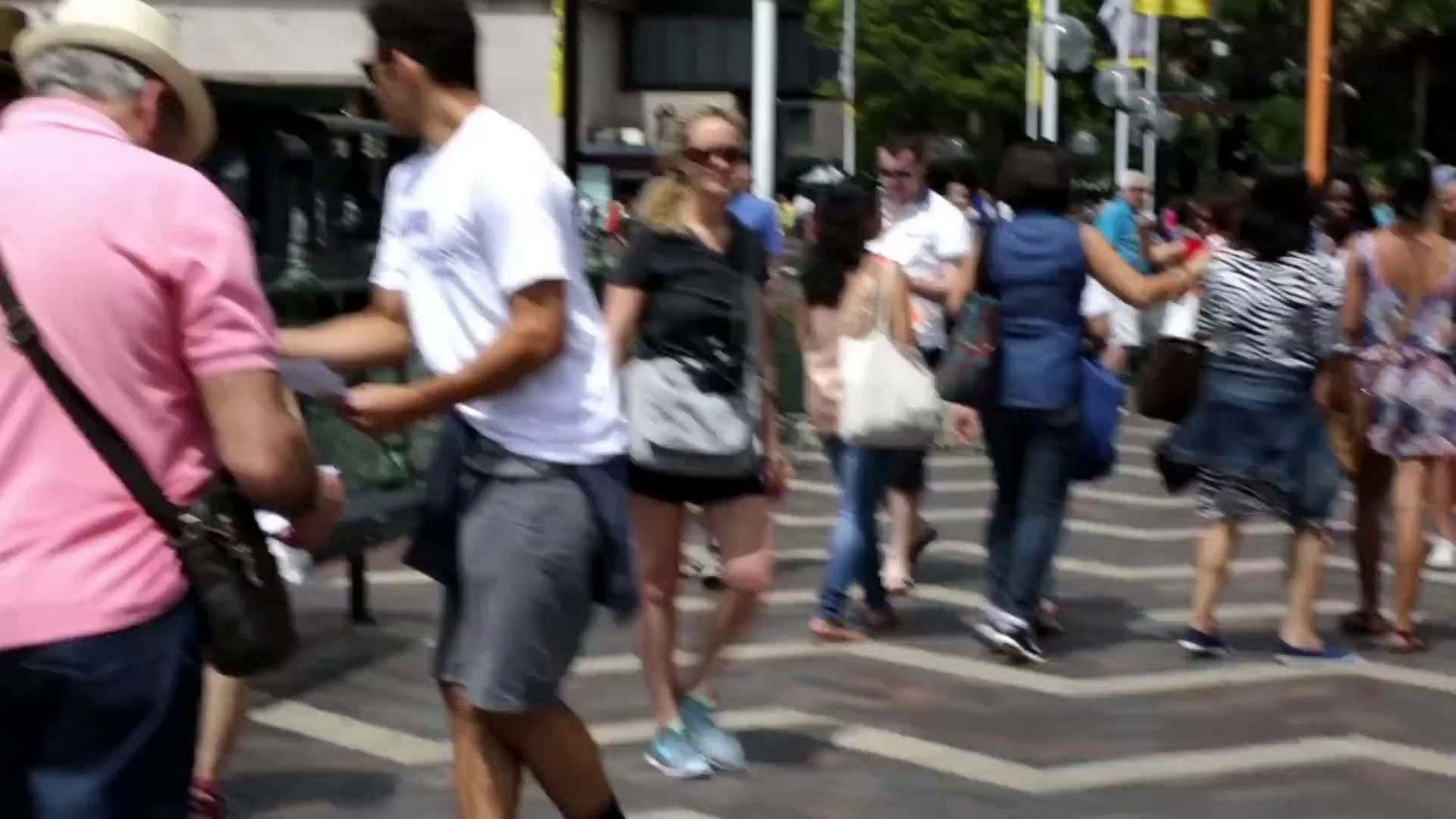}
 	&\includegraphics[width=.245\textwidth]{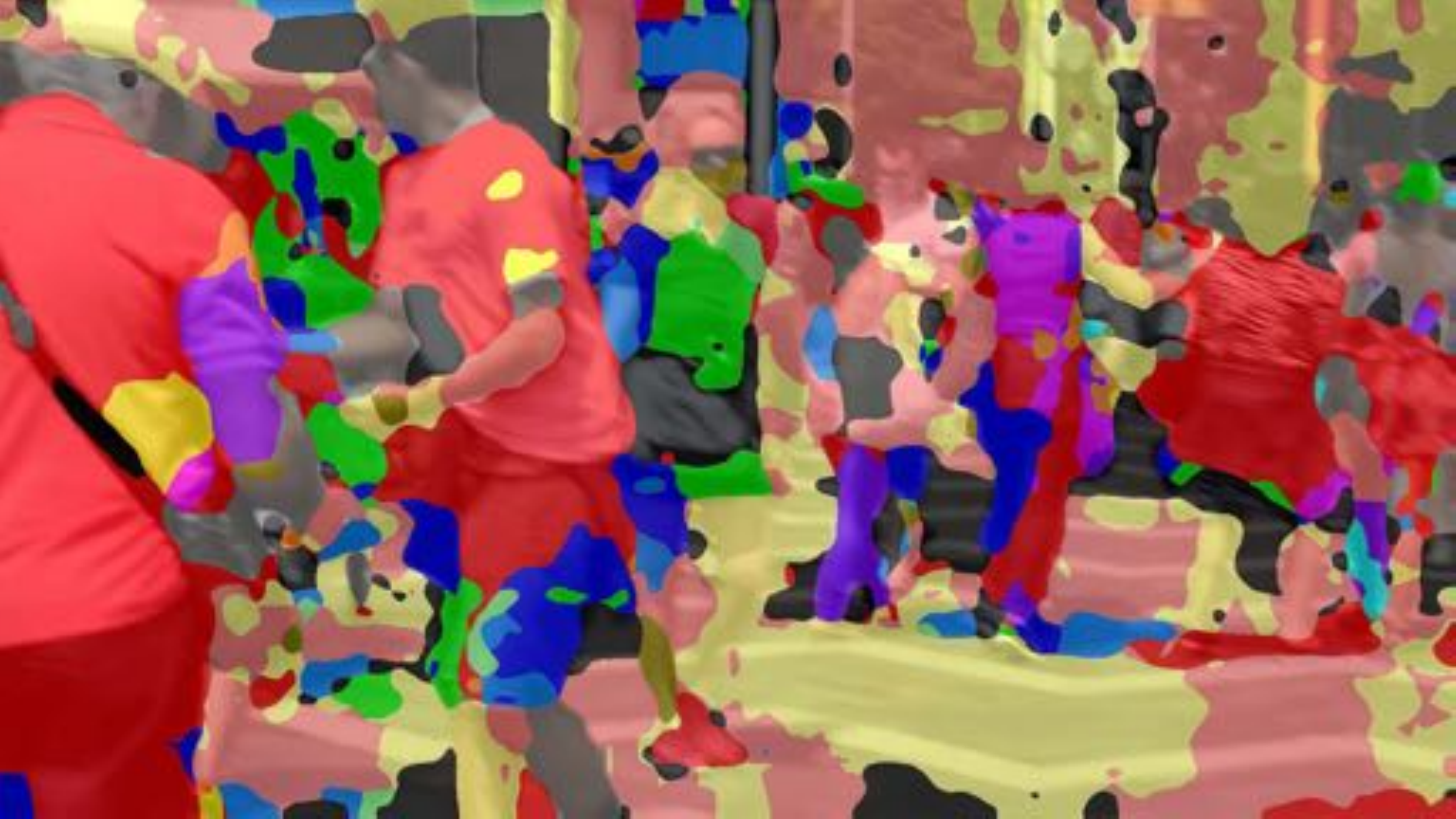}
    &\includegraphics[width=.245\textwidth]{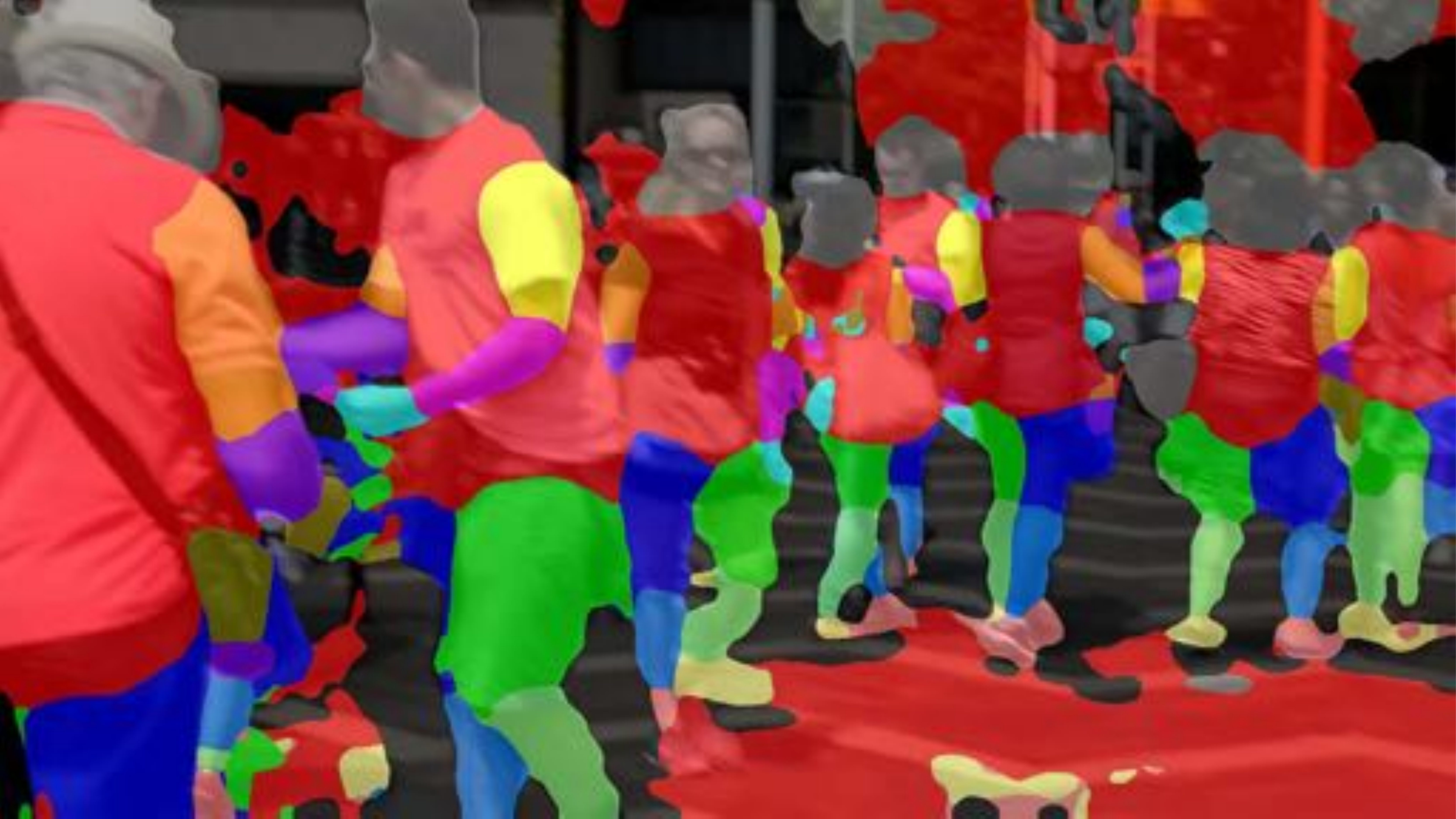}
 	&\includegraphics[width=.245\textwidth]{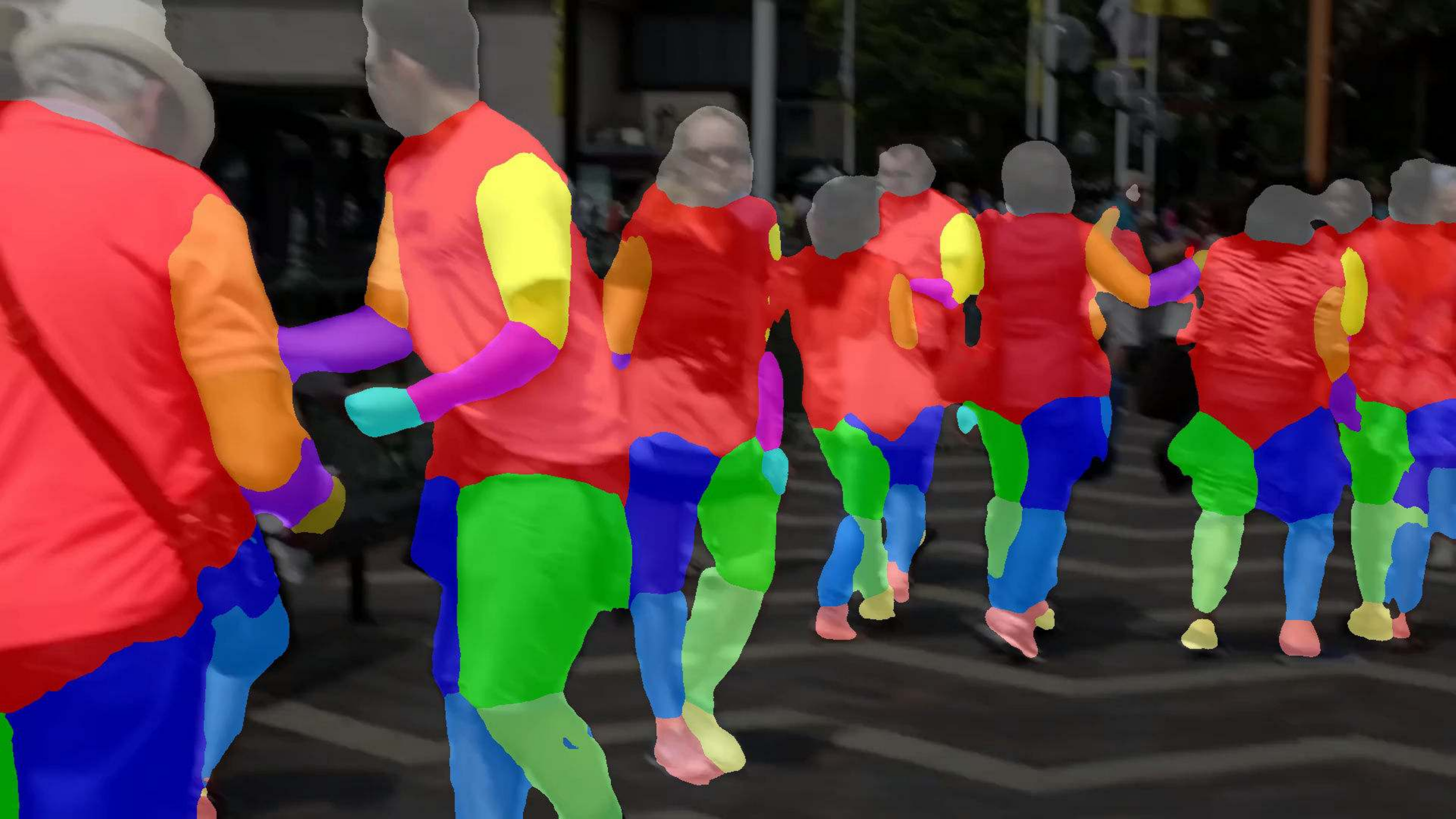}\\
 	Input & SYN & ADV & Ours\\
    \end{tabular}
 \caption{Qualitative comparison on a challenging video~\cite{youtube-video}. The left column shows the input images. The second column from the left shows the results of training using synthetic data only (SYN). The third column from the left shows the results of adversarial training (ADV). The right column shows our results. We can see that SYN failed completely for the real-world images due to domain gap. ADV produces many false alarms in the background. In contrast, our approach performs better than the previous approaches on the tested frames. More video results can be found at \href{https://youtu.be/8QaGfdHwH48}{https://youtu.be/8QaGfdHwH48}}
\label{fig:visual} 
\end{figure*}

We create $30$ novel keypoints for each avatar in the graphics simulator, and use the proposed method to learn the new set of keypoints. Figure~\ref{fig:joint-define} shows the definition of the novel keypoints. To enable our existing network to learn such a new task, we add two additional head networks in our framework to learn the newly created $30$ keypoints and their Part Affinity Fields, resulting in a total of $5$ head networks in our network architecture. Figure~\ref{fig:addition-kpts} shows the qualitative results of our novel keypoint detection.
With small modifications of the existing network, our method learns the novel skeleton representations from the synthetic data and transfers the knowledge to the real domain. It eliminates the needs of ground truth labeling of the additional joints on the real data.

\section{Qualitative comparison}
Recent study~\cite{varol2017learning} proposed to estimate body part segmentation by learning with synthetic data, which is closely related to our method. 
Since MPII dataset~\cite{andriluka20142d} does not have part segmentation labels for quantitative evaluation, Varol~\etal~\cite{varol2017learning} showed qualitative results on selected images from MPII. 
Given a test image, Varol~\etal~\cite{varol2017learning} used additional preprocessing to normalize the input. From their results on MPII dataset with multiple people, it appears that they cropped each image centered at a specific person before feeding to their network. In contrast, our method does not require such preprocessing. Furthermore, our method produces better results as shown in Figure~\ref{fig:demo}. For each example, we show the original image from MPII dataset~\cite{andriluka20142d}, our part segmentation result on the original image, the cropped version which was used as the network input in~\cite{varol2017learning}, and the part segmentation result of~\cite{varol2017learning}.

We further conduct qualitative comparison with the adversarial training approach on a challenging video~\cite{youtube-video}. We compare our method with the adversarial network models presented in Sec~\ref{sec:adversarial}, and Figure~\ref{fig:visual} shows the results. We can see that our method performs consistently better than the previous baseline approaches on the tested frames. The results validate the effectiveness of the proposed approach.
\section{Conclusion}\label{sec:conclusion}
We presented a cross-domain complementary learning framework for multi-person part segmentation. Without using any real data part segmentation labels, our method is able to achieve a comparable or better performance than several state-of-the-art techniques that use real part segmentation data for training. We further demonstrated that our technique can also be used to learn novel keypoint detection from synthetic data.


%

\section*{Acknowledgment}

We would like to thank Alvin Chia, Jon Hanzelka, and Pedro Urbina for their help with the synthetic data generation. We would like to thank Jamie Shotton for his support.

\ifCLASSOPTIONcaptionsoff
  \newpage
\fi



\bibliographystyle{IEEEtran}


%

%

\begin{IEEEbiography}[{\includegraphics[width=1in,height=1.25in,clip,keepaspectratio]{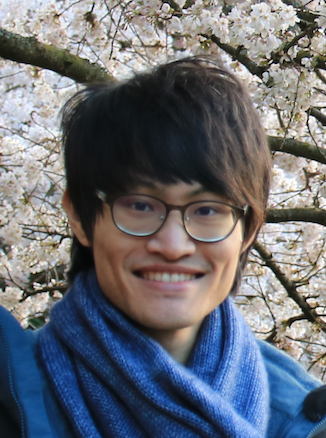}}]{Kevin Lin} (S'10) received the M.S. degree from the Graduate Institute of Networking and Multimedia, National Taiwan University, in 2014. He is currently working toward the Ph.D. degree in electrical engineering at the University of Washington. During his study, he was a research intern at Microsoft Research. His research interests include computer vision, machine learning, and natural language processing.
\end{IEEEbiography}

\begin{IEEEbiography}[{\includegraphics[width=1in,height=1.25in,clip,keepaspectratio]{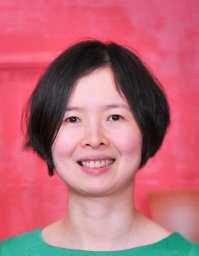}}]{Lijaun Wang} received the B.E. degree from Huazhong University of Science and Technology, and the Ph.D. degree from Tsinghua University, in 2001 and 2006 respectively. She joined Microsoft Research in 2006, where she is currently a principle research manager. She has been the key contributor in developing technologies on computer vision, printed and handwritten text recognition, avatar animation, and speech synthesis and recognition, which have been shipped into various Microsoft products. Current research interests include vision-language pre-training, image captioning, human pose estimation and part segmentation, and machine learning. She was part of the team that developed Azure Kinect Body Tracking SDK. She has published more than 40 papers on the top conferences and journals, and she has been awarded more than 15 US patents. She is a senior member of IEEE.
\end{IEEEbiography}

\begin{IEEEbiography}[{\includegraphics[width=1in,height=1.25in,clip,keepaspectratio]{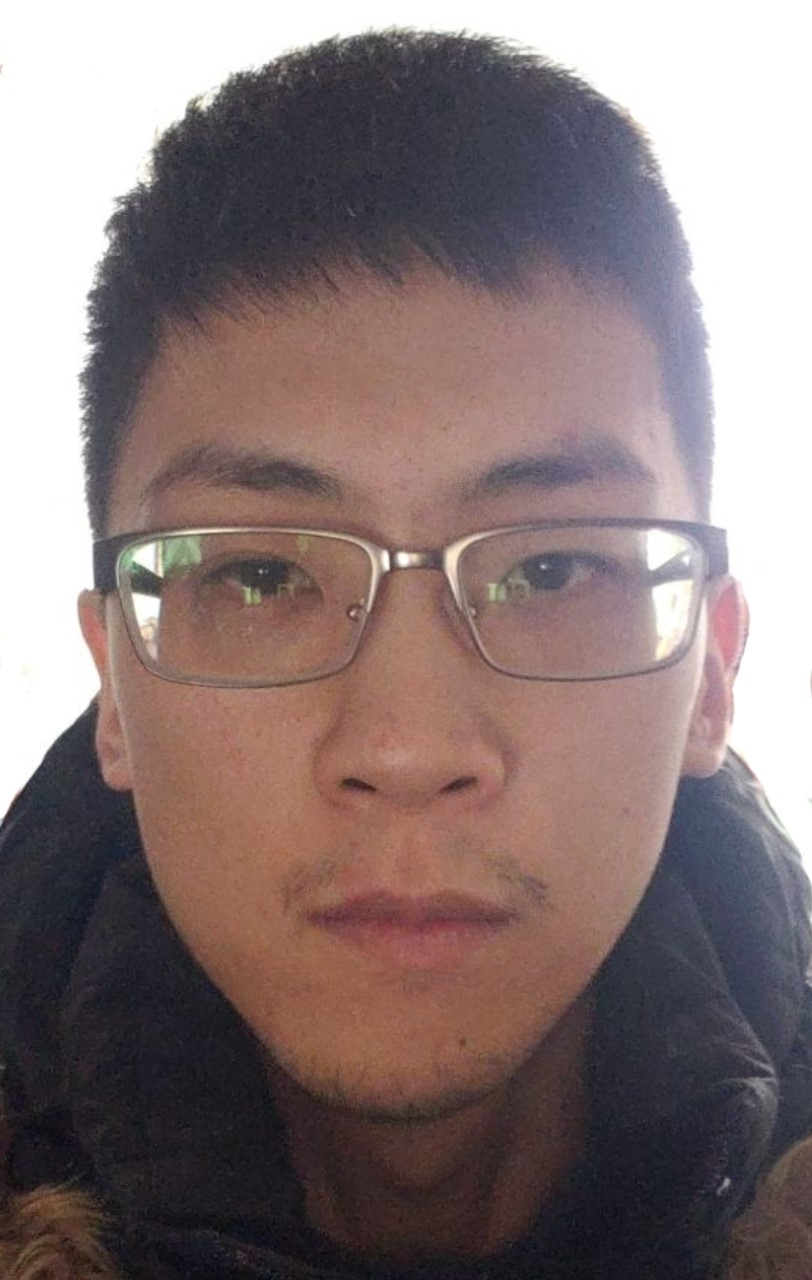}}]{Kun Luo} received his M.S. degree in computer science from University of California San Diego. He is currently a Research Software Development Engineer at Microsoft. He mainly works on building systems solving computer vision problems via deep learning methods. 
\end{IEEEbiography}

\begin{IEEEbiography}[{\includegraphics[width=1in,height=1.25in,clip,keepaspectratio]{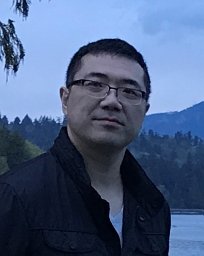}}]{Yinpeng Chen} received the Ph.D. degree in electrical engineering from Arizona State University, Tempe, in 2009. He received the B.S. and M.S. degree in electrical engineering from Tsinghua University, Beijing, in 2000 and 2003, respectively. He is currently a researcher at Microsoft. His current research interests include computer vision and multimedia. 
\end{IEEEbiography}

\begin{IEEEbiography}[{\includegraphics[width=1in,height=1.25in,clip,keepaspectratio]{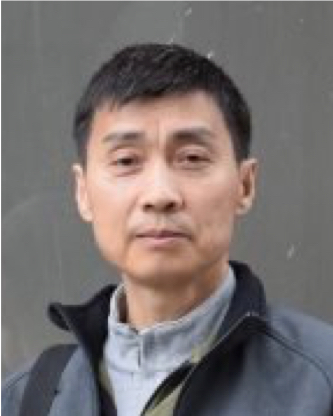}}]{Zicheng Liu} (SM'05-F'15) received the B.S. degree in mathematics from Huazhong Normal University, Wuhan, China, in 1984, the M.S. degree in operations research from the Institute of Applied Mathematics, Chinese Academy of Sciences, in 1989, and the Ph.D. degree in computer science from Princeton University in 1996. He joined Microsoft Research in 1997, and he is currently a principal research manager. Before joining Microsoft Research, he was with Silicon Graphics, Inc., as a Member of Technical Staff for two years, where he developed the trimmed NURBS tessellator shipped in both OpenGL and the OpenGL Optimizer. Current research interests include human activity recognition, human pose estimation and part segmentation, and machine learning. He was part of the team that developed Azure Kinect Body Tracking SDK. 

Liu has co-authored three books. He served as the chair of the Multimedia Systems and Applications Technical Committee of IEEE CAS society. He was a steering committee member of \textit{IEEE Transactions on Multimedia}. He is the Editor-in- Chief of \textit{Journal of Visual Communications and Image Representation}. He is an affiliate professor in the department of Electrical and Computer Engineering, University of Washington. He was an IEEE distinguished lecturer from 2015-2016. He is a fellow of IEEE.
\end{IEEEbiography}

\begin{IEEEbiography}[{\includegraphics[width=1in,height=1.25in,clip,keepaspectratio]{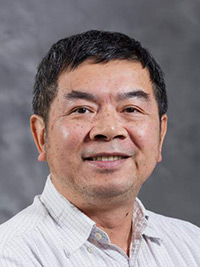}}]{Ming-Ting Sun} (S'79-M'81-SM'89-F'96-LF'20) received the B.S. degree from National Taiwan University and the Ph.D. degree from University of California, Los Angeles, all in electrical engineering.

Dr. Sun joined the University of Washington in 1996 where he is a Professor. Previously, he was the Director of Video Signal Processing Research at Bellcore. He has been a chaired/visiting professor at several universities. His main research interest is video and multimedia signal processing.

Dr. Sun holds 13 patents and has published about 300 technical papers, including 17 book chapters in the area of video and multimedia technologies.  He co-edited a book ``Compressed Video over Networks''.  He has guest-edited 12 special issues for various journals and given keynotes for several international conferences. He was an Editor-in-Chief of the \textit{Journal for Visual Communication and Image Representation (JVCI)} from 2012 to 2016, the Editor-in-Chief for \textit{IEEE Transactions on Multimedia (TMM)} from 2000 to 2001, and the Editor-in-Chief of the \textit{IEEE Transactions on Circuits and Systems for Video Technology (TCSVT)} from 1995 to 1997. He was a Distinguished Lecturer of the Circuits and Systems Society from 2000 to 2001. He received an IEEE CASS Golden Jubilee Medal in 2000. He served as a General Co-Chair of ICME (International Conference on Multimedia and Expo) 2016, an Honorary Chair of VCIP (Visual Communication and Image Processing) 2015, a General Co-Chair chair of Visual Communications and Image Processing 2000. He served as the Chair of the VSPC (Visual Signal Processing and Communications) Technical Committee of IEEE CAS (Circuits and Systems) Society from 1993 to 1994. He received the \textit{TCSVT} Best Paper Award in 1993.  From 1988 to 1991, he was the chairman of the IEEE CAS Standards Committee and established the IEEE Inverse Discrete Cosine Transform Standard. He is a Life Fellow of IEEE.

\end{IEEEbiography}




\end{document}